\documentclass{article}

\newif\ifanon

\ifanon
  \usepackage[dblblindworkshop]{neurips_2026}
\else
  \usepackage[dblblindworkshop,final]{neurips_2026}
\fi

\usepackage[utf8]{inputenc}
\usepackage[T1]{fontenc}
\usepackage[hypertexnames=false]{hyperref}
\usepackage{url}
\usepackage{booktabs}
\usepackage{amsfonts}
\usepackage{nicefrac}
\usepackage{microtype}
\usepackage[table]{xcolor}
\usepackage{pifont}
\usepackage{longtable} 
\usepackage{pdflscape} 
\usepackage{graphicx} 
\usepackage{float} 
\usepackage{siunitx} 

\definecolor{linkbrown}{RGB}{140,70,20}
\hypersetup{colorlinks=true,linkcolor=linkbrown,citecolor=linkbrown,urlcolor=linkbrown}
\definecolor{tablegreen}{RGB}{0,125,55}
\definecolor{tablered}{RGB}{190,35,35}
\definecolor{tableorange}{RGB}{180,105,0}
\newcommand{\cmark}{\textcolor{tablegreen}{\ding{51}}}
\newcommand{\xmark}{\textcolor{tablered}{\ding{55}}}
\newcommand{\pmark}{\textcolor{tableorange}{\(\circ\)}}

\title{MIKASA-Robo-VLA: Benchmarking Memory in VLA Models for Long-Horizon Manipulation}
\workshoptitle{RoboPAD}

\ifanon
\else
  \author{%
    \textbf{Egor Cherepanov}$^{1,2}$
    \quad \textbf{Nikita Kachaev}$^{1, 3}$
    \quad \textbf{Aleksandr I. Panov}$^{1,2}$
    \quad \textbf{Alexey K. Kovalev}$^{1,2}$ \\
    $^1$AXXX, Moscow, Russia
    \quad $^2$MIRIAI, Moscow, Russia
    \quad $^3$HSE University, Moscow, Russia\\
    {\tt\small cherepanov@axxx.tech} \\[2pt]
    \textbf{\href{https://mikasarobo.github.io/}{mikasarobo.github.io}}
  }
\fi

\ifanon
  \newcommand{\repourl}{public, location withheld for double-blind review}
  \newcommand{\dataurl}{public, location withheld for double-blind review}
  
\else
  \newcommand{\repourl}{\url{https://github.com/CognitiveAISystems/MIKASA-Robo}}
  \newcommand{\dataurl}{RLDS: \url{https://huggingface.co/datasets/mikasa-robo/mikasa-robo-vla-rlds}; LeRobot: \url{https://huggingface.co/datasets/mikasa-robo/mikasa-robo-vla-lerobot}}
  
\fi

\newcommand{\NumTasks}{90}
\newcommand{\NumEpisodes}{22{,}500}
\newcommand{\NumFrames}{6{,}306{,}953}
\newcommand{\NumHours}{175.2}
\newcommand{\NumMemoryTypes}{10}
\newcommand{\NumHorizonSplits}{3}
\newcommand{\MinHorizon}{25}
\newcommand{\MaxHorizon}{2{,}160}
\newcommand{\MeanOracleSuccess}{1.000}
\newcommand{\StateDim}{7}
\newcommand{\ActionDim}{7}
\newcommand{\ImageShape}{128\times128\times3}
\newcommand{\ControlRateHz}{10}
\newcommand{\BaselineMean}{0.211}
\newcommand{\BaselineSteps}{139{,}999}
\newcommand{\BaselineEpisodesPerTask}{20}
\newcommand{\BaselineUpscale}{224}
\newcommand{\BaselineActionDimPadded}{32}
\newcommand{\BaselineGlobalBatch}{128}
\newcommand{\BaselineNumGpus}{8}
\newcommand{\BaselineEvalBaseSeed}{10{,}000}
\newcommand{\NumExcludedTasks}{5}
\newcommand{\GripperRangeLower}{-0.01}
\newcommand{\GripperRangeUpper}{0.04}
\newcommand{\NumGapDetermined}{70}

\newcommand{\NumTasksAbovePublishedContext}{28}
\newcommand{\PublishedContextFrames}{16}
\newcommand{\NumTasksZeroGap}{15}
\newcommand{\NumGapNoRetention}{10}
\newcommand{\NumGapUndetermined}{10}
\newcommand{\NumBaselineGapNoRetention}{2}
\newcommand{\NumBaselineGapUndetermined}{3}
\newcommand{\BaselineTrainEpisodes}{3{,}500}
\newcommand{\BaselineTrainFrames}{798{,}296}
\newcommand{\NumBaselineTasks}{14}
\newcommand{\BaselineSEMin}{0.049}
\newcommand{\BaselineSEMax}{0.112}
\newcommand{\RldsSizeGiB}{287.4}
\newcommand{\LerobotSizeGiB}{25.9}
\newcommand{\NumTasksLengthsAmbiguous}{90}
\newcommand{\NumTasksPPO}{34}
\newcommand{\NumTasksMP}{56}
\newcommand{\NumStripFrames}{8}
\newcommand{\NumFamilies}{26}
\newcommand{\EpisodesPerTask}{250}
\newcommand{\BaselineFramePercent}{12.7}
\newcommand{\NumDistinctInstructions}{30}
\newcommand{\NumTasksSharedInstruction}{80}
\newcommand{\RandomFloorTasks}{90}
\newcommand{\RandomFloorEpisodesPerTask}{50}
\newcommand{\RandomFloorEpisodes}{4{,}500}
\newcommand{\RandomFloorCorroborationSplit}{Short}
\newcommand{\RandomFloorSplitEpisodes}{1{,}900}
\newcommand{\RandomFloorSplitSuccesses}{0}
\newcommand{\RandomFloorSuccesses}{1}
\newcommand{\RandomFloorEndSuccesses}{0}
\newcommand{\RandomFloorZeroTasks}{89}
\newcommand{\RandomFloorNonzeroTask}{ShellGameShuffleTouch-Long-VLA-v0}
\newcommand{\RandomFloorNonzeroRate}{0.02}
\newcommand{\RandomFloorNonzeroSE}{0.020}
\newcommand{\RandomFloorStartSeed}{500{,}000}
\newcommand{\RandomFloorCurriculumTasks}{23}
\newcommand{\RandomFloorCorroborationRate}{0.000526}

\newcommand{\RandomFloorSimBackend}{cpu}
\newcommand{\RandomFloorManiSkillVersion}{3.0.0b14}
\newcommand{\RandomFloorGymnasiumVersion}{0.29.1}
\newcommand{\RandomFloorTorchVersion}{2.2.1+cu121}
\newcommand{\RandomFloorPythonVersion}{3.11.13}
\newcommand{\NumCueKeyTasks}{90}
\newcommand{\NumCueLeakTasks}{10}
\newcommand{\NumCueLeakReactive}{4}
\newcommand{\NumCueLeakConcealed}{6}

\newcommand{\NumPredecessorTypes}{4}
\newcommand{\NumChanceFloorEvaluated}{14}
\newcommand{\NumChanceFloorTasks}{7}
\newcommand{\NumChanceFloorWithinSE}{5}
\newcommand{\NumChanceFloorBelow}{2}
\newcommand{\NumChanceFloorAbove}{0}
\newcommand{\ChanceFloorPooledEpisodes}{140}
\newcommand{\ChanceFloorPooledSuccesses}{33}
\newcommand{\ChanceFloorPooledExpected}{35.3}
\newcommand{\ChanceFloorPooledSD}{5.1}
\newcommand{\ChanceFloorCandidateSet}{$3$, $5$, and $6$}
\newcommand{\RldsSchemaDistinctDigests}{1}
\newcommand{\RldsSchemaDigestAlgorithm}{MD5}
\newcommand{\RldsSchemaDigest}{\texttt{01a88910206da914198d39bd053a8def}}

\newcommand{\TypesAllGapsZero}{Spatial and Procedural}

\newcommand{\TypesNoDeterminedGap}{Prospective and Checklist}
\newcommand{\BaselineLongTasks}{3}
\newcommand{\BaselineLongSuccesses}{1}
\newcommand{\BaselineLongEpisodes}{60}
\newcommand{\BaselineLongMean}{0.017}
\newcommand{\BaselineNonLongTasks}{11}
\newcommand{\BaselineNonLongSuccesses}{58}
\newcommand{\BaselineNonLongEpisodes}{220}
\newcommand{\BaselineNonLongMean}{0.264}
\newcommand{\BaselineHorizonP}{0.008}
\newcommand{\BaselineMeanSE}{0.044}

\newcommand{\BaselineChunk}{8}

\newcommand{\NumBaselineMemoryTypes}{7}
\newcommand{\BaselineUnevaluatedTypes}{Temporal, Negative, and Checklist}
\newcommand{\NumBaselineGapDetermined}{9}
\newcommand{\NumBaselineNoDeterminedGap}{5}
\newcommand{\BaselineDeterminedGapMean}{0.167}
\newcommand{\BaselineNoDeterminedGapMean}{0.290}
\newcommand{\BaselineGapGroupDiff}{0.123}
\newcommand{\BaselineGapGroupP}{0.209}

\newcommand{\BaselineGapGroupDiffNoLong}{0.048}
\newcommand{\BaselineGapGroupPNoLong}{0.617}
\newcommand{\BaselineGapSpearman}{0.161}
\newcommand{\BaselineGapSpearmanP}{0.673}

\newcommand{\ExecRegimeMaxDelta}{0.650}
\newcommand{\ChunkHighlightTask}{RememberColor5-Medium}
\newcommand{\ChunkHighlightDelayMin}{37}
\newcommand{\ChunkHighlightDelayMax}{40}

\newcommand{\ChunkShortHighlightSuccess}{0.080}
\newcommand{\ChunkLongHighlightSuccess}{0.540}

\begin{document}

\maketitle

\vspace{-2em}
\begin{center}
  \includegraphics[width=\linewidth]{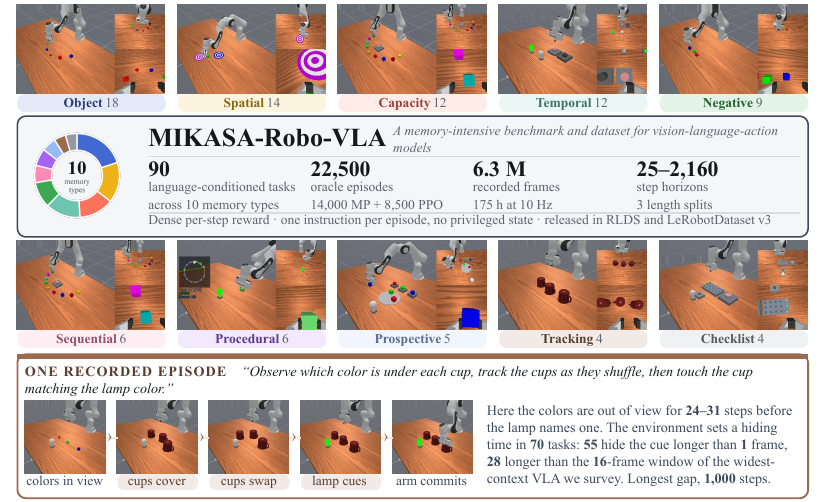}
\end{center}

\begin{abstract}
Vision-language-action policies often see only one or a few recent frames,
which makes it difficult to evaluate how they use information that disappears
during a task. We introduce MIKASA-Robo-VLA, a benchmark of \NumTasks{}
language-conditioned manipulation tasks.
All but \NumGapNoRetention{} hide the cue an action depends on. Those
\NumGapNoRetention{} are reactive controls.
MIKASA-Robo, the suite it rebuilds, has 32 tasks and uses language only in a
representative VLA subset. Here every task provides an instruction, while
memory-dependent tasks hide a task-relevant cue and reactive controls keep it
available. For \NumGapDetermined{} tasks, environment phase timings specify an
information gap, and for \NumTasksAbovePublishedContext{} of them the gap
exceeds the \PublishedContextFrames{}-frame window of the widest
fixed-context VLA we survey. The gap counts only the interval the cue is
provably absent, not the full duration a policy must retain it, so every
memory-dependent task still requires memory by construction, including the
ones whose measured gap is short. We release \NumEpisodes{}
oracle trajectories across \NumMemoryTypes{} memory types in RLDS and
LeRobotDataset v3.
A reference \(\pi_{0.5}\) baseline with current images and proprioception,
but no observation history or explicit memory module, is fine-tuned on
\NumBaselineTasks{} tasks and achieves $\BaselineMean{} \pm \BaselineMeanSE{}$
mean task success. Its lower success on the evaluated Long-split tasks is
confounded by open-loop chunking and the memory types represented in that
subset.
\ifanon\else
Project page: \mbox{\href{https://mikasarobo.github.io/}{mikasarobo.github.io}}.
\fi
\end{abstract}

\section{Introduction}
\label{sec:introduction}

Vision-language-action (VLA) models increasingly use pretrained visual and
linguistic representations to control robots from demonstrations. Models such
as OpenVLA and $\pi_{0.5}$ execute diverse manipulation skills and generalize
across tasks and environments~\citep{kim2024openvla,intelligence2025pi05}.
Sustained interaction makes memory a central requirement: a robot must act on
information that may no longer be visible.
It may need to recall an object's identity after an occlusion, reproduce a
previously observed sequence, or track which steps of a task have already
been completed. The instruction specifies the goal, but the observation
history determines how to achieve it.

Explicit-memory methods address this challenge in different ways. MemER
retrieves relevant visual experiences for hierarchical
control~\citep{sridhar2025memer}, while MEM combines visual and textual
memory at different temporal scales~\citep{torne2026mem}. Their physical
setups make systematic memory sweeps costly to reproduce. Simulation offers
a more scalable alternative, yet
widely used suites such as CALVIN and LIBERO emphasize skill composition
and knowledge transfer~\citep{mees2022calvin,liu2023libero} rather than
explicit control of what must be remembered and for how long. In complex multi-stage
tasks, aggregate success also mixes memory demands with perception, planning,
and motor execution. Memory-focused suites, including MemoryBench,
LIBERO-Mem, RoboMME, and RoboMemArena, make these demands more explicit, but
cover a narrower range of tasks and memory types
\citep{fang2025sam2act,chung2026liberomem,dai2026robomme,lei2026robomemarena}.

We introduce \textbf{MIKASA-Robo-VLA}, an extension of
MIKASA-Robo~\citep{cherepanov2026mikasarobo}, originally developed to evaluate
memory in reinforcement-learning (RL) agents, to language-conditioned VLA
training and evaluation. The suite expands from 32 tasks and
\NumPredecessorTypes{} memory types to \NumTasks{} tasks spanning
\NumMemoryTypes{} types, with natural-language instructions and a shared
observation and action interface. It asks two questions: \emph{what must a
policy remember, and for how long?} Task families vary the
content and structure of memory, including object attributes, sequences,
counts, and pending actions. Difficulty variants adjust factors such as the
number of items, sequence length, and delay before recall. Configured episode
horizons range from \MinHorizon{} to \MaxHorizon{} steps, organized into
Short, Medium, and Long splits, and retention intervals are characterized
separately from episode length.

Dense rewards support online RL fine-tuning, and demonstrations for every
task support imitation learning. A common evaluation protocol fixes the
policy interface, episode seeds, and success metric.

Our main contributions are as follows:
\begin{itemize}
  \item \textbf{A broad, parameterized benchmark for memory in VLA policies.}
  We provide \NumTasks{} language-conditioned manipulation tasks across
  \NumMemoryTypes{} memory types, with difficulty variants spanning memory
  content and temporal demands, and three horizon splits covering
  \MinHorizon{}--\MaxHorizon{} steps (Section~\ref{sec:benchmark}).

  \item \textbf{Demonstrations and rewards for policy training.}
  We release \NumEpisodes{} oracle demonstration episodes, comprising
  \NumFrames{} timesteps, collected using PPO policies and motion planning.
  Every task is available in both RLDS and LeRobotDataset v3 formats, with
  dense environment rewards supporting subsequent RL fine-tuning
  (Section~\ref{sec:datasets}).

  \item \textbf{Standardized evaluation and a reference analysis.}
  We provide a common evaluation protocol in the released code, a per-task characterization
  of retention intervals, and a reference $\pi_{0.5}$ baseline on
  \NumBaselineTasks{} tasks (Section~\ref{sec:analysis}).
\end{itemize}

\section{Related Work}
\label{sec:related}

\paragraph{Memory in reinforcement learning.}
RL memory evaluation includes both diagnostic tasks and diverse partially
observable environments. T-Maze and bsuite probe retention across delays
\citep{bakker2001rllstm,osband2020bsuite}, while POPGym and Memory Gym
broaden the range of observations, task structures, and temporal demands
\citep{morad2023popgym,pleines2025memorygym}. Other work separates memory from
credit assignment~\citep{ni2023transformers} or develops taxonomies of memory
capabilities~\citep{cherepanov2025unraveling}. MIKASA-Robo applies this
diagnostic framing to robotic manipulation~\citep{cherepanov2026mikasarobo}.
We extend its RL setting to language-conditioned policies, with an expanded
taxonomy, a common VLA interface, demonstrations, and dense rewards.

\paragraph{Memory architectures for robot policies.}
Generalist policies typically condition actions on limited visual history:
OpenVLA uses a single image~\citep{kim2024openvla}, while RT-1, Octo, and
RoboVLMs use windows of six, two, and \PublishedContextFrames{} frames,
respectively, in their reported main configurations
\citep{brohan2023rt1,octomodelteam2024octo,li2024robovlms}.
Explicit memory extends this design space through perceptual and cognitive
storage in MemoryVLA~\citep{shi2026memoryvla}, recurrent state in
$\mu$VLA~\citep{cherepanov2026uvla}, experience retrieval in
MemER~\citep{sridhar2025memer}, and visual--textual memory at multiple scales
in MEM~\citep{torne2026mem}. MIKASA-Robo-VLA provides tasks for comparing such
designs across memory contents and retention intervals. Its information-gap
analysis relates task structure to observation windows without treating window
size as a limit on recurrent or external memory.

\paragraph{General-purpose manipulation benchmarks.}
CALVIN evaluates language-conditioned skill composition
\citep{mees2022calvin}, LIBERO studies lifelong knowledge transfer
\citep{liu2023libero}, SIMPLER targets correspondence between simulated
and real-world policy performance~\citep{li2024simpler}, and RoboCasa
broadens evaluation to everyday kitchen activities
\citep{nasiriany2024robocasa}. Their standard evaluations do not systematically
vary memory content and retention delay. A long sequence can remain locally
observable, while a short task can depend on a cue that has disappeared.
MIKASA-Robo-VLA makes this distinction explicit.

\paragraph{Memory-dependent manipulation benchmarks.}
MemoryBench introduces spatial-memory and action-recall tasks
\citep{fang2025sam2act}. LIBERO-Mem formalizes four object-centric non-Markov
factors~\citep{chung2026liberomem}, and RMBench distinguishes one-shot from
repeated-exploration memory~\citep{chen2026rmbench}. RoboMME covers temporal,
spatial, object, and procedural memory across sixteen tasks
\citep{dai2026robomme}, while RoboMemArena composes longer tasks and annotates
memory-relevant keyframes~\citep{lei2026robomemarena}. RoboTwin-MeM isolates
transient intermediate visual evidence and explicitly scales the number of
required keyframes~\citep{yang2026eventvla}, and MEMOBench instead diagnoses
storage, update, and compression through executable checkpoints
\citep{sun2026memobench}.
MIKASA-Robo-VLA adds \NumTasks{} parameterized tasks over \NumMemoryTypes{}
memory types (Table~\ref{tab:taxonomy}), dense rewards, a common evaluation
protocol, and \NumEpisodes{} demonstrations in RLDS and LeRobotDataset v3 --
the only suite among these to combine controlled retention and load
variation, horizon stratification, dense rewards, and GPU-parallel execution
(Table~\ref{tab:comparison}, Appendix~\ref{sec:appendix-comparison}).

\section{MIKASA-Robo-VLA Benchmark}
\label{sec:benchmark}

\begin{figure}[t]
  \centering
  \includegraphics[width=\linewidth]{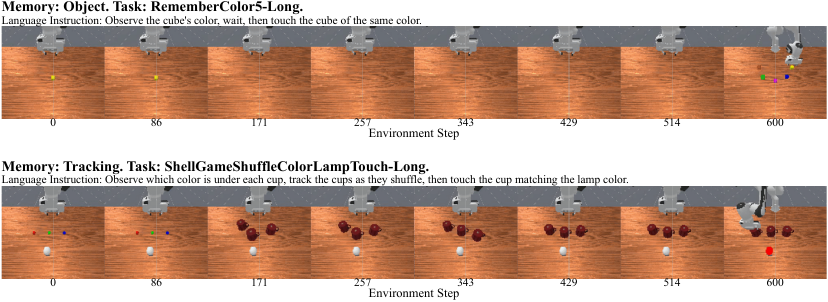}
  \vspace{-8pt}
  \caption{\textbf{Memory-dependent manipulation in MIKASA-Robo-VLA.} Each row shows
  \NumStripFrames{} evenly spaced frames from a 600-step rollout horizon,
  together with the task and language instruction. Ticks locate each frame at
  its nominal environment step. The relevant cue disappears before the
  dependent action begins. Images show the top-camera view at render
  resolution, and policy inputs have resolution $\ImageShape$.
  Appendix~\ref{sec:appendix-contact-sheet} provides rollouts for all tasks.}
  \label{fig:teaser}
\end{figure}

MIKASA-Robo-VLA comprises \NumTasks{} language-conditioned manipulation
tasks implemented in ManiSkill~3~\citep{tao2024maniskill3}. The benchmark
organizes evaluation around the information a policy must retain and the
temporal structure of the interaction. Most tasks present a cue that becomes
unavailable before or during the action it determines. \NumGapNoRetention{}
tasks keep the relevant information visible and serve as reactive controls.
Figure~\ref{fig:teaser} illustrates the cue-to-action structure in two
representative tasks.

The registered variants belong to \NumFamilies{} task families, which in turn
are assigned to \NumMemoryTypes{} memory categories. A typical
memory-dependent episode presents a cue, hides or occludes it, and then opens
the action phase that depends on it. This is not a universal template:
reactive controls retain a visible cue, tracking tasks require following a
moving referent, and \texttt{TimedTransfer} requires acting in a specified
time window.

\subsection{Memory taxonomy and task families}
\label{sec:taxonomy}

The suite covers \NumMemoryTypes{} memory types (Table~\ref{tab:taxonomy}).
These categories describe the content
and operations involved in memory: retaining an identity or location,
remembering a set or ordered sequence, tracking changes, or maintaining
pending intentions and completed actions. Within these categories,
\NumFamilies{} task families share an underlying environment design.
Difficulty variants change factors such as the number of distractors, the
number of items to retain, and the length of a demonstrated sequence.
For example, \texttt{RememberColor} varies the candidate set, while its
timing variants vary the sampled interval between cue and action. These
semantic and temporal changes are useful controls, but the benchmark does not
form a fully crossed factorial design.
Category assignments follow the released manifest.
Section~\ref{sec:limitations} discusses an inconsistency between one family's
label and its success condition.

\definecolor{mtObject}{HTML}{4269D0}
\definecolor{mtSpatial}{HTML}{EFB118}
\definecolor{mtCapacity}{HTML}{FF725C}
\definecolor{mtTemporal}{HTML}{6CC5B0}
\definecolor{mtNegative}{HTML}{3CA951}
\definecolor{mtSequential}{HTML}{FF8AB7}
\definecolor{mtProcedural}{HTML}{A463F2}
\definecolor{mtProspective}{HTML}{97BBF5}
\definecolor{mtTracking}{HTML}{9C6B4E}
\definecolor{mtChecklist}{HTML}{9498A0}
\begin{table}[t]
\centering\scriptsize
\setlength{\tabcolsep}{3pt}
\caption{\textbf{Memory-type taxonomy: definitions, task and episode counts, and the realized episode-length distribution in steps (minimum, mean, median, maximum).} Figure~\ref{fig:teaser} shows a rollout from two of these types. Appendix~\ref{sec:appendix-contact-sheet} shows one for every task. Each swatch is that type's color in every figure of this paper.}
\label{tab:taxonomy}
\begin{tabular}{@{}lp{6.15cm}rrrrrr@{}}
\toprule
\textbf{Memory type} & \textbf{Definition} & \textbf{Tasks} & \textbf{Episodes} & \rotatebox{60}{\textbf{Min t}} & \rotatebox{60}{\textbf{Mean t}} & \rotatebox{60}{\textbf{Median t}} & \rotatebox{60}{\textbf{Max t}} \\
\midrule
\textcolor{mtObject}{\rule[-0.05em]{1.1ex}{1.1ex}}\,Object & The cue reveals which object identity (color or shape) is the target. The agent must retain that identity while acting among visually similar distractors. & 18 & 4{,}500 & 10 & 178 & 53 & 599 \\
\textcolor{mtSpatial}{\rule[-0.05em]{1.1ex}{1.1ex}}\,Spatial & The goal is a place rather than an object identity: which container hides the ball, where a moving ball will arrive, a target angle, or where an object began. Where that place is hidden, the agent must recall it rather than read it off the current frame. 10 of the 14 are reactive controls whose cue is never hidden, so recall is demanded only in the other 4. & 14 & 3{,}500 & 10 & 30 & 28 & 90 \\
\textcolor{mtCapacity}{\rule[-0.05em]{1.1ex}{1.1ex}}\,Capacity & The cue presents a set of items to be collected or touched in any order. The agent must retain the full set membership across the interaction, with set size scaling the memory load. & 12 & 3{,}000 & 121 & 431 & 355 & 1199 \\
\textcolor{mtTemporal}{\rule[-0.05em]{1.1ex}{1.1ex}}\,Temporal & The cue is a countable or timed event (e.g., a number of blinks). The agent must retain a count or duration and reproduce it during the action phase. & 12 & 3{,}000 & 57 & 338 & 204 & 1004 \\
\textcolor{mtNegative}{\rule[-0.05em]{1.1ex}{1.1ex}}\,Negative & The cue presents a set of items and the goal is defined by exclusion. The agent must retain the whole set it was shown in order to touch the one later item whose color or shape was not in it. & 9 & 2{,}250 & 10 & 15 & 15 & 28 \\
\textcolor{mtSequential}{\rule[-0.05em]{1.1ex}{1.1ex}}\,Sequential & The cue presents an ordered sequence of items. The agent must retain both the set and the order and reproduce that order during the action phase. & 6 & 1{,}500 & 131 & 489 & 371 & 1199 \\
\textcolor{mtProcedural}{\rule[-0.05em]{1.1ex}{1.1ex}}\,Procedural & The cue demonstrates a motion or shape being traced. The agent must retain a trajectory rather than a static fact and reproduce it during the action phase. & 6 & 1{,}500 & 188 & 460 & 307 & 1480 \\
\textcolor{mtProspective}{\rule[-0.05em]{1.1ex}{1.1ex}}\,Prospective & A brief cue appears part-way through a longer manipulation task and specifies an action to perform only once that task is complete. The agent must hold the pending intention while the manipulation continues to occupy it. & 5 & 1{,}250 & 104 & 513 & 514 & 940 \\
\textcolor{mtTracking}{\rule[-0.05em]{1.1ex}{1.1ex}}\,Tracking & The cue's referents are shuffled after the cue ends. The agent must keep updating where the cued content now is rather than retaining a single static fact. & 4 & 1{,}000 & 29 & 188 & 102 & 513 \\
\textcolor{mtChecklist}{\rule[-0.05em]{1.1ex}{1.1ex}}\,Checklist & The task requires testing several items in an unspecified order. The agent must retain both which items it has already tested and how each test came out in order to report the result set without repeats or omissions. & 4 & 1{,}000 & 351 & 805 & 720 & 1523 \\
\bottomrule
\end{tabular}
\end{table}

\subsection{Temporal structure and horizon splits}
\label{sec:splits}

Configured episode horizons range from \MinHorizon{} to \MaxHorizon{} steps.
The Short, Medium, and Long splits group tasks by these limits
(Table~\ref{tab:splits}), providing a common stratification for training and
evaluation. Several families include longer-horizon variants that extend the
interval between observing a cue and acting on it while preserving the
underlying task semantics. Semantic and temporal parameters therefore offer
distinct ways to vary difficulty, although the release is not a fully
crossed experimental design. Episode horizon also includes observation,
waiting, and motor execution, so it is not itself a measure of retention
(Section~\ref{sec:gap-analysis}).

\begin{table}[t]
\centering\small
\caption{\textbf{Horizon-split composition: task and episode counts, frame totals, realized episode length, and the design horizon range.}}
\label{tab:splits}
\setlength{\tabcolsep}{4pt}
\begin{tabular}{@{}lrrrrrl@{}}
\toprule
\textbf{Split} & \textbf{Tasks} & \textbf{Episodes} & \textbf{Frames} & \textbf{Median length (steps)} & \textbf{Max length (steps)} & \textbf{Horizon range} \\
\midrule
Short & 38 & 9{,}500 & 303{,}527 & 18 & 197 & 25--200 \\
Medium & 30 & 7{,}500 & 2{,}154{,}062 & 272 & 599 & 250--600 \\
Long & 22 & 5{,}500 & 3{,}849{,}364 & 698 & 1{,}523 & 700--2{,}160 \\
\bottomrule
\end{tabular}
\end{table}

\subsection{Extension over MIKASA-Robo}

MIKASA-Robo-VLA extends MIKASA-Robo from a representative VLA subset to a
full-suite language interface. It broadens task and memory coverage, adds
temporal variants, standardizes policy evaluation, and releases demonstrations
in both RLDS and LeRobotDataset v3. The two suites share a memory-first
design, but task-name overlap alone is not evidence that every variant has an
identical environment implementation. Task parameters are meant to be
re-specified: Appendix~\ref{sec:appendix-customization} demonstrates this with
custom variants built to test a different memory update rate.

\subsection{Observations, actions, and language}
\label{sec:obs-action}

All tasks expose a shared policy interface. Visual observations consist of
a static top-down RGB image and a wrist-camera RGB image, each of resolution
$\ImageShape$. A \StateDim{}-dimensional proprioceptive vector contains
end-effector position, roll--pitch--yaw orientation, and gripper state.
Actions comprise \ActionDim{} dimensions for delta end-effector pose and
gripper control, executed using \texttt{pd\_ee\_delta\_pose} at
\SI{\ControlRateHz}{\hertz}.

Language instructions specify the goal and the relevant memory operation.
For example, \texttt{RememberColor} asks the robot to observe a cube's color
and later select a matching cube. Difficulty variants change the distractor
set while retaining the same instruction. The suite uses
\NumDistinctInstructions{} instruction templates across \NumTasks{} tasks.
Most are shared within families. \texttt{TimedTransfer} includes the required
step count, and the Rotate families insert the episode's target angle.

\subsection{Environment interface and evaluation}
\label{sec:wrappers}

The shared interface standardizes camera and proprioceptive observations and,
where required by the collection curriculum, freezes actions during cue
presentation or waiting phases. Optional diagnostic overlays affect rendered
inspection videos rather than policy images, and the released demonstrations
were collected with overlays disabled. The exact factory call and wrapper behavior
are documented in Appendix~\ref{sec:appendix-wrappers}.

The benchmark supplies a common evaluation procedure with fixed episode
seeds, a 50-episode-per-task default, and success credited if the task
condition holds at any step (\texttt{success\_once}). The reference baseline
uses a smaller budget (Section~\ref{sec:baseline}), and known interface
issues are summarized in Section~\ref{sec:limitations}.

\subsection{Rewards and success criteria}
\label{sec:rewards}

Every task provides a dense reward for RL training and an explicit binary
success condition for evaluation. Reward shaping combines reaching and
placement terms with task-specific progress signals, such as the number of
correctly recalled items, and penalties for unnecessary motion or disturbing
protected objects. Terminal success receives a fixed reward. Success is
assessed directly from task conditions --- for example, reaching a goal region,
matching a count, or completing a sequence in order --- rather than by applying
a threshold to the shaped reward.
Appendix~\ref{sec:appendix-families} specifies the reward components and
success conditions for each family.

\section{Datasets}
\label{sec:datasets}

We release demonstrations for every benchmark task to support supervised
VLA fine-tuning and subsequent evaluation. The collection contains
\NumEpisodes{} episodes and \NumFrames{} timesteps, corresponding to
\NumHours{} hours of simulated interaction at
\SI{\ControlRateHz}{\hertz}. Both release formats use the policy interface
specified in Section~\ref{sec:obs-action}.

\subsection{Demonstration collection}
\label{sec:collection}

Each task contributes \EpisodesPerTask{} successful oracle episodes.
Demonstrations are generated by PPO policies trained with dense rewards for
\NumTasksPPO{} tasks and by scripted motion planners for the remaining
\NumTasksMP{}. Both use privileged simulator state to resolve the task, and initial scenes
are sampled through the environment's reset procedure. Simulator seeds are
retained in the RLDS metadata for reproducing initial conditions.

The collected rollouts are rendered through the VLA observation wrappers to
produce camera images and proprioception in the same representation used
for policy evaluation. Only episodes satisfying the task's success condition
are retained. Consequently, the reported oracle success rate of
\MeanOracleSuccess{} describes the filtered dataset, rather than an
unconditional estimate of oracle reliability. The release provides successful
behavior for imitation learning but does not systematically cover failure
recovery or states induced by a learned policy's errors.

\subsection{Release formats}
\label{sec:formats}

The same demonstrations are available in RLDS~\citep{ramos2021rlds} and
LeRobotDataset v3~\citep{cadene2026lerobot}. Both contain language instructions, two RGB camera
streams, proprioception, and actions. The RLDS release follows the TFDS
organization used by Open X-Embodiment~\citep{oxecollaboration2024}, stores
camera frames as PNG images, and retains per-step rewards. The LeRobot
release stores state and action arrays in Parquet and camera streams as
AV1 video. It occupies \LerobotSizeGiB{}\,GiB, compared with
\RldsSizeGiB{}\,GiB for RLDS. The releases differ in both image encoding and
schema, so their size difference should not be attributed to compression alone.

The formats differ in metadata availability: the released LeRobot schema
omits per-step rewards, so reward-dependent training should use RLDS. No
shared episode identifier is provided, so episode indices must be treated as
local to each format, and seeds, rewards, and success metadata should be read
from the corresponding RLDS records rather than joined by index.
Appendix~\ref{sec:datasheet} documents the schemas and metadata limitations.

\subsection{Dataset composition}
\label{sec:aggregate-stats}

\begin{figure}[t]
  \centering
  \includegraphics[width=\linewidth]{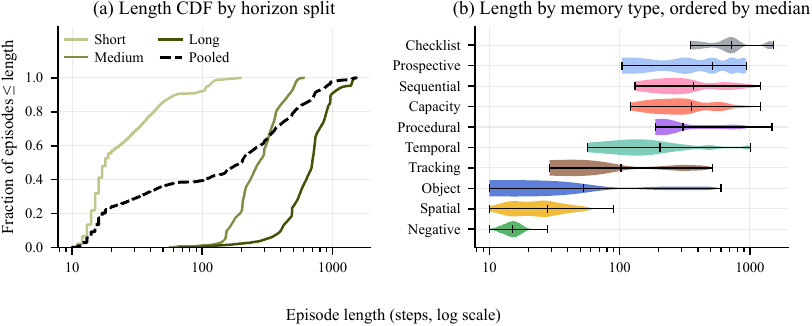}
  \vspace{-8pt}
  \caption{\textbf{Realized episode lengths across \NumEpisodes{}
  demonstrations.} (a) Empirical cumulative distributions for the full release and each
  horizon split, with a logarithmic horizontal axis. (b) Length distributions
  by memory type, ordered by median. Violins show kernel-density estimates.
  Horizontal lines indicate the observed range and ticks mark the median.}
  \label{fig:lengths}
\end{figure}

The fixed number of demonstrations per task balances episode counts, but
longer tasks contribute more timesteps. Figure~\ref{fig:lengths} shows the
realized episode-length distribution by horizon split and memory type.
The horizon splits describe
task duration, rather than prescribing a single train/test partition, and
evaluation must separately specify held-out seeds or tasks.
Table~\ref{tab:splits} summarizes the split-level composition, and
Appendix~\ref{sec:appendix-all-tasks} reports per-task statistics.

\section{Analysis}
\label{sec:analysis}

We first characterize environment-level information gaps, then report a
reference policy experiment, and finally examine associations within its
limited evaluated subset.

\subsection{Information gap and observation context}
\label{sec:gap-analysis}

The \emph{information gap} measures the interval between the last availability
of a task-relevant cue and the start of the phase that requires a
cue-dependent action. We extract this interval from the environment's cue
and blackout phase boundaries, rather than from a policy's realized action
time.
Fixed phase timings yield a constant gap, while timings sampled at reset
yield a distribution, summarized by its median and interquartile range.
Related formulations characterize memory length or complexity through
optimal policies~\citep{ni2023transformers,chen2026rmbench}. Our measure
provides an environment-based description without requiring an optimal
policy estimate.
Per-family descriptions in Appendix~\ref{sec:appendix-families} specify the
underlying task structure.

An environment-determined gap is available for \NumGapDetermined{} tasks.
The remaining tasks comprise \NumGapNoRetention{} reactive controls, whose
cue remains visible, and \NumGapUndetermined{} tasks whose retention interval
depends on reaching a state and cannot be determined from phase timing alone.
Among the determined gaps, \NumTasksAbovePublishedContext{} exceed
\PublishedContextFrames{} steps, the widest fixed observation window
reported for the configurations reviewed in Section~\ref{sec:related},
counted in successive observations at the benchmark's control rate. This
threshold is a reference point, not a statement that all VLA policies use
the same context length, and it does not imply that the tasks below it are
memory-free -- the gap counts only the provably-cue-absent interval, and
every memory-dependent task in the suite requires retention by
construction.

The gap measures an explicit interval of information unavailability, rather
than the total duration for which a successful policy must retain it.
For \NumTasksZeroGap{} tasks, the gap is zero because the dependent action
phase begins immediately when the cue disappears, and recall may still be
needed throughout that phase. The threshold counts therefore omit some
memory demands, including those in \TypesAllGapsZero{} tasks with zero
determined gaps and \TypesNoDeterminedGap{} tasks without determined gaps.
Conversely, a positive gap alone does not establish that a policy must fail:
correlated observations, embodied strategies, or explicit memory may provide
alternative access to the relevant information.

\subsection{Reference baseline}
\label{sec:baseline}

\paragraph{Training and evaluation.}
We fully fine-tune $\pi_{0.5}$~\citep{intelligence2025pi05} from the public
\texttt{pi05\_base} checkpoint on \NumBaselineTasks{} tasks, using
\BaselineTrainEpisodes{} LeRobot episodes comprising
\BaselineTrainFrames{} timesteps. The subset spans
\NumBaselineMemoryTypes{} memory types and \BaselineFramePercent{}\% of the
released timesteps. It follows the available training mixture and is not a
representative sample of the full benchmark. \BaselineUnevaluatedTypes{}
are not evaluated.

The policy receives the current images from both cameras, resized to
$\BaselineUpscale{}\times\BaselineUpscale{}$, and proprioception discretized
into the text prompt. It has no observation history or explicit memory
module. The \ActionDim{} active action dimensions are padded to the
checkpoint's \BaselineActionDimPadded{} dimensions, and predictions are
executed in open-loop chunks of \BaselineChunk{} actions. We evaluate one training run
at checkpoint \BaselineSteps{}, using \BaselineEpisodesPerTask{} episodes
per task with seeds disjoint from the training set. Success follows
\texttt{success\_once}. This evaluation uses fewer episodes and a different
seed range than the released defaults, with the wrapper adjustments of
Section~\ref{sec:limitations}.
Appendix~\ref{sec:appendix-baseline} reports the full configuration.

\begin{table}[!t]
\centering\small
\caption{\textbf{Reference \(\pi_{0.5}\) performance with current images and proprioception but no observation history or explicit memory module, evaluated on 20 episodes per task.} The aggregate is the unweighted mean across tasks.}
\label{tab:baseline}
\setlength{\tabcolsep}{5pt}
\begin{tabular}{lllr}
\toprule
\textbf{Task} & \textbf{Memory type} & \textbf{Horizon} & \textbf{Success} \\
\midrule
\texttt{BunchOfColors3-Long-VLA-v0} & Capacity & Long & 0.000 \\
\texttt{ChainOfColors3-Long-VLA-v0} & Sequential & Long & 0.000 \\
\texttt{GatherAndRecall1-VLA-v0} & Prospective & Short & 0.350 \\
\texttt{GatherAndRecall3-VLA-v0} & Prospective & Medium & 0.150 \\
\texttt{InterceptGrabMedium-VLA-v0} & Spatial & Short & 0.050 \\
\texttt{InterceptMedium-VLA-v0} & Spatial & Short & 0.400 \\
\texttt{RememberColor5-Long-VLA-v0} & Object & Medium & 0.300 \\
\texttt{RememberColor5-VLA-v0} & Object & Short & 0.250 \\
\texttt{RememberShape5-VLA-v0} & Object & Short & 0.050 \\
\texttt{RememberShapeAndColor3x2-VLA-v0} & Object & Short & 0.200 \\
\texttt{ShellGameShuffleColorLampTouch-Long-VLA-v0} & Tracking & Medium & 0.350 \\
\texttt{TakeItBack-VLA-v0} & Spatial & Short & 0.500 \\
\texttt{TraceShapeEasy-VLA-v0} & Procedural & Medium & 0.300 \\
\texttt{TraceShapeSeqEasy-VLA-v0} & Procedural & Long & 0.050 \\
\midrule
Mean (14 tasks) & & & 0.211 \\
\bottomrule
\end{tabular}
\end{table}

\paragraph{Overall performance and chance references.}
Mean task success is $\BaselineMean{}\pm\BaselineMeanSE{}$
(Table~\ref{tab:baseline}), with the aggregate standard error computed across
tasks. Per-task binomial standard errors range from
\BaselineSEMin{} to \BaselineSEMax{} for nonzero success rates.
Zero-success rows instead report exact one-sided 95\% upper bounds.
The small evaluation budget limits the precision of individual task
comparisons.

Uniformly random actions achieve \RandomFloorSuccesses{} success in
\RandomFloorEpisodes{} episodes across all \RandomFloorTasks{} tasks.
This establishes a low floor for unstructured control, but does not exclude
success through competent manipulation combined with guessing. For the
\NumChanceFloorTasks{} evaluated tasks with a discrete choice among $k$
equally likely candidates, a cue-independent uniform choice has success
probability $1/k$. Across these tasks, the baseline obtains
\ChanceFloorPooledSuccesses{} successes in \ChanceFloorPooledEpisodes{}
episodes, compared with \ChanceFloorPooledExpected{} expected successes and
a standard deviation of \ChanceFloorPooledSD{} under this reference.
No task exceeds its choice reference by more than one reported standard
error. These results provide no clear evidence of cue-dependent selection
on this subset. The analytic reference assumes successful execution after
a candidate is selected and is not a learned-policy baseline.

\subsection{Memory architecture comparison}
\label{sec:arch-comparison}

$\mu$VLA~\citep{cherepanov2026uvla}, an explicit-memory extension of the
benchmark's task family, reports the comparison this baseline alone cannot:
OpenVLA-OFT with and without a 64-token recurrent memory module raises mean
success from 0.42 to 0.84 on five trained tasks and from 0.07 to 0.23 on 18
held-out ones~\citep{cherepanov2026uvla} (per-task breakdown in
Appendix~\ref{sec:appendix-external}). A separate execution-regime analysis
there shows the same open-loop chunking this baseline uses shifts per-task
success by up to \ExecRegimeMaxDelta{}, a genuine confound on the Long-split
result below.

\subsection{Performance across memory demands and horizons}
\label{sec:failure-analysis}

All permutation tests below use tasks as the unit of analysis, enumerate
all assignments with the observed group sizes, and describe associations
within the evaluated subset.

\paragraph{Information gaps.}
The \NumBaselineGapDetermined{} tasks with determined gaps have mean success
\BaselineDeterminedGapMean{}, compared with
\BaselineNoDeterminedGapMean{} for the other
\NumBaselineNoDeterminedGap{} tasks (two-sided permutation
$p=\BaselineGapGroupP{}$). The latter group includes
\NumBaselineGapNoRetention{} reactive controls and
\NumBaselineGapUndetermined{} tasks with undetermined retention intervals,
so this comparison does not contrast memory-dependent and memory-free tasks.
Excluding the \BaselineLongTasks{} Long-split tasks reduces the difference
from \BaselineGapGroupDiff{} to \BaselineGapGroupDiffNoLong{}
($p=\BaselineGapGroupPNoLong{}$). Among tasks with determined gaps,
Spearman's correlation between median gap and success is
$\rho=\BaselineGapSpearman{}$ (two-sided permutation
$p=\BaselineGapSpearmanP{}$). Neither comparison establishes an association
at this sample size, and the small subset also limits the ability to detect
one.

\paragraph{Episode horizons.}
The \BaselineLongTasks{} evaluated Long-split tasks achieve
\BaselineLongSuccesses{} success in \BaselineLongEpisodes{} episodes
(\BaselineLongMean{}), compared with \BaselineNonLongSuccesses{} in
\BaselineNonLongEpisodes{} (\BaselineNonLongMean{}) for the remaining
\BaselineNonLongTasks{} tasks. A one-sided task-level permutation test gives
$p=\BaselineHorizonP{}$. Split membership follows Table~\ref{tab:splits}'s
horizon bands, not the \texttt{-Long-} name suffix.

This contrast identifies a concentration of failures in the longest-horizon
tasks, but its cause is unresolved: those tasks include the subset's only
Capacity and Sequential tasks and one Procedural task, coupling duration
with the need to reproduce sets, orders, or trajectories, and Section~
\ref{sec:arch-comparison} above already documents open-loop chunking as an
independent confound on longer rollouts.

\section{Limitations and Future Work}
\label{sec:limitations}

The reference results cover one training run and one checkpoint on
\NumBaselineTasks{} of \NumTasks{} tasks, and a separate architecture
comparison from a related project, not a matched run, does not show which
tasks discriminate under identical training. The baseline's
\BaselineChunk{}-action open-loop chunks and training on successful
demonstrations with limited recovery mean a long-task failure can reflect
memory, motor error, or distribution shift. Phase timing captures explicit
information gaps, not all policy-dependent retention demands, and does not
rule out shortcuts. The released gaps are not a dense delay sweep, and
\texttt{TimedTransfer} measures elapsed-time tracking, not recall of a hidden
visual cue. All environments and demonstrations are simulated, collected
with privileged-state PPO policies or motion planners, so transfer to
physical robots is unmeasured and the release contains no human
demonstrations. The instruction set is limited to
\NumTasksSharedInstruction{} tasks sharing an instruction and no held-out
paraphrases, so results concern task execution, not linguistic
generalization.
\section{Conclusion}
\label{sec:conclusion}

MIKASA-Robo-VLA extends MIKASA-Robo from a reinforcement-learning
suite into a language-conditioned benchmark for memory in
vision-language-action policies. Every one of its \NumTasks{} tasks carries
an instruction, spans \NumMemoryTypes{} memory types, and falls into one of
\NumHorizonSplits{} horizon splits, with step horizons from \MinHorizon{} to
\MaxHorizon{}. We release \NumEpisodes{} oracle trajectories,
\NumFrames{} frames total, in both RLDS and LeRobotDataset v3, and for
\NumTasksAbovePublishedContext{} tasks the environment's own phase timings
place the median information gap beyond the widest fixed-context window we
survey. Every memory-dependent task in the suite requires retention by
construction, regardless of the measured gap length. A
reference \(\pi_{0.5}\) baseline with current images and proprioception but
no observation history or explicit memory module reaches
$\BaselineMean{} \pm \BaselineMeanSE{}$ mean success on \NumBaselineTasks{}
tasks. That number describes one model under one training and evaluation
protocol, not a ceiling on what memory-free policies can do on this suite,
and several confounds keep it from separating horizon length from retention
demand. The benchmark's contribution is the standardized task suite, the
dataset, and the diagnostic apparatus around it (the taxonomy, the gap
metric, the chance floors) rather than this single evaluated point. We see
MIKASA-Robo-VLA as infrastructure for the question its title asks, not an
answer to it. Its \NumFamilies{} task families and released oracle
trajectories give later work a fixed target for training and comparing
memory-augmented VLA policies against a documented baseline, with the same
per-family reward and success definitions available to every method that
evaluates against it.

\bibliographystyle{plainnat}
\bibliography{refs}

\appendix

\section*{Appendix Overview}

\renewcommand*{\theHtable}{appendix.\arabic{table}}

\begin{itemize}
\itemsep0pt
\item Appendix~\ref{sec:datasheet} (p.~\pageref{sec:datasheet}): Datasheet
\item Appendix~\ref{sec:appendix-schemas} (p.~\pageref{sec:appendix-schemas}): Dataset Schemas
\item Appendix~\ref{sec:appendix-obs-action} (p.~\pageref{sec:appendix-obs-action}): Observation and Action Details
\item Appendix~\ref{sec:appendix-eval-protocol} (p.~\pageref{sec:appendix-eval-protocol}): Evaluation Protocol and Release Issues
\item Appendix~\ref{sec:appendix-wrappers} (p.~\pageref{sec:appendix-wrappers}): Wrapper Cookbook
\item Appendix~\ref{sec:appendix-families} (p.~\pageref{sec:appendix-families}): Per-Family Task Details
\item Appendix~\ref{sec:appendix-all-tasks} (p.~\pageref{sec:appendix-all-tasks}): Full Per-Task Statistics
\item Appendix~\ref{sec:appendix-baseline} (p.~\pageref{sec:appendix-baseline}): Reference Baseline: Full Configuration
\item Appendix~\ref{sec:appendix-comparison} (p.~\pageref{sec:appendix-comparison}): Comparison to Existing Benchmarks
\item Appendix~\ref{sec:appendix-external} (p.~\pageref{sec:appendix-external}): Supplementary External Evaluations
\item Appendix~\ref{sec:appendix-contact-sheet} (p.~\pageref{sec:appendix-contact-sheet}): Rollout Contact Sheet
\end{itemize}

\section{Datasheet}
\label{sec:datasheet}

Following the framework of \citet{gebru2021datasheets}.

\subsection*{Motivation}

\textbf{For what purpose was the dataset created?} This dataset provides
training data and environments for evaluating language-conditioned
manipulation tasks with documented memory demands. It supports comparisons of
policies that use recent observations, recurrent state, or external memory
across task semantics and retention intervals.

\subsection*{Composition}

\textbf{What do the instances represent?} Each episode is one simulated
manipulation rollout: two RGB camera streams, a \StateDim{}-dimensional
proprioceptive state, and a \ActionDim{}-dimensional action per step, plus the
language instruction for that task (Section~\ref{sec:obs-action}).

\textbf{Do the released frames contain rendered debug text?} No. Overlay
wrappers modify the output of \texttt{render}, not the observation returned by
\texttt{step}, and collection used \texttt{include\_overlays=False}. Debug
text appears only in the rendered rollout videos used for this paper's
filmstrip figures
(Appendix~\ref{sec:appendix-families}).

\textbf{How many instances are there?} \NumTasks{} tasks, \EpisodesPerTask{}
episodes each, \NumEpisodes{} episodes and \NumFrames{} frames in total
(\NumHours{} hours of control at \SI{\ControlRateHz}{\hertz}).

\textbf{Does the dataset contain all possible instances, or a sample?}
Every task's release is a fixed sample of \EpisodesPerTask{} oracle
rollouts, not an exhaustive enumeration of the task's state space -- see
\emph{Collection Process} below for how each episode's initial state is
drawn.

\textbf{Is there a label or target associated with each instance?} Each RLDS
step stores a normalized dense reward, and the RLDS sidecar stores per-episode
\texttt{success\_once}, reward sums, and simulator seeds. LeRobot's native
schema stores neither reward nor success. Every released episode satisfies
\texttt{success\_once}. Appendix~\ref{sec:appendix-schemas} describes the
format-specific fields and their alignment.

\textbf{Are there recommended data splits?} The benchmark's own Short/
Medium/Long horizon splits (Table~\ref{tab:splits}) and its \NumMemoryTypes{} memory
types (Table~\ref{tab:taxonomy}) are the natural stratification for
held-out evaluation subsets. This release does not impose a single
train/test split, since different consumers will want to hold out
different subsets of the \NumTasks{} tasks.

\textbf{Does the dataset relate to people?} No. Every frame is rendered from a
simulated scene holding a robot arm and rigid props, and the demonstrations
were generated by policies and motion planners rather than by human
teleoperation (Section~\ref{sec:collection}), so the release contains no human
subjects, no recorded operators, and no personally identifiable information.

\subsection*{Collection Process}

\textbf{How was the data associated with each instance acquired?} Every
episode is a full simulator rollout, not a derived or inferred value. On
\NumTasksPPO{} tasks a PPO policy trained against the task's dense
reward is the oracle, and the remaining \NumTasksMP{} use a scripted motion
planner, for tasks whose action sequence is fully determined once the cue
is resolved (Section~\ref{sec:collection}).

\textbf{What mechanisms were used to collect the data?} Both oracle types
run inside the same ManiSkill/Gymnasium simulator the released observation
and action spaces are defined for. Video and proprioception are re-rendered
from each rollout after collection using the VLA-facing observation wrapper
stack (Section~\ref{sec:wrappers}), so the released data matches exactly
what a policy would receive at evaluation time, not the oracle's privileged
view during collection.

\textbf{Over what timeframe was the data collected?} The collection dates are
not recorded in the release metadata. Each released episode retains a
simulator seed in \texttt{episode\_seeds}. Recreating an initial state also
requires matching simulator software, assets, and environment configuration.
The PPO checkpoints used for \NumTasksPPO{} tasks are not part of the release,
so those rollouts cannot be replayed from the published data alone.

\textbf{Warning: the two release branches are ordered differently, and
episode index is not a join key.}
The RLDS and LeRobot branches contain the same episodes for every task, but
their order can differ. The converter does not record a cross-format episode
mapping. Episode length cannot supply an unambiguous mapping: all
\NumTasksLengthsAmbiguous{} tasks contain at least two episodes with the same
length.

This matters because of where the per-episode metadata lives. Each task's
\texttt{source\_rlds\_metadata.json} -- carrying that task's episode lengths,
\texttt{success\_once} flags, reward sums, and simulator seeds -- sits
\emph{inside} the LeRobot tree, but its arrays are indexed in \textbf{RLDS}
order. The LeRobot branch's own \texttt{meta/info.json} carries no reward, no
success flag, and no seed, so that file is the only source for them. A reader
who trains on the LeRobot branch and joins this metadata by episode index can
assign a different episode's reward, success flag, or seed.

The release provides no unambiguous cross-format join key. The simulator seed
appears only in \texttt{source\_rlds\_metadata.json}, in RLDS order. Read an
episode's reward, success flag, or seed from the RLDS branch, where index and
metadata agree, and treat the LeRobot episode index as branch-local. This
limitation does not affect any number reported in this paper:
every statistic here is a count, sum, or median computed from
\texttt{source\_rlds\_metadata.json} alone, whose own arrays are mutually
aligned, and no analysis in this paper joins across the two branches.

\subsection*{Preprocessing, Cleaning, and Labeling}

\textbf{Was any preprocessing done?} Oracle rollouts are filtered to
episodes that reach \texttt{success\_once}. Both release formats apply the
same observation wrapper stack (Section~\ref{sec:wrappers}) before writing
video and proprioception, so raw simulator state is never released
directly. No other filtering, augmentation, or relabeling is applied.

\textbf{Is the raw data also available?} No. Only the wrapped,
VLA-facing observation is released. The underlying ManiSkill/SAPIEN scene
state is not retained per episode.

\subsection*{Uses}

\textbf{What tasks has the dataset been used for?} Training and evaluating
the reference \(\pi_{0.5}\) baseline reported in Section~\ref{sec:baseline}
of this paper, on a \NumBaselineTasks{}-task subset.

\textbf{Is there anything about the composition or collection that might
impact future uses?} The oracle demonstrations come from PPO policies and
motion planners with privileged simulator access, not human teleoperation
(Section~\ref{sec:limitations}). They are a controlled source of training
data for exactly these tasks, not a proxy for the demonstration
distributions a real deployment would collect. Every episode succeeds by
construction (Section~\ref{sec:collection}), so the release contains no
failure trajectories for a policy to learn to avoid.

\textbf{Are there tasks for which the dataset should not be used?} It should
not be used to argue that a policy is ready for hardware. Every episode is
simulated and every one succeeds, so the release holds no failure and no
recovery for a policy to learn from, and no number in this paper was measured
against a physical robot. A score here is also not by itself evidence of
retention: Section~\ref{sec:gap-analysis} measures what a task demands, not
what a policy does.

\subsection*{Distribution}

\textbf{Will the dataset be distributed to third parties?} It already is, in
both formats (\dataurl{}), alongside the environment suite and the wrapper
stack that produced it (\repourl{}). Both predate this submission.

\textbf{Under what license?} The release distributes the benchmark code and
trajectories under the MIT License. Users should consult the licenses of
upstream software and assets separately.

\subsection*{Maintenance}

\textbf{Will the dataset be updated?} Corrections to the manifest or
extraction scripts, if found after release, will be documented in the
repository's changelog rather than silently changing released files.

\section{Dataset Schemas}
\label{sec:appendix-schemas}

Section~\ref{sec:formats} names the two release formats and what distinguishes
them. This section gives their per-key schemas, so that a loader can be written
against this release rather than against RLDS and LeRobot in general. Every
literal below is quoted from a named source file or from the released schema
itself. None is a measurement of the data.

\textbf{Actions are normalized, but proprioception is not.} The control mode is
\texttt{pd\_ee\_delta\_pose} (Section~\ref{sec:obs-action}), which resolves to
ManiSkill's \texttt{PDEEPoseControllerConfig} for the arm and
\texttt{PDJointPosMimicControllerConfig} for the gripper. Both set
\texttt{normalize\_action=True} by default, and the Panda's registered
configuration does not override it, so the environment's action space is
$[-1, 1]^{\ActionDim{}}$ and the controller rescales into physical units
internally. The stored actions are the normalized ones. Reading channel by
channel, against the bounds the Panda configuration sets:

\begin{itemize}
  \item Channels 1--3, delta end-effector position:
  \texttt{pos\_lower=-0.1}, \texttt{pos\_upper=0.1}. A stored $+1$ commands
  \SI{0.1}{\metre} of translation on that axis for that step.
  \item Channels 4--6, delta end-effector orientation:
  \texttt{rot\_lower=-0.1}, \texttt{rot\_upper=0.1}. These three are not
  rescaled the way the position channels are. The controller clips them by
  their joint norm and multiplies by \texttt{rot\_lower} alone.
  \texttt{rot\_upper} sets the declared action-space bound and takes no part
  in the scaling. A stored $+1$ therefore commands $-\SI{0.1}{\radian}$. The
  magnitude a reader would infer from the position channels is right and the
  sign is backwards. Replaying stored actions through the same controller is
  unaffected, since the convention is self-consistent, but converting them to
  physical units is not.
  \item Channel \ActionDim{}, gripper: \texttt{lower=-0.01},
  \texttt{upper=0.04}, and \texttt{use\_delta} is \texttt{False} on this
  controller. Unlike the six above it, this channel is an absolute target
  finger position, not a delta, and its bounds are not symmetric about zero.
  A stored $0$ commands the midpoint of that range, not an unchanged gripper.
\end{itemize}

\noindent The \StateDim{}-dimensional proprioceptive vector is stored raw, in
the simulator's own units, and is not normalized or centered:
end-effector $xyz$ in metres, roll-pitch-yaw in radians as
\texttt{ConvertJointsToEEFXyzRpyGripperWrapper} derives them from the
end-effector quaternion, and a gripper channel holding the sum of the two
Panda finger joint positions, also in metres. A loader that rescales state to
match action, or rescales action to match state, produces a policy that trains
without error and acts on the wrong scale.

\textbf{LeRobotDataset v3 branch.} Each task directory holds
\texttt{meta/info.json}, columnar Parquet under \texttt{data/}, and one
AV1-encoded MP4 per camera under \texttt{videos/}. The metadata declares
\texttt{codebase\_version} \texttt{v3.0} and \texttt{robot\_type}
\texttt{mikasa\_robo}. Table~\ref{tab:schema-lerobot} lists the feature
schema, which is identical across all \NumTasks{} tasks.

\begin{table}[h]
\centering\small
\caption{\textbf{LeRobotDataset v3 feature schema, from each task's
\texttt{meta/info.json}.} Video frames decode to HWC \texttt{uint8}. The
declared shape is the frame shape, not the per-row shape.}
\label{tab:schema-lerobot}
\begin{tabular}{@{}llp{6.4cm}@{}}
\toprule
\textbf{Key} & \textbf{Type} & \textbf{Notes} \\
\midrule
\texttt{observation.images.top} & video & $\ImageShape$, AV1 in
\texttt{yuv420p}, \SI{\ControlRateHz}{\hertz}, static camera \\
\texttt{observation.images.wrist} & video & as above, wrist-mounted camera \\
\texttt{observation.state} & \texttt{float32}\,[\StateDim{}] & physical units,
not normalized \\
\texttt{action} & \texttt{float32}\,[\ActionDim{}] & normalized to $[-1, 1]$ \\
\texttt{timestamp} & \texttt{float32}\,[1] & seconds from episode start \\
\texttt{frame\_index} & \texttt{int64}\,[1] & step index within the episode \\
\texttt{episode\_index} & \texttt{int64}\,[1] & branch-local, not a join key
across branches (Appendix~\ref{sec:datasheet}) \\
\texttt{index} & \texttt{int64}\,[1] & row index within the task \\
\texttt{task\_index} & \texttt{int64}\,[1] & resolves against
\texttt{meta/tasks.parquet}, where the language instruction lives \\
\bottomrule
\end{tabular}
\end{table}

There is no reward, no success flag, and no simulator seed anywhere in this
branch's own metadata. All three come from the \texttt{source\_rlds\_metadata.json}
file that sits in the same directory but is indexed in RLDS order, which is the
alignment hazard the datasheet documents (Appendix~\ref{sec:datasheet}).

\textbf{RLDS branch.} Each task directory holds a single TFDS version
directory, \texttt{1.0.0}, with \texttt{dataset\_info.json},
\texttt{features.json}, and sharded TFRecords. The top-level
\texttt{FeaturesDict} has two members: a \texttt{steps} sequence, whose
per-step schema is Table~\ref{tab:schema-rlds}, and an
\texttt{episode\_metadata} dict whose only field is \texttt{file\_path}, a
\texttt{Text} pointing at the source file the episode was converted from.
Nothing else is carried per episode. The seed and the success flag are absent
from the TFDS record as well. They come from the sidecar this branch writes
beside \texttt{features.json} as \texttt{metadata.json}, which is the same file,
byte for byte, that the LeRobot branch ships in each task directory under the
name \texttt{source\_rlds\_metadata.json}. The schema below is
read off one task's \texttt{features.json}. The \NumTasks{} files share
\RldsSchemaDistinctDigests{} \RldsSchemaDigestAlgorithm{} digest,
\RldsSchemaDigest{}, so it describes the whole branch.

\begin{table}[h]
\centering\small
\caption{\textbf{RLDS per-step schema, from each task's
\texttt{features.json}.} The Notes column gives the corresponding LeRobot
key where one exists.}
\label{tab:schema-rlds}
\begin{tabular}{@{}llp{6.2cm}@{}}
\toprule
\textbf{Key under \texttt{steps}} & \textbf{Type} & \textbf{Notes} \\
\midrule
\texttt{observation/image} & \texttt{uint8} $\ImageShape$ & PNG-encoded,
$\rightarrow$ \texttt{observation.images.top} \\
\texttt{observation/wrist\_image} & \texttt{uint8} $\ImageShape$ & PNG-encoded,
$\rightarrow$ \texttt{observation.images.wrist} \\
\texttt{observation/proprio} & \texttt{float32}\,[\StateDim{}] &
$\rightarrow$ \texttt{observation.state} \\
\texttt{action} & \texttt{float32}\,[\ActionDim{}] & normalized to $[-1, 1]$ \\
\texttt{reward} & \texttt{float32} & no LeRobot counterpart, see below \\
\texttt{discount} & \texttt{float32} & written as the constant \texttt{1.0} on
every step, so it carries nothing \\
\texttt{is\_first}, \texttt{is\_last} & \texttt{bool} & episode boundaries \\
\texttt{is\_terminal} & \texttt{bool} & the environment's \texttt{done} flag \\
\texttt{language\_instruction} & \texttt{Text} & repeated per step,
$\rightarrow$ \texttt{task\_index} plus \texttt{meta/tasks.parquet} \\
\bottomrule
\end{tabular}
\end{table}

\textbf{Which reward is stored.} The \texttt{reward} field holds each
environment's \texttt{compute\_normalized\_dense\_reward}, not its
\texttt{compute\_dense\_reward}. Both dataset collectors construct the
environment with \texttt{reward\_mode="normalized\_dense"}
(Appendix~\ref{sec:appendix-wrappers} shows the same call), which routes
ManiSkill's \texttt{get\_reward} to the normalized method, and the RLDS builder
writes that per-step value through unchanged. Each environment implements the
normalized method as its dense reward divided by its own maximum-reward
constant, and that constant differs between families. The stored values are
therefore on a common per-step scale across tasks, and are a fixed rescaling of
the reward terms the per-family summaries describe
(Appendix~\ref{sec:appendix-families}). Recovering the unscaled dense reward
means multiplying by the environment's own constant.

Two things in the released schema will mislead a reader who trusts the
documentation strings over the values. The \texttt{reward} field's own
description reads ``Reward if provided, 1 on final step for demos'', and
\texttt{is\_terminal}'s reads ``True for demos''. Both are inherited verbatim
from the Open X-Embodiment builder template and neither describes what the
builder writes. The stored rewards are dense and per-step, as the episode
reward sums in the sidecar metadata show.

\textbf{The branches were checked to agree.} Section~\ref{sec:formats}
states that both branches hold the same trajectories. Each task's LeRobot
\texttt{total\_frames}, written by the converter, equals the sum of that task's
RLDS-order \texttt{episode\_lengths} on \NumTasks{} of \NumTasks{} tasks, for
\NumFrames{} frames each way. The camera keys, image shape, control rate, and
state and action dimensions are the same value on every task. Both checks run
in this paper's test suite rather than having been done once by hand. What they
constrain is totals and schema. A per-episode correspondence does not follow
from them, and the datasheet's warning about episode order stands
(Appendix~\ref{sec:datasheet}).

\section{Observation and Action Space Details}
\label{sec:appendix-obs-action}

Section~\ref{sec:obs-action} gives the observation and action space at the
level a reader needs to understand the benchmark. This appendix gives the
level a reader needs to reproduce it.

\textbf{Observation dictionary.} After the wrapper stack
(Section~\ref{sec:wrappers}) is applied, every task's observation is a
dictionary with four keys, not the two a policy needs. Two are the intended
policy input: the two stacked camera images (top and wrist, each
$\ImageShape$ uint8) and the \StateDim{}-dimensional
end-effector state (position, roll-pitch-yaw orientation, and a scalar
gripper state, derived from the robot's joint state). The
other two are diagnostic channels carrying privileged state, one of which
was observed holding a real value on all \NumCueKeyTasks{} canonical tasks
when each was reset and stepped. They
are injected by the innermost wrapper
in the stack, which is applied unconditionally to every task and cannot be
switched off. A task that defines neither gets the same sentinel value in
both slots (the numeral 4242424242, coincidentally the same literal the
evaluation module uses as its start seed, Appendix~\ref{sec:appendix-eval-protocol})
rather than having the keys omitted.

The released episodes carry neither key -- the extraction path writes
only the two camera images, the proprioceptive state, and the action -- so
this concerns live evaluation and any pipeline that steps an environment
directly.

\textbf{The language instruction is not in the observation.} It is returned
alongside the reward, success flag, and other per-step information rather
than as part of the observation dictionary. Each task carries one sentence,
shared with the other variants of its family (Section~\ref{sec:obs-action}) and
identical across that task's episodes except in the two Rotate families, whose
instruction names that episode's target angle.
Those two families are also the only ones whose field is a list rather than a
string: they format one sentence per parallel environment, where every other
task returns the single string its class declares. A loader written against
the common case tokenizes a list on them. One written against them indexes
the first element and gets a single character everywhere else. Beyond the four observation
keys, no other simulator state -- object poses, joint torques, contact
forces -- is exposed. Oracle policies see this information during data
collection, but the released episodes are re-rendered from the same rollouts
using the VLA-facing wrapper stack, so training data matches exactly what a
policy receives at evaluation time.

\textbf{Warning: one diagnostic channel leaks the answer on
\NumCueLeakConcealed{} tasks, and a policy must drop it.}
The suite's own reinforcement-learning training wrapper drops both diagnostic
keys on its image-observation path, before the observation reaches a policy.
The VLA wrapper stack has no equivalent step: the innermost wrapper injects
the two keys and nothing downstream removes them
(Appendix~\ref{sec:appendix-wrappers}). A task that assigns the affected
channel a real value assigns it once, at reset, and echoes
it unchanged with nothing gating it on elapsed time, so it survives
into the observation a policy sees at every step, including the steps after the
cue is meant to be hidden. On \NumCueLeakTasks{} canonical ids it was read
holding a real value at both reset and episode end -- the target color, the
shape sequence, or the goal angle:

\begin{itemize}
  \item \texttt{RotateLenientPos}, \texttt{RotateLenientPosNeg},
  \texttt{RotateStrictPos}, \texttt{RotateStrictPosNeg}
  \item \texttt{ShellGameColorLampTouch},
  \texttt{ShellGameShuffleColorLampTouch},
  \texttt{ShellGameShuffleColorLampTouch-Long}
  \item \texttt{TraceShapeSeqEasy}, \texttt{TraceShapeSeqMedium},
  \texttt{TraceShapeSeqHard}
\end{itemize}

\noindent (each with the \texttt{-VLA-v0} suffix). The two groups are not the
same problem. On the \NumCueLeakReactive{} \texttt{Rotate*} ids the cue is
visible \emph{by design}: those tasks regenerate the target angle into the
language instruction at every step, the angle is therefore already available
to the policy through a channel it is supposed to read, and this paper classes
all \NumCueLeakReactive{} as reactive controls that demand no retention
(Section~\ref{sec:gap-analysis}). Nothing is concealed there for
that diagnostic channel to reveal.
The instruction rounds the angle to whole degrees and the diagnostic channel does
not, so the channel is the more precise of the two. It is not the more
useful one: these tasks accept any final angle within \SI{0.1}{\radian} of the
target, an order of magnitude past the half-degree the rounding can cost.

On the remaining \NumCueLeakConcealed{} tasks listed
above, the cue is supposed to be hidden
after the cue phase, and these tasks are counted in this paper's retention
statistics on that basis. That the diagnostic channel still carries the target
color or the shape sequence on them is unintended, and the released
wrapper stack does not filter these two keys out by default. A policy
that flattens or concatenates the observation dictionary without filtering
receives the answer for free on those \NumCueLeakConcealed{} tasks, and any
success rate so obtained measures nothing about retention. Drop both
diagnostic keys explicitly before the observation reaches a policy
(Appendix~\ref{sec:appendix-wrappers} describes how). Results on those
\NumCueLeakConcealed{} ids obtained without that filter should be discarded rather than
reinterpreted: the shortcut is available at every step and leaves no trace in
the score, so there is nothing in the number to correct.

This paper's own reference baseline is unaffected. It does not consume the
observation dictionary at all: its prompt is assembled from a single
per-dataset string, and its harness reads only the camera images and the
discretized proprioceptive vector (Appendix~\ref{sec:appendix-baseline}).
Neither diagnostic channel ever reaches it.

\textbf{Action space.} Actions are \ActionDim{}-dimensional end-effector commands
($\Delta xyz$, $\Delta$roll-pitch-yaw, gripper) under delta end-effector pose
control, issued at \SI{\ControlRateHz}{\hertz} -- the same \StateDim{} dimensions as
the proprioceptive state. Not every channel is a delta. The gripper channel is
an absolute mimic-joint target over the range
$[\GripperRangeLower{}, \GripperRangeUpper{}]$~\si{\metre}, which is
not symmetric about zero, so the zero vector does not mean ``hold position'': after
the normalization to $[-1,1]$ that both channels share, a stored zero commands the
midpoint of that range and opens the fingers part-way. A policy that means ``change
nothing'' has to predict the current gripper width, not a zero.

\textbf{Coordinate conventions.} Positions and orientations are in the
robot base frame, matching ManiSkill's default convention for the
underlying simulator. No additional coordinate transform is applied by the
VLA wrapper stack beyond the joints-to-end-effector conversion itself.

\section{Evaluation Protocol}
\label{sec:appendix-eval-protocol}

Section~\ref{sec:baseline} describes the reference baseline's evaluation in
prose. The protocol it describes is not itself prose: the wrapper stack, the
seeding, the episode count, and the success criterion are all fixed by the
benchmark suite's evaluation module, which is released with the code.
Reproducing a number from this benchmark means running that module, not
reimplementing a description of it.

That module is a library rather than a runnable program: it supplies no
command-line interface of its own. A separate example script, released
alongside it, supplies the command line and a placeholder policy adapter to
replace with one's own. That script exposes options to choose which split
and task to run, to override the start seed and episode count quoted below,
to choose the simulator backend, to save videos, to choose an output
directory, and to resume a partially completed run.

The first part of this appendix records what the released code fixes, so a
reader can check the description against it rather than trust it. The second
records where the reference baseline's own evaluation departed from the
released protocol, because the baseline was scored by the source training
project's evaluation code and not by this one. The third records the chance
floor Section~\ref{sec:baseline} reads that baseline against: what a policy
that never looks at the observation scores on every task.

\subsection{What the released evaluation code fixes}

\textbf{Wrapper stack.} Every environment is constructed through one call to
the released wrapper-construction function before any episode is stepped,
with the same overlay setting (disabled) that produced the released episodes
(Appendix~\ref{sec:appendix-wrappers}). A run
therefore sees exactly the observation and action space of
Appendix~\ref{sec:appendix-obs-action}, including the two diagnostic keys
described there: there is no evaluation-only wrapper and no evaluation-only
observation key.

\textbf{Debug overlays.} Rendering the debug overlays and saving videos are
both disabled by default in the evaluation module's configuration, and the
overlay wrappers are attached only when one of the two is switched on. Asking
for videos therefore turns the overlays on as a side effect, so a run recorded
for inspection is a visibly different configuration from the default one a
reported number comes from. Note that this default is the opposite of the
released wrapper-construction function's own default
(Appendix~\ref{sec:appendix-wrappers}). The evaluation module's default is
the one the release was collected under.

\textbf{Seeding and episode count.} Both are fixed by the evaluation module:
a start seed of \texttt{4242424242} and 50 episodes per task, quoted here in
full so that a reproduction can be checked against them without reading the
source.

A task's $k$-th evaluation episode therefore carries the same seed
on every run rather than a seed drawn afresh, and one episode count applies to
every task in a run rather than a per-task choice. Because the simulator is
deterministic given a seed (Appendix~\ref{sec:datasheet}), an independent
group running the same start seed reproduces the same initial states, not
merely the same distribution over them. The episode count is independent of
the \EpisodesPerTask{} episodes per task the dataset ships, which come from
the collectors rather than from this evaluation module. The start-seed value
is also reused elsewhere in the source as the wrapper stack's missing-value
sentinel (Appendix~\ref{sec:appendix-obs-action}). The two uses are
unrelated, and a reader checking the source for that number will find both.

\textbf{Success criterion.} An episode is scored by whether the task's
success condition held at \emph{any} step of the episode, not only at the
final step. This is the looser of the two readings, and it is the one behind
every success rate in this paper, including the oracle success column of
Table~\ref{tab:all-tasks}. A task whose success condition can be satisfied
and then undone is credited for the step at which it was satisfied.

\subsection{How the reference baseline's evaluation differed}
\label{sec:eval-deviations}

The reference baseline was not run through the released benchmark module. It
was scored by the source training project's own evaluation code, which builds
each environment itself rather than through the released wrapper-construction
function. That code transcribes the per-task wrapper chain out of the
released wrapper-configuration table and cross-checks the transcription at
build time against a separately hardcoded set of the task ids that should
carry the curriculum wrapper, raising an error if the two disagree, so the
chains cannot drift apart unnoticed. Six differences remain. We list them so
that a reader who reproduces the baseline through the released code and gets a
different figure knows which to look at first, rather than concluding the
result does not reproduce.

\begin{enumerate}
  \item \textbf{An additional freeze wrapper on the \texttt{-Long-} variants.}
  The baseline's evaluation attaches the cue-phase freeze wrapper
  to the \texttt{-Long-} variants, holding the arm still through the cue
  phase -- zero end-effector delta, gripper held open -- which the released
  evaluation module does not attach to those variants by default. A
  reproduction through the released code should apply the same wrapper to
  match the baseline's numbers reported here.
  \item \textbf{Render overlays on.} The baseline's evaluation attaches the
  task-specific overlays to every task, where the released default is off and
  the released episodes were collected with them off. Each overlay only
  changes what is drawn for its own rendering and leaves the rest of the
  environment step unchanged (Appendix~\ref{sec:appendix-wrappers}), so this
  changes the recorded video and neither the observation the policy receives
  nor the success it is scored on.
  \item \textbf{An extra outermost debug wrapper.} The evaluation also mounts
  a reward-debugging wrapper that appears nowhere in the released
  wrapper-configuration table. It comes from the collector that recorded
  the PPO-sourced tasks, which mounted it outermost. It records the step's
  internal state for its own rendering and leaves the rest of the step
  unchanged, so the previous item's argument covers it.
  \item \textbf{A different seed lane and a smaller episode count.} The
  baseline uses its own start seed, chosen to be disjoint from the seeds the
  training episodes were drawn from (Appendix~\ref{sec:appendix-baseline}),
  and \BaselineEpisodesPerTask{} episodes per task. Both differ from the
  start seed and episode count above, so the
  initial states behind the baseline's numbers are a different set from the
  ones a default run visits, and a smaller one.
  \item \textbf{Episodes stop at the first success.} The baseline's rollout
  loop stops as soon as a step reports success, where the released
  code steps every episode to its horizon and takes the disjunction over
  steps. The success statistic is identical either way. The episode lengths
  are not, so any quantity derived from episode length, including wall-clock
  cost, will not match.
  \item \textbf{A split simulator backend.} The baseline's evaluation runs the
  motion-planned tasks on the CPU simulator backend and the PPO-sourced tasks
  on the GPU backend, splitting them by the \texttt{Source} column of
  Table~\ref{tab:all-tasks}, where the released evaluation default is the GPU
  backend for every task.
  The split is not arbitrary: it matches each task to the
  backend its own demonstrations were collected on, since the motion-planning
  collector builds its environments on the CPU backend and the collector
  behind the PPO-sourced tasks builds them on the GPU one.
\end{enumerate}

Items 2 and 3 are transparent to the score, since both wrappers leave the
rest of the environment step unchanged. Item 5 is transparent for a different
reason, and not one the wrappers can settle: the loop stops on
the same condition the released path takes its disjunction over, so a run
that stops early and a run that continues report the same success either way.
Item 4 changes which episodes are run, so it can move a rate within the
interval width of Section~\ref{sec:limitations} without any of the numbers
being wrong. Item 6 is the deviation this paper has not bounded. The two
backends do not integrate identical actions into identical trajectories, so
the split is not a matter of sampling and the interval width above says
nothing about it. We ran no paired comparison that would give its size or its
direction. A run on the released default is therefore a run under a different
physics backend, and a disagreement there is not evidence against either
number. Item 1
is the one that would make a reproduction disagree outright, and it disagrees
in the released code's favor until that code attaches the same wrapper.

\subsection{Chance floor: what a uniform-random policy scores}
\label{sec:appendix-random-floor}

Section~\ref{sec:baseline} reads the reference baseline against a measured
floor rather than an assumed one. This subsection is that measurement.

\textbf{Policy.} Every action is drawn uniformly from the environment's
action space. The policy has no parameters and never
reads the observation. Its action space is re-seeded at the start of each
episode from the same seed as the environment reset, so a rerun visits the
same trajectories.

\textbf{Protocol.} All \RandomFloorTasks{} canonical tasks,
\RandomFloorEpisodesPerTask{} episodes each, \RandomFloorEpisodes{} episodes
in total, from start seed \RandomFloorStartSeed{}. That lane is an
independent draw rather than a repeat of any other run in this paper: it is
disjoint from the evaluation module's own start seed above, from the
baseline's \BaselineEvalBaseSeed{} (Appendix~\ref{sec:appendix-baseline}),
and from the seeds the released demonstrations were collected under.
Success uses the same any-step criterion as everywhere else in this paper.

\textbf{One deviation from the released stack.} The released path builds
every environment with overlays disabled, which requires image-plus-state
observations. This run used state-only observations on the
\texttt{\RandomFloorSimBackend{}} simulator backend, because rendering
\RandomFloorEpisodes{} RGB rollouts was not possible in the window the
measurement had: an unrelated job held every GPU on the machine at full
utilization and near-full memory, and the renderer would not initialize.
Two observation-processing layers of the released chain are therefore
absent: the one that flattens the RGB-D observation into the released
format, and the one that converts joint positions to end-effector pose.

Neither can move a success statistic in this run, for a structural reason
rather than an empirical one. Both only rewrite what the
policy is shown, in different ways, and leave the action and the rest of the
environment step unchanged (Appendix~\ref{sec:appendix-wrappers}).
An observation-only layer can only change what a policy sees, and this policy
sees nothing. Neither layer touches the action, the physics, or the success
condition. The one layer in the released chain that does modify actions, the
cue-phase freeze wrapper, was applied on the
\RandomFloorCurriculumTasks{} tasks the released wrapper-configuration
table names and on no others.

\textbf{Result.} Of the \RandomFloorEpisodes{} episodes,
\RandomFloorSuccesses{} succeeded. Of the \RandomFloorTasks{} tasks,
\RandomFloorZeroTasks{} scored exactly zero. The remaining task,
\texttt{\RandomFloorNonzeroTask{}}, scored \RandomFloorNonzeroRate{}, whose
binomial standard error at $n=\RandomFloorEpisodesPerTask{}$ is
\RandomFloorNonzeroSE{}. Scored at the final step of the episode instead of
at any step, the count over the same rollouts is \RandomFloorEndSuccesses{}:
the one success was not held to the end of its episode.

This bounds one thing and no more. Chance contributes nothing to a success
rate on these tasks, so a low score is not an artifact of tasks that can be
passed by accident. The floor does not show that any task requires memory,
and it does not bound what a policy without retention could reach: a policy
that manipulates competently and remembers nothing is not a random one, and
nothing here measures one.

\textbf{A selection floor for the forgetful policy.} That second policy can be
bounded without running it, on the tasks whose success condition is a choice
among a fixed candidate set. There the environment source fixes the number of
candidates $k$, so an arm that manipulates competently and has forgotten the
cue still lands on the target with probability $1/k$. Of the
\NumChanceFloorEvaluated{} evaluated tasks, \NumChanceFloorTasks{} are
selections of this kind, with $k$ equal to \ChanceFloorCandidateSet{}. The
reference arm clears the floor on \NumChanceFloorAbove{} of them:
\NumChanceFloorWithinSE{} of its rates fall within one standard error of $1/k$
and the other \NumChanceFloorBelow{} fall below. Pooled over the
\ChanceFloorPooledEpisodes{} episodes those tasks contribute, it solves
\ChanceFloorPooledSuccesses{} against the \ChanceFloorPooledExpected{} a policy
sitting exactly at each floor would average -- a shortfall well inside the
\ChanceFloorPooledSD{} that null's own spread allows, so the arm is not below
the floor either. It is indistinguishable from it. Which of the two readings
this supports, a policy that has learned the manipulation and not the cue or a
policy that is failing for some other reason, the comparison does not decide.
Section~\ref{sec:failure-analysis} takes up the alternatives.

\begin{table}[h]
\centering\scriptsize
\caption{\textbf{Selection floors for the 7 evaluated tasks whose success condition names a finite candidate set.} $k$ is the number of candidates the success condition selects among, never transcribed by hand. Success is the reference arm's rate over 20 episodes. A rate below $1/k$ does not mean the arm is worse than forgetful: at this episode count the two are not separable. Nor does a rate above it mean the arm is better -- 0 of these rows clear their floor by more than the row's own standard error.}
\label{tab:chance-floor}
\begin{tabular}{lrrr}
\toprule
\textbf{Task} & $\mathbf{k}$ & $\mathbf{1/k}$ & \textbf{Success} \\
\midrule
\texttt{GatherAndRecall1-VLA-v0} & 3 & 0.333 & 0.35 \\
\texttt{GatherAndRecall3-VLA-v0} & 3 & 0.333 & 0.15 \\
\texttt{RememberColor5-Long-VLA-v0} & 5 & 0.200 & 0.30 \\
\texttt{RememberColor5-VLA-v0} & 5 & 0.200 & 0.25 \\
\texttt{RememberShape5-VLA-v0} & 5 & 0.200 & 0.05 \\
\texttt{RememberShapeAndColor3x2-VLA-v0} & 6 & 0.167 & 0.20 \\
\texttt{ShellGameShuffleColorLampTouch-Long-VLA-v0} & 3 & 0.333 & 0.35 \\
\bottomrule
\end{tabular}
\end{table}

Table~\ref{tab:chance-floor} gives each task's $k$, its floor, and the arm's
rate. The tasks that get no floor are excluded for one of two reasons: some
are continuous rather than selections, and the rest
need several correct choices at once, where $1/k$ describes no policy anyone
could run. Three limits apply throughout. The floor bounds a hypothetical
policy rather than any model that was run, and clearing it is not evidence of
retention, since an observation shortcut would clear it too. On the
\texttt{GatherAndRecall} family the bound is conditional on the manipulation
succeeding first, which makes it weaker than the rest. And the rule declines
to score \texttt{BunchOfColors3-Long-VLA-v0} and
\texttt{ChainOfColors3-Long-VLA-v0}, the evaluated tasks whose conditions
enumerate the largest candidate sets, so every floor that survives is one of
the highest available: a laxer rule would set a lower bar. Both score zero,
under the $1/k$ such a rule would have given them, so admitting them would not
soften the comparison.

\textbf{Where the single success fell.} On
\texttt{\RandomFloorNonzeroTask{}}, one of the \texttt{-Long-} variants that
the released evaluation module does not freeze through the cue phase by
default. Its base variant
carries the wrapper and scored zero. An arm that is not frozen through the
cue phase has that phase in which to satisfy a success condition by
accident, which is consistent with, though not proof of, why the one
success landed there rather than on the base variant. One episode is far
too few to attribute the success to the missing wrapper.

\textbf{Corroboration.} An independent uniform-policy run made earlier on the
benchmark's \RandomFloorCorroborationSplit{} split, from a different
implementation and at the evaluation module's own start seed rather than
this lane,
reports a success rate of \RandomFloorCorroborationRate{} over the
\RandomFloorSplitEpisodes{} episodes that split holds. The matching figure
from this run is \RandomFloorSplitSuccesses{} successes in the same
\RandomFloorSplitEpisodes{}. That is the comparison: the
\RandomFloorSuccesses{}-in-\RandomFloorEpisodes{} rate quoted above pools all
three splits and is not the same quantity. Neither run resolves a rate this
small. What the two agree on is that it is not distinguishable from zero.

\textbf{Software.} ManiSkill \texttt{\RandomFloorManiSkillVersion{}},
Gymnasium \texttt{\RandomFloorGymnasiumVersion{}}, PyTorch
\texttt{\RandomFloorTorchVersion{}}, Python
\texttt{\RandomFloorPythonVersion{}}.

\section{Wrapper Cookbook}
\label{sec:appendix-wrappers}

Every task in this benchmark is registered as a standard Gymnasium
environment. The VLA-facing observation and action space
(Appendix~\ref{sec:appendix-obs-action}) is produced by applying one
wrapper function immediately after the environment is constructed. The
released benchmark's own environment-construction helper performs the same
steps internally, described here rather than listed as code. That helper is
library code, invoked through a separate runnable example script
(Appendix~\ref{sec:appendix-eval-protocol}).

Constructing a task environment for this benchmark takes six explicit
settings, none left to a simulator default: the number of parallel
environments, the observation mode, the control mode, the render mode, the
reward mode, and the simulator backend. The observation mode and control
mode determine the shapes described in
Appendix~\ref{sec:appendix-obs-action}; the reward mode selects the
normalized dense reward Appendix~\ref{sec:appendix-schemas} describes. The
wrapper-construction function is then applied to that environment with
overlays disabled, producing the released observation and action space.
Resetting the environment with a given seed, sampling an action from the
action space, and stepping the environment once yields an observation, a
reward, a success flag, and the language instruction for that episode. Two
observation keys must be dropped before the observation reaches a policy, as
the next paragraph explains. The control
mode in particular has no safe default here: the curriculum wrapper
emits an action shaped for the delta end-effector pose control mode, and
constructing the environment under any other control mode produces a
runtime warning and a wrapper that no longer matches the action space.

\textbf{Two observation keys must be dropped before a policy sees them.}
Before filtering, the observation dictionary holds four keys, not two: the
two uint8 camera images (top and wrist) at $\ImageShape$ each, the
\StateDim{}-dimensional end-effector state, and two diagnostic channels
carrying privileged state that the wrapper stack injects on every task. On
\NumCueLeakConcealed{} of the benchmark's tasks, one of those diagnostic
channels is set at reset and echoed unchanged on every step
thereafter, so a policy fed the unfiltered dictionary can score well without
retaining anything. The affected ids and the reason the guard is missing are in
Appendix~\ref{sec:appendix-obs-action}. The instruction string is not part of
the observation dictionary. It is returned separately, alongside the reward
and success flag. The sampled action
is a \ActionDim{}-dimensional delta end-effector pose.

\textbf{Overlays default on in the wrapper function, but off in the
released evaluation.} The wrapper-construction function defaults to
rendering the debug overlays, but the released evaluation module's
configuration defaults that off, and the released episodes were collected
with it off. A caller who reuses the wrapper function directly without
disabling overlays therefore gets a configuration that is not the released
one. Debug render overlays draw a step
counter, live reward components, and task-specific state such as which cup
currently hides the target. Each overlay only changes what is drawn for
rendering, so it does not change what a policy is shown. Turning the
overlays on therefore changes the rendered frame and nothing a policy sees.
Video runs are the exception: the evaluation module switches to a
frame-returning render mode when video saving is requested and turns the
overlays on in the same branch, whatever the configured overlay flag says
(Appendix~\ref{sec:appendix-eval-protocol}), so a run that writes videos is
not built the same way a default run is. Both changes stay inside
rendering: the observation mode decides the observation, and neither change
reaches a policy.

\textbf{The wrapper table is the benchmark's boundary. The environment
registry is wider.} Registering the benchmark's environments registers
several task ids that are not covered by the released
wrapper-configuration table -- a shell-game variant and several larger
batteries-checker sizes among them -- so constructing one of those
environments succeeds on tasks the manifest does not list and this paper
does not cover. Applying the wrapper-construction function to one of them
raises an error naming the supported ids, rather than silently applying a
default configuration that may not match the task, so the mismatch surfaces
at construction rather than in the results.

\section{Per-Family Task Details}
\label{sec:appendix-families}

Section~\ref{sec:taxonomy} groups the \NumTasks{} tasks into \NumMemoryTypes{}
memory types. Within a memory type, tasks further group into \NumFamilies{} environment
families sharing the same underlying reward and success-condition code, at
different variant sizes or horizons. Each subsection below covers one
family: a motivating everyday analogy, its language instruction, task
description, a representative
rollout filmstrip, a table of every variant in the family, its verbatim
success condition, and a summary of its dense reward structure, written
from its \texttt{compute\_dense\_reward} source rather than quoted from it
(Section~\ref{sec:rewards}).

Every filmstrip in this appendix is sampled from the environment's render
channel, which draws debug text over the scene: a step counter, live reward
components, and task-specific state such as which cup currently hides the
target. None of it reaches the released observation, for two reasons the
datasheet states in full (Appendix~\ref{sec:datasheet}): the overlay classes
touch \texttt{render} and not \texttt{step}, and collection ran with them
switched off.

\subsection{Batteries Checker Easy}
\label{sec:family-batteries-checker-easy}

\paragraph{Motivation.}
An electrician sorting a box of batteries pulled from old equipment tests each one in a socket, reads the result, and sets it aside -- retesting a battery already confirmed working wastes time, and skipping one leaves the sort incomplete.

\begin{quote}\itshape
``Find all working batteries by inserting each one into the socket, observing the lamp result, and then pressing the button to confirm.''
\end{quote}

\paragraph{Task.}
Sequential verification: insert batteries into a socket, observe the lamp result, remember which batteries work, and press the confirmation button.

\begin{figure}[H]
\centering
\includegraphics[width=0.95\linewidth]{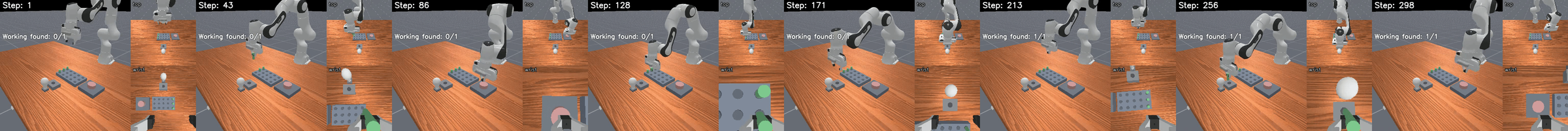}
\vspace{-8pt}
\caption{\textbf{Rollout filmstrip, \texttt{BatteriesCheckerEasy-3-VLA-v0} (a representative variant).}}
\end{figure}

\begin{table}[H]
\centering\small
\caption{\textbf{Batteries Checker Easy: variants, horizon, data source, median realized episode length, and information gap.} Lengths and gaps are in environment steps. Every variant contributes 250 retained demonstrations, and their oracle success rate is 1.000. Each horizon's split is given by Table~\ref{tab:splits}.}
\begin{tabular}{@{}lrlrl@{}}
\toprule
\textbf{Env ID} & \shortstack[r]{\textbf{Horizon}\\(steps)} & \textbf{Source} & \shortstack[r]{\textbf{Median}\\length} & \textbf{Gap} \\
\midrule
\texttt{BatteriesCheckerEasy-3-VLA-v0} & 540 & MP & 363 & undetermined \\
\texttt{BatteriesCheckerEasy-6-VLA-v0} & 1080 & MP & 727 & undetermined \\
\bottomrule
\end{tabular}
\end{table}

\paragraph{Information gap: provenance.}
Every gap in the table above is read from the environment source.
\begin{itemize}
\item \texttt{BatteriesCheckerEasy-3-VLA-v0} (gap undetermined), \texttt{BatteriesCheckerEasy-6-VLA-v0} (): The lamp cue (working/not) is revealed at insertion and stays visible for only 7 steps after a working hit. This variant returns the tested battery to its tray slot automatically rather than requiring the agent to carry it back, advancing to the confirmation stage on the first step that stage is reached. The button-press requirement therefore starts once the agent has inserted the battery and read the lamp, an interval timed by the agent's own motion, not by a source constant.
\end{itemize}

\paragraph{Success condition.}
Same criterion as the hard variant: all working batteries must be found through completed check-confirm cycles.

\paragraph{Dense reward structure.}
The easy variant does not define its own dense reward. It reuses the hard variant's shaping unchanged (see batteries\_checker\_hard below) -- only the docstring, and the environment's own auto-return-to-slot behavior, differ.

\subsection{Batteries Checker Hard}
\label{sec:family-batteries-checker-hard}

\paragraph{Motivation.}
The same sorting task at a larger box and a longer shift: the electrician has to hold a running tally of which batteries are already confirmed across many more items and a much longer interval before the sort is done.

\begin{quote}\itshape
``Find all working batteries by inserting each one into the socket, observing the lamp result, returning it from the socket to its initial slot, and then pressing the button to confirm.''
\end{quote}

\paragraph{Task.}
Hard sequential verification: test each battery, remember the result, return it to its slot, and submit the working set.

\begin{figure}[H]
\centering
\includegraphics[width=0.95\linewidth]{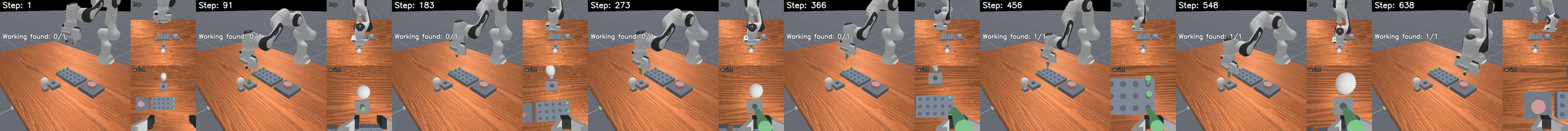}
\vspace{-8pt}
\caption{\textbf{Rollout filmstrip, \texttt{BatteriesCheckerHard-3-VLA-v0} (a representative variant).}}
\end{figure}

\begin{table}[H]
\centering\small
\caption{\textbf{Batteries Checker Hard: variants, horizon, data source, median realized episode length, and information gap.} Lengths and gaps are in environment steps. Every variant contributes 250 retained demonstrations, and their oracle success rate is 1.000. Each horizon's split is given by Table~\ref{tab:splits}.}
\begin{tabular}{@{}lrlrl@{}}
\toprule
\textbf{Env ID} & \shortstack[r]{\textbf{Horizon}\\(steps)} & \textbf{Source} & \shortstack[r]{\textbf{Median}\\length} & \textbf{Gap} \\
\midrule
\texttt{BatteriesCheckerHard-3-VLA-v0} & 1080 & MP & 699 & undetermined \\
\texttt{BatteriesCheckerHard-6-VLA-v0} & 2160 & MP & 1416 & undetermined \\
\bottomrule
\end{tabular}
\end{table}

\paragraph{Information gap: provenance.}
Every gap in the table above is read from the environment source.
\begin{itemize}
\item \texttt{BatteriesCheckerHard-3-VLA-v0} (gap undetermined), \texttt{BatteriesCheckerHard-6-VLA-v0} (): The lamp cue (working/not) is revealed at insertion and stays visible for only 7 steps after a working hit, but the button-press requirement only starts once the battery is physically returned to its tray slot, a transition timed by the agent's own motion, not a source constant.
\end{itemize}

\paragraph{Success condition.}
Every working battery must be discovered through the full insert-return-confirm procedure. Partial progress does not count as success.

\paragraph{Dense reward structure.}
A stage-gated sum keyed on the current action stage: during the insert stage, reach-toward-the-nearest-unchecked-battery plus carry-the-grasped-battery-toward-the-socket terms. During the return stage, a term that shrinks the distance between the active battery and its home tray slot. During the confirm stage, a button-reach plus button-press-depth term. On top of this, one-off bonuses fire on each new insert, return, and confirm event, a larger bonus fires the first time a battery is confirmed working, and a penalty applies to pressing the confirm button outside the confirm stage. Minus the smoothness penalty throughout, the whole stage-gated sum is masked to steps where the action is actually expected, then a fixed success bonus replaces it once every working battery has been found.

\subsection{Blink Count Button Press}
\label{sec:family-blink-count-button-press}

\paragraph{Motivation.}
Consumer electronics report internal fault codes by blinking a status LED a fixed number of times. A technician reading one has to count silently through the blink sequence and only act -- pressing a diagnostic button the matching number of times -- once the light goes solid.

\begin{quote}\itshape
``Count how many times the blue lamp blinks, press the red button exactly that many times when the red lamp turns green, then press the black button to submit your answer.''
\end{quote}

\paragraph{Task.}
Temporal counting: count blue-lamp blinks, wait for execution, press the red button exactly that many times, then submit.

\begin{figure}[H]
\centering
\includegraphics[width=0.95\linewidth]{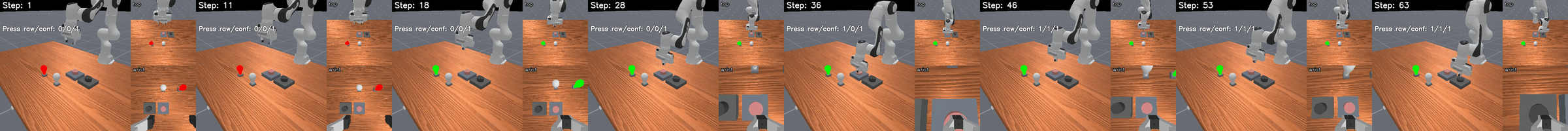}
\vspace{-8pt}
\caption{\textbf{Rollout filmstrip, \texttt{BlinkCountButtonPressEasy-VLA-v0} (a representative variant).}}
\end{figure}

\begin{table}[H]
\centering\small
\caption{\textbf{Blink Count Button Press: variants, horizon, data source, median realized episode length, and information gap.} Lengths and gaps are in environment steps. Every variant contributes 250 retained demonstrations, and their oracle success rate is 1.000. Each horizon's split is given by Table~\ref{tab:splits}.}
\begin{tabular}{@{}lrlrl@{}}
\toprule
\textbf{Env ID} & \shortstack[r]{\textbf{Horizon}\\(steps)} & \textbf{Source} & \shortstack[r]{\textbf{Median}\\length} & \textbf{Gap} \\
\midrule
\texttt{BlinkCountButtonPressEasy-VLA-v0} & 150 & MP & 94 & 0 \\
\texttt{BlinkCountButtonPressMedium-VLA-v0} & 200 & MP & 121 & 0 \\
\texttt{BlinkCountButtonPressHard-VLA-v0} & 300 & MP & 153 & 0 \\
\texttt{BlinkCountButtonPressEasy-Long-VLA-v0} & 1200 & MP & 199 & 0 \\
\texttt{BlinkCountButtonPressMedium-Long-VLA-v0} & 1200 & MP & 490 & 0 \\
\texttt{BlinkCountButtonPressHard-Long-VLA-v0} & 1200 & MP & 788 & 0 \\
\bottomrule
\end{tabular}
\end{table}

\paragraph{Information gap: provenance.}
Every gap in the table above is read from the environment source.
\begin{itemize}
\item \texttt{BlinkCountButtonPressEasy-VLA-v0} (gap 0), \texttt{BlinkCountButtonPressMedium-VLA-v0} (), \texttt{BlinkCountButtonPressHard-VLA-v0} (), \texttt{BlinkCountButtonPressEasy-Long-VLA-v0} (), \texttt{BlinkCountButtonPressMedium-Long-VLA-v0} (), \texttt{BlinkCountButtonPressHard-Long-VLA-v0} (): The manipulation requirement begins on the very step the cue phase ends, with no intervening empty phase: a 0-step gap.
\end{itemize}

\paragraph{Success condition.}
\begin{itemize}
\item Success is produced only when the black submit button is pressed.
\item At submit time, the counted number of valid red-button presses must exactly match the target blink count.
\end{itemize}

\paragraph{Dense reward structure.}
A phase-decomposed sum over four regimes (cue-watching, pre-lift press-and-count, post-press lift-and-release, and submit) rather than one continuous shaping term. During the cue it rewards staying near the button without pressing. During the count phase it rewards reach, press depth, and progress toward matching the target blink count, with a one-off bonus per new raw press. After each press it rewards lifting clear and releasing before the next press is allowed, with lift/hover penalties that discourage skipping the lift. Once the counted presses equal the target it rewards reaching and pressing the black confirm button. Early presses during the cue and holding the button down during the lift phase are both penalized. Minus the smoothness penalty throughout, a fixed penalty applies on an incorrect confirm.

\subsection{Bunch Of Colors}
\label{sec:family-bunch-of-colors}

\paragraph{Motivation.}
A shopper glances at a fruit display, notices which colors are represented, and only later, after being asked to fetch the colors you saw, has to act on that memory without the display still in view.

\begin{quote}\itshape
``Observe which colored cubes appear during the cue, wait, then touch all of them in any order and press the center button.''
\end{quote}

\paragraph{Task.}
Set memory: observe a set of colored cubes, then touch all remembered colors in any order and press the center button.

\begin{figure}[H]
\centering
\includegraphics[width=0.95\linewidth]{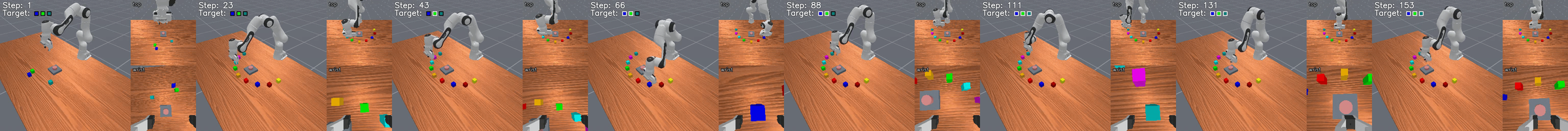}
\vspace{-8pt}
\caption{\textbf{Rollout filmstrip, \texttt{BunchOfColors3-VLA-v0} (a representative variant).}}
\end{figure}

\begin{table}[H]
\centering\small
\caption{\textbf{Bunch Of Colors: variants, horizon, data source, median realized episode length, and information gap.} Lengths and gaps are in environment steps. Every variant contributes 250 retained demonstrations, and their oracle success rate is 1.000. Each horizon's split is given by Table~\ref{tab:splits}.}
\begin{tabular}{@{}lrlrl@{}}
\toprule
\textbf{Env ID} & \shortstack[r]{\textbf{Horizon}\\(steps)} & \textbf{Source} & \shortstack[r]{\textbf{Median}\\length} & \textbf{Gap} \\
\midrule
\texttt{BunchOfColors3-VLA-v0} & 400 & MP & 158 & 3 (2--4) \\
\texttt{BunchOfColors5-VLA-v0} & 400 & MP & 242 & 3 (2--4) \\
\texttt{BunchOfColors7-VLA-v0} & 400 & MP & 327 & 3 (2--4) \\
\texttt{BunchOfColors3-Long-VLA-v0} & 700 & MP & 449 & 225 (138--313) \\
\texttt{BunchOfColors5-Long-VLA-v0} & 700 & MP & 524 & 225 (138--313) \\
\texttt{BunchOfColors7-Long-VLA-v0} & 700 & MP & 569 & 225 (138--313) \\
\bottomrule
\end{tabular}
\end{table}

\paragraph{Information gap: provenance.}
Every gap in the table above is read from the environment source.
\begin{itemize}
\item \texttt{BunchOfColors3-VLA-v0} (gap 3 (2--4)), \texttt{BunchOfColors5-VLA-v0} (), \texttt{BunchOfColors7-VLA-v0} (): The manipulation phase begins after an empty phase whose length is drawn uniformly between 1 and 5 steps at reset, so the gap is that draw itself.
\item \texttt{BunchOfColors3-Long-VLA-v0} (gap 225 (138--313)), \texttt{BunchOfColors5-Long-VLA-v0} (), \texttt{BunchOfColors7-Long-VLA-v0} (): The manipulation phase begins after an empty phase whose length is drawn uniformly between 50 and 400 steps at reset, so the gap is that draw itself.
\end{itemize}

\paragraph{Success condition.}
The robot must touch all target colors and avoid wrong selections.

\paragraph{Dense reward structure.}
Reach-toward-goal shaping plus a count-based term that scales linearly with the fraction of the target color set touched so far (independent of order), a smaller bonus for being close to static, a flat penalty applies per wrong-color touch, and a displacement penalty discourages disturbing untouched cubes during the active (post-cue) phase. The whole reward is masked to the active phase and replaced by a fixed terminal value once every target color has been touched with no wrong touches.

\subsection{Chain Of Colors}
\label{sec:family-chain-of-colors}

\paragraph{Motivation.}
The same glance at a display, but the shopper must reproduce not just which colors were there but the exact order they were arranged in -- closer to memorizing a combination than a shopping list.

\begin{quote}\itshape
``Observe which colored cubes appear during the cue, wait, then touch all of them in the same order as the cubes were shown and press the center button.''
\end{quote}

\paragraph{Task.}
Ordered sequence memory: observe colors in order, then touch the remembered colors in the same order before submission.

\begin{figure}[H]
\centering
\includegraphics[width=0.95\linewidth]{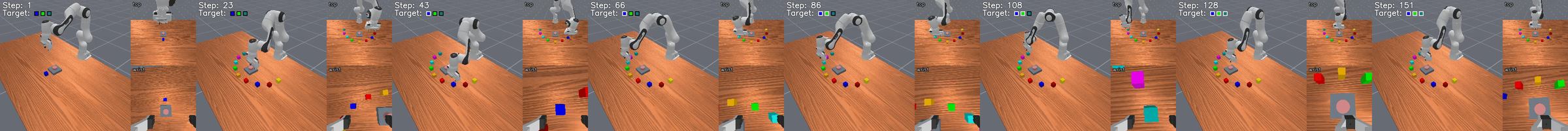}
\vspace{-8pt}
\caption{\textbf{Rollout filmstrip, \texttt{ChainOfColors3-VLA-v0} (a representative variant).}}
\end{figure}

\begin{table}[H]
\centering\small
\caption{\textbf{Chain Of Colors: variants, horizon, data source, median realized episode length, and information gap.} Lengths and gaps are in environment steps. Every variant contributes 250 retained demonstrations, and their oracle success rate is 1.000. Each horizon's split is given by Table~\ref{tab:splits}.}
\begin{tabular}{@{}lrlrl@{}}
\toprule
\textbf{Env ID} & \shortstack[r]{\textbf{Horizon}\\(steps)} & \textbf{Source} & \shortstack[r]{\textbf{Median}\\length} & \textbf{Gap} \\
\midrule
\texttt{ChainOfColors3-VLA-v0} & 400 & MP & 165 & 3 (2--4) \\
\texttt{ChainOfColors5-VLA-v0} & 400 & MP & 253 & 3 (2--4) \\
\texttt{ChainOfColors7-VLA-v0} & 400 & MP & 342 & 3 (2--4) \\
\texttt{ChainOfColors3-Long-VLA-v0} & 800 & MP & 559 & 225 (138--313) \\
\texttt{ChainOfColors5-Long-VLA-v0} & 1000 & MP & 740 & 225 (138--313) \\
\texttt{ChainOfColors7-Long-VLA-v0} & 1200 & MP & 916 & 225 (138--313) \\
\bottomrule
\end{tabular}
\end{table}

\paragraph{Information gap: provenance.}
Every gap in the table above is read from the environment source.
\begin{itemize}
\item \texttt{ChainOfColors3-VLA-v0} (gap 3 (2--4)), \texttt{ChainOfColors5-VLA-v0} (), \texttt{ChainOfColors7-VLA-v0} (): The manipulation phase begins after an empty phase whose length is drawn uniformly between 1 and 5 steps at reset, so the gap is that draw itself.
\item \texttt{ChainOfColors3-Long-VLA-v0} (gap 225 (138--313)), \texttt{ChainOfColors5-Long-VLA-v0} (), \texttt{ChainOfColors7-Long-VLA-v0} (): The manipulation phase begins after an empty phase whose length is drawn uniformly between 50 and 400 steps at reset, so the gap is that draw itself.
\end{itemize}

\paragraph{Success condition.}
Every target color must be touched in the correct order, with no wrong or out-of-order selections.

\paragraph{Dense reward structure.}
Structurally identical to bunch\_of\_colors -- reach shaping, a per-touch count reward, a smaller bonus for being close to static, a wrong-touch penalty, and a cube-displacement penalty, all masked to the active phase with a fixed terminal bonus on success -- except the count reward is computed from the length of the correctly *ordered* touch prefix (a cumulative-max of an elementwise match against the true sequence) rather than from set membership, so an out-of-order touch does not extend the credited count even if the color is correct.

\subsection{Find Imposter Color}
\label{sec:family-find-imposter-color}

\paragraph{Motivation.}
A line inspector is shown a reference tray of parts, the tray is rearranged out of view, and the inspector has to find the one part now present that was not in the original set -- comparing against a memory of the reference, not a live side-by-side.

\begin{quote}\itshape
``Observe the cubes shown, wait, then touch the cube whose color was not present before.''
\end{quote}

\paragraph{Task.}
Out-of-set color detection: remember shown colors and touch the cube whose color was absent from the cue.

\begin{figure}[H]
\centering
\includegraphics[width=0.95\linewidth]{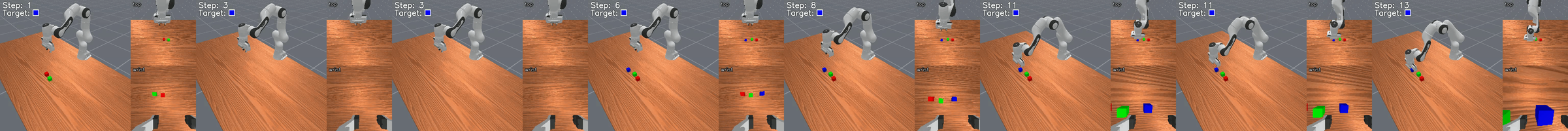}
\vspace{-8pt}
\caption{\textbf{Rollout filmstrip, \texttt{FindImposterColor3-VLA-v0} (a representative variant).}}
\end{figure}

\begin{table}[H]
\centering\small
\caption{\textbf{Find Imposter Color: variants, horizon, data source, median realized episode length, and information gap.} Lengths and gaps are in environment steps. Every variant contributes 250 retained demonstrations, and their oracle success rate is 1.000. Each horizon's split is given by Table~\ref{tab:splits}.}
\begin{tabular}{@{}lrlrl@{}}
\toprule
\textbf{Env ID} & \shortstack[r]{\textbf{Horizon}\\(steps)} & \textbf{Source} & \shortstack[r]{\textbf{Median}\\length} & \textbf{Gap} \\
\midrule
\texttt{FindImposterColor3-VLA-v0} & 25 & PPO & 15 & 3 (2--4) \\
\texttt{FindImposterColor5-VLA-v0} & 25 & PPO & 15 & 3 (2--4) \\
\texttt{FindImposterColor9-VLA-v0} & 25 & PPO & 16 & 3 (2--4) \\
\bottomrule
\end{tabular}
\end{table}

\paragraph{Information gap: provenance.}
Every gap in the table above is read from the environment source.
\begin{itemize}
\item \texttt{FindImposterColor3-VLA-v0} (gap 3 (2--4)), \texttt{FindImposterColor5-VLA-v0} (), \texttt{FindImposterColor9-VLA-v0} (): The manipulation phase begins after an empty phase whose length is drawn uniformly between 1 and 5 steps at reset, so the gap is that draw itself.
\end{itemize}

\paragraph{Success condition.}
TCP within GOAL\_THRESH of the imposter cube in the manipulation phase.

\paragraph{Dense reward structure.}
A single-phase reach-and-hold shaping term, saturating with the distance from the tool center point to the target cube, plus a smaller bonus for being close to static and for being static while the object is placed, minus the smoothness penalty. No separate distractor-avoidance term. The task is entirely encoded through which cube counts as the goal position. A fixed terminal bonus replaces the shaped reward on success.

\subsection{Find Imposter Shape}
\label{sec:family-find-imposter-shape}

\paragraph{Motivation.}
The same inspection task, but the odd part is identified by its geometry rather than its color -- the kind of sorting-by-profile an inspector does on a line where parts are backlit or unpainted and color is not available as a cue.

\begin{quote}\itshape
``Observe the shapes shown, wait, then touch the object whose shape was not present before.''
\end{quote}

\paragraph{Task.}
Out-of-set shape detection: remember shown shapes and touch the object whose shape was absent from the cue.

\begin{figure}[H]
\centering
\includegraphics[width=0.95\linewidth]{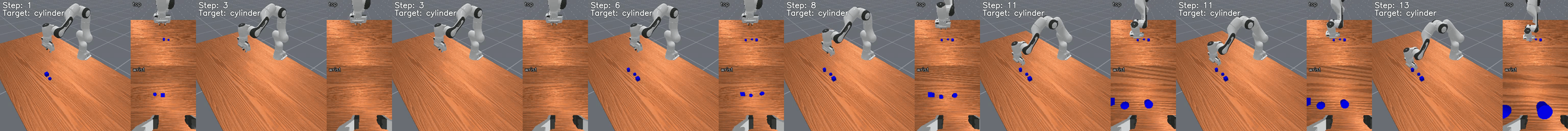}
\vspace{-8pt}
\caption{\textbf{Rollout filmstrip, \texttt{FindImposterShape3-VLA-v0} (a representative variant).}}
\end{figure}

\begin{table}[H]
\centering\small
\caption{\textbf{Find Imposter Shape: variants, horizon, data source, median realized episode length, and information gap.} Lengths and gaps are in environment steps. Every variant contributes 250 retained demonstrations, and their oracle success rate is 1.000. Each horizon's split is given by Table~\ref{tab:splits}.}
\begin{tabular}{@{}lrlrl@{}}
\toprule
\textbf{Env ID} & \shortstack[r]{\textbf{Horizon}\\(steps)} & \textbf{Source} & \shortstack[r]{\textbf{Median}\\length} & \textbf{Gap} \\
\midrule
\texttt{FindImposterShape3-VLA-v0} & 25 & PPO & 14 & 3 (2--4) \\
\texttt{FindImposterShape5-VLA-v0} & 25 & PPO & 15 & 3 (2--4) \\
\texttt{FindImposterShape9-VLA-v0} & 25 & PPO & 15 & 3 (2--4) \\
\bottomrule
\end{tabular}
\end{table}

\paragraph{Information gap: provenance.}
Every gap in the table above is read from the environment source.
\begin{itemize}
\item \texttt{FindImposterShape3-VLA-v0} (gap 3 (2--4)), \texttt{FindImposterShape5-VLA-v0} (), \texttt{FindImposterShape9-VLA-v0} (): The manipulation phase begins after an empty phase whose length is drawn uniformly between 1 and 5 steps at reset, so the gap is that draw itself.
\end{itemize}

\paragraph{Success condition.}
TCP within GOAL\_THRESH of the imposter shape in the manipulation phase.

\paragraph{Dense reward structure.}
Identical reward code to find\_imposter\_color: reach-to-goal shaping, a small static-robot bonus, and the smoothness penalty, with a fixed terminal bonus on success.

\subsection{Find Imposter Shape And Color}
\label{sec:family-find-imposter-shape-and-color}

\paragraph{Motivation.}
The inspection task again, now requiring both shape and color to match the memorized reference for a part to be accepted -- a part that is the right shape in the wrong color, or vice versa, still counts as the imposter.

\begin{quote}\itshape
``Observe the objects shown, wait, then touch the object whose shape and color combination was not present before.''
\end{quote}

\paragraph{Task.}
Out-of-set binding detection: remember shape-color combinations and touch the novel combination.

\begin{figure}[H]
\centering
\includegraphics[width=0.95\linewidth]{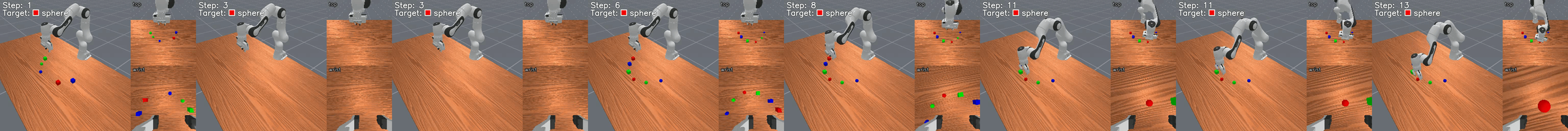}
\vspace{-8pt}
\caption{\textbf{Rollout filmstrip, \texttt{FindImposterShapeAndColor3x2-VLA-v0} (a representative variant).}}
\end{figure}

\begin{table}[H]
\centering\small
\caption{\textbf{Find Imposter Shape And Color: variants, horizon, data source, median realized episode length, and information gap.} Lengths and gaps are in environment steps. Every variant contributes 250 retained demonstrations, and their oracle success rate is 1.000. Each horizon's split is given by Table~\ref{tab:splits}.}
\begin{tabular}{@{}lrlrl@{}}
\toprule
\textbf{Env ID} & \shortstack[r]{\textbf{Horizon}\\(steps)} & \textbf{Source} & \shortstack[r]{\textbf{Median}\\length} & \textbf{Gap} \\
\midrule
\texttt{FindImposterShapeAndColor3x2-VLA-v0} & 25 & PPO & 16 & 3 (2--4) \\
\texttt{FindImposterShapeAndColor3x3-VLA-v0} & 25 & PPO & 15 & 3 (2--4) \\
\texttt{FindImposterShapeAndColor5x3-VLA-v0} & 40 & PPO & 15 & 3 (2--4) \\
\bottomrule
\end{tabular}
\end{table}

\paragraph{Information gap: provenance.}
Every gap in the table above is read from the environment source.
\begin{itemize}
\item \texttt{FindImposterShapeAndColor3x2-VLA-v0} (gap 3 (2--4)), \texttt{FindImposterShapeAndColor3x3-VLA-v0} (), \texttt{FindImposterShapeAndColor5x3-VLA-v0} (): The manipulation phase begins after an empty phase whose length is drawn uniformly between 1 and 5 steps at reset, so the gap is that draw itself.
\end{itemize}

\paragraph{Success condition.}
TCP within GOAL\_THRESH of the imposter object in the manipulation phase.

\paragraph{Dense reward structure.}
Identical reward code to find\_imposter\_color: reach-to-goal shaping, a small static-robot bonus, and the smoothness penalty, with a fixed terminal bonus on success.

\subsection{Gather And Recall}
\label{sec:family-gather-and-recall}

\paragraph{Motivation.}
A warehouse picker is told to keep filling a cart from the shelves and, when a floor light flashes, immediately drop everything and deliver to the marked dock -- the pending instruction has to survive an unrelated, longer primary task until its cue fires.

\begin{quote}\itshape
``Move all cubes onto the disc. A lamp will briefly flash while you work. After all cubes are placed, press the button matching the flash color.''
\end{quote}

\paragraph{Task.}
Concurrent manipulation plus cue recall: gather cubes while a lamp briefly flashes, then press the button matching the remembered color.

\begin{figure}[H]
\centering
\includegraphics[width=0.95\linewidth]{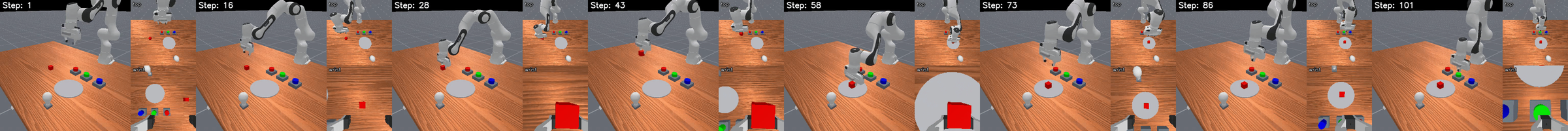}
\vspace{-8pt}
\caption{\textbf{Rollout filmstrip, \texttt{GatherAndRecall1-VLA-v0} (a representative variant).}}
\end{figure}

\begin{table}[H]
\centering\small
\caption{\textbf{Gather And Recall: variants, horizon, data source, median realized episode length, and information gap.} Lengths and gaps are in environment steps. Every variant contributes 250 retained demonstrations, and their oracle success rate is 1.000. Each horizon's split is given by Table~\ref{tab:splits}.}
\begin{tabular}{@{}lrlrl@{}}
\toprule
\textbf{Env ID} & \shortstack[r]{\textbf{Horizon}\\(steps)} & \textbf{Source} & \shortstack[r]{\textbf{Median}\\length} & \textbf{Gap} \\
\midrule
\texttt{GatherAndRecall1-VLA-v0} & 200 & MP & 115 & undetermined \\
\texttt{GatherAndRecall3-VLA-v0} & 400 & MP & 318 & undetermined \\
\texttt{GatherAndRecall5-VLA-v0} & 600 & MP & 514 & undetermined \\
\texttt{GatherAndRecall7-VLA-v0} & 800 & MP & 721 & undetermined \\
\texttt{GatherAndRecall9-VLA-v0} & 1000 & MP & 900 & undetermined \\
\bottomrule
\end{tabular}
\end{table}

\paragraph{Information gap: provenance.}
Every gap in the table above is read from the environment source.
\begin{itemize}
\item \texttt{GatherAndRecall1-VLA-v0} (gap undetermined), \texttt{GatherAndRecall3-VLA-v0} (), \texttt{GatherAndRecall5-VLA-v0} (), \texttt{GatherAndRecall7-VLA-v0} (), \texttt{GatherAndRecall9-VLA-v0} (): The lamp flash triggers once a randomly drawn number of cubes are on the disc and lasts a randomly drawn duration between 8 and 14 steps, but the button-press requirement only starts once all cubes are on the disc -- a transition timed by how fast the agent places the remaining cubes after the flash ends, not a source constant.
\end{itemize}

\paragraph{Success condition.}
All cubes detected on the disc AND the correct color button is pressed.

\paragraph{Dense reward structure.}
A concurrent-task sum: reach-to-nearest-unplaced-cube shaping, a placement-progress term proportional to the fraction of cubes already on the disc, and, only once every cube is on the disc, a reach term toward the button whose color matches the remembered lamp flash plus a press-depth term for that same button. Minus the smoothness penalty throughout. A fixed penalty applies when a wrong-color button is pressed after all cubes are placed, and a fixed success bonus replaces the shaped reward on success.

\subsection{Intercept}
\label{sec:family-intercept}

\paragraph{Motivation.}
A part rolls down a conveyor toward the end of the line. A worker has to judge its path and reach the interception point before it falls off, acting on a trajectory rather than a fixed target location.

\begin{quote}\itshape
``Intercept the rolling ball by moving to its path and deflecting it toward the target.''
\end{quote}

\paragraph{Task.}
Predictive interception: infer a rolling ball path and deflect it toward the target.

\begin{figure}[H]
\centering
\includegraphics[width=0.95\linewidth]{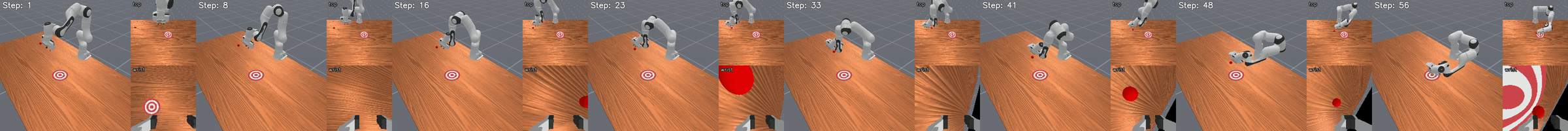}
\vspace{-8pt}
\caption{\textbf{Rollout filmstrip, \texttt{InterceptSlow-VLA-v0} (a representative variant).}}
\end{figure}

\begin{table}[H]
\centering\small
\caption{\textbf{Intercept: variants, horizon, data source, median realized episode length, and information gap.} Lengths and gaps are in environment steps. Every variant contributes 250 retained demonstrations, and their oracle success rate is 1.000. Each horizon's split is given by Table~\ref{tab:splits}.}
\begin{tabular}{@{}lrlrl@{}}
\toprule
\textbf{Env ID} & \shortstack[r]{\textbf{Horizon}\\(steps)} & \textbf{Source} & \shortstack[r]{\textbf{Median}\\length} & \textbf{Gap} \\
\midrule
\texttt{InterceptSlow-VLA-v0} & 60 & PPO & 43 & cue never hidden \\
\texttt{InterceptMedium-VLA-v0} & 60 & PPO & 41 & cue never hidden \\
\texttt{InterceptFast-VLA-v0} & 60 & PPO & 30 & cue never hidden \\
\bottomrule
\end{tabular}
\end{table}

\paragraph{Information gap: provenance.}
Every gap in the table above is read from the environment source.
\begin{itemize}
\item \texttt{InterceptSlow-VLA-v0} (gap cue never hidden), \texttt{InterceptMedium-VLA-v0} (), \texttt{InterceptFast-VLA-v0} (): The ball is spawned moving and stays physically present and observable for the entire episode, with its position read directly at every step. There is no cue-hide phase or occlusion constant, so no cue-to-action retention gap exists to measure.
\end{itemize}

\paragraph{Success condition.}
The ball center must end up inside the goal radius.

\paragraph{Dense reward structure.}
Two-regime shaping switched by a sticky reached-target flag: before the tool center point first comes within a fixed distance of a hit-offset pose next to the moving ball, the reward is dominated by reach-to-hit-pose shaping. After that flag is set, the reward is dominated by a placement term that shrinks the ball's xy distance to the goal region. A fixed terminal bonus replaces the shaped reward on success. Note this family has no explicit smoothness penalty term.

\subsection{Intercept Grab}
\label{sec:family-intercept-grab}

\paragraph{Motivation.}
The same rolling part now has to be caught and held rather than merely deflected -- closer to fielding a dropped tool before it hits the floor than tapping it aside.

\begin{quote}\itshape
``Intercept the rolling ball and grasp it to stop it.''
\end{quote}

\paragraph{Task.}
Predictive capture: infer a rolling ball path and grasp or stop it.

\begin{figure}[H]
\centering
\includegraphics[width=0.95\linewidth]{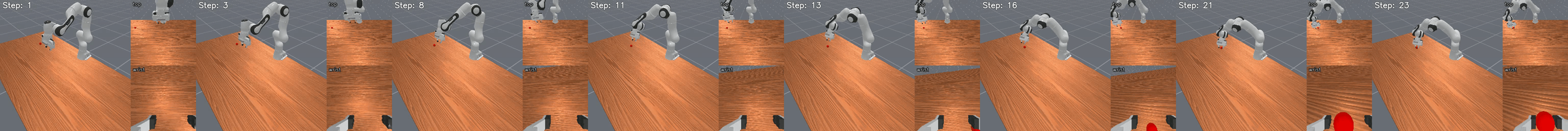}
\vspace{-8pt}
\caption{\textbf{Rollout filmstrip, \texttt{InterceptGrabSlow-VLA-v0} (a representative variant).}}
\end{figure}

\begin{table}[H]
\centering\small
\caption{\textbf{Intercept Grab: variants, horizon, data source, median realized episode length, and information gap.} Lengths and gaps are in environment steps. Every variant contributes 250 retained demonstrations, and their oracle success rate is 1.000. Each horizon's split is given by Table~\ref{tab:splits}.}
\begin{tabular}{@{}lrlrl@{}}
\toprule
\textbf{Env ID} & \shortstack[r]{\textbf{Horizon}\\(steps)} & \textbf{Source} & \shortstack[r]{\textbf{Median}\\length} & \textbf{Gap} \\
\midrule
\texttt{InterceptGrabSlow-VLA-v0} & 60 & PPO & 24 & cue never hidden \\
\texttt{InterceptGrabMedium-VLA-v0} & 60 & PPO & 33 & cue never hidden \\
\texttt{InterceptGrabFast-VLA-v0} & 60 & PPO & 50 & cue never hidden \\
\bottomrule
\end{tabular}
\end{table}

\paragraph{Information gap: provenance.}
Every gap in the table above is read from the environment source.
\begin{itemize}
\item \texttt{InterceptGrabSlow-VLA-v0} (gap cue never hidden), \texttt{InterceptGrabMedium-VLA-v0} (), \texttt{InterceptGrabFast-VLA-v0} (): The docstring states explicitly 'the robot has no separate observation phase: it must react immediately'. The ball is continuously visible, so there is no cue-hide phase or occlusion constant and no retention gap to measure.
\end{itemize}

\paragraph{Success condition.}
The ball must be grasped and the robot must be static at the end of the episode step.

\paragraph{Dense reward structure.}
A cascade of proximity- and grasp-state-gated bonuses rather than one shaping term: base reach-to-ball shaping, a flat bonus once the ball's position is inside the gripper's bounding geometry, an added proximity bump within a very close radius, an exponential-decay bonus that rewards closing the gripper toward the ball's exact diameter once it is bracketed, another flat bonus once gripper width is near that target, and a large flat bonus once the ball is grasped. A still-larger fixed bonus replaces the shaped reward on success. Two computed static-robot bonus quantities are never added into the returned reward -- unused in the source, not part of the actual shaping.

\subsection{Remember Color}
\label{sec:family-remember-color}

\paragraph{Motivation.}
A sorter is shown one reference cube's color, the cube is taken away, and several identical-looking cubes of different colors are set out -- the sorter has to pick the one matching a color no longer visible.

\begin{quote}\itshape
``Observe the cube's color, wait, then touch the cube of the same color.''
\end{quote}

\paragraph{Task.}
Delayed color recall: observe one target color, wait, and touch the cube with the same color.

\begin{figure}[H]
\centering
\includegraphics[width=0.95\linewidth]{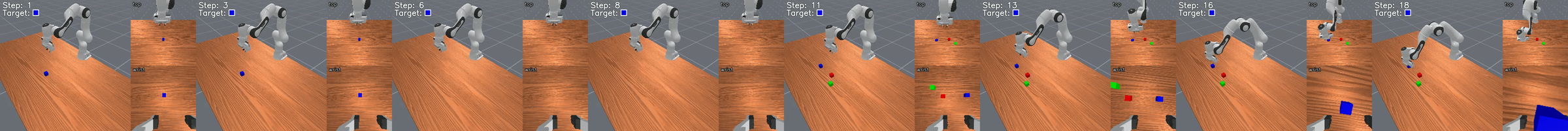}
\vspace{-8pt}
\caption{\textbf{Rollout filmstrip, \texttt{RememberColor3-VLA-v0} (a representative variant).}}
\end{figure}

\begin{table}[H]
\centering\small
\caption{\textbf{Remember Color: variants, horizon, data source, median realized episode length, and information gap.} Lengths and gaps are in environment steps. Every variant contributes 250 retained demonstrations, and their oracle success rate is 1.000. Each horizon's split is given by Table~\ref{tab:splits}.}
\begin{tabular}{@{}lrlrl@{}}
\toprule
\textbf{Env ID} & \shortstack[r]{\textbf{Horizon}\\(steps)} & \textbf{Source} & \shortstack[r]{\textbf{Median}\\length} & \textbf{Gap} \\
\midrule
\texttt{RememberColor3-VLA-v0} & 25 & PPO & 15 & 3 (2--4) \\
\texttt{RememberColor5-VLA-v0} & 25 & PPO & 15 & 3 (2--4) \\
\texttt{RememberColor9-VLA-v0} & 25 & PPO & 15 & 3 (2--4) \\
\texttt{RememberColor3-Long-VLA-v0} & 600 & MP & 337 & 250 (150--350) \\
\texttt{RememberColor5-Long-VLA-v0} & 600 & MP & 390 & 250 (150--350) \\
\texttt{RememberColor9-Long-VLA-v0} & 600 & MP & 378 & 250 (150--350) \\
\bottomrule
\end{tabular}
\end{table}

\paragraph{Information gap: provenance.}
Every gap in the table above is read from the environment source.
\begin{itemize}
\item \texttt{RememberColor3-VLA-v0} (gap 3 (2--4)), \texttt{RememberColor5-VLA-v0} (), \texttt{RememberColor9-VLA-v0} (): The manipulation phase begins after an empty phase whose length is drawn uniformly between 1 and 5 steps at reset, so the gap is that draw itself.
\item \texttt{RememberColor3-Long-VLA-v0} (gap 250 (150--350)), \texttt{RememberColor5-Long-VLA-v0} (), \texttt{RememberColor9-Long-VLA-v0} (): The manipulation phase begins after an empty phase whose length is drawn uniformly between 50 and 450 steps at reset, so the gap is that draw itself.
\end{itemize}

\paragraph{Success condition.}
The robot must reach the cube whose color matches the cue and satisfy the environment reach threshold.

\paragraph{Dense reward structure.}
Identical reward code to find\_imposter\_color: reach-to-goal shaping, a small static-robot bonus, and the smoothness penalty, with a fixed terminal bonus on success.

\subsection{Remember Shape}
\label{sec:family-remember-shape}

\paragraph{Motivation.}
The same sorting task by geometry instead of color, the way a machinist might be shown a reference bolt head and then asked to find its match among a mixed bin by feel and shape alone.

\begin{quote}\itshape
``Observe the object's shape, wait, then touch the object of the same shape.''
\end{quote}

\paragraph{Task.}
Delayed shape recall: observe one target shape, wait, and touch the object with the same shape.

\begin{figure}[H]
\centering
\includegraphics[width=0.95\linewidth]{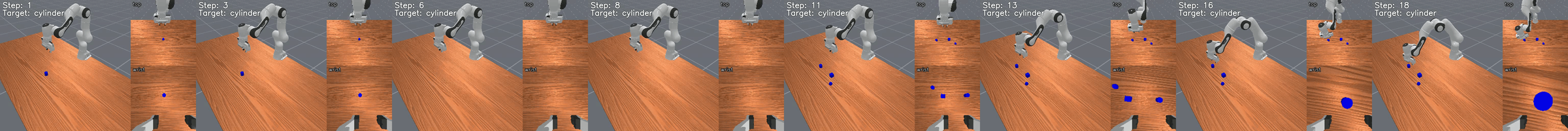}
\vspace{-8pt}
\caption{\textbf{Rollout filmstrip, \texttt{RememberShape3-VLA-v0} (a representative variant).}}
\end{figure}

\begin{table}[H]
\centering\small
\caption{\textbf{Remember Shape: variants, horizon, data source, median realized episode length, and information gap.} Lengths and gaps are in environment steps. Every variant contributes 250 retained demonstrations, and their oracle success rate is 1.000. Each horizon's split is given by Table~\ref{tab:splits}.}
\begin{tabular}{@{}lrlrl@{}}
\toprule
\textbf{Env ID} & \shortstack[r]{\textbf{Horizon}\\(steps)} & \textbf{Source} & \shortstack[r]{\textbf{Median}\\length} & \textbf{Gap} \\
\midrule
\texttt{RememberShape3-VLA-v0} & 25 & PPO & 15 & 3 (2--4) \\
\texttt{RememberShape5-VLA-v0} & 25 & PPO & 15 & 3 (2--4) \\
\texttt{RememberShape9-VLA-v0} & 25 & PPO & 15 & 3 (2--4) \\
\texttt{RememberShape3-Long-VLA-v0} & 600 & MP & 337 & 250 (150--350) \\
\texttt{RememberShape5-Long-VLA-v0} & 600 & MP & 349 & 250 (150--350) \\
\texttt{RememberShape9-Long-VLA-v0} & 600 & MP & 328 & 250 (150--350) \\
\bottomrule
\end{tabular}
\end{table}

\paragraph{Information gap: provenance.}
Every gap in the table above is read from the environment source.
\begin{itemize}
\item \texttt{RememberShape3-VLA-v0} (gap 3 (2--4)), \texttt{RememberShape5-VLA-v0} (), \texttt{RememberShape9-VLA-v0} (): The manipulation phase begins after an empty phase whose length is drawn uniformly between 1 and 5 steps at reset, so the gap is that draw itself.
\item \texttt{RememberShape3-Long-VLA-v0} (gap 250 (150--350)), \texttt{RememberShape5-Long-VLA-v0} (), \texttt{RememberShape9-Long-VLA-v0} (): The manipulation phase begins after an empty phase whose length is drawn uniformly between 50 and 450 steps at reset, so the gap is that draw itself.
\end{itemize}

\paragraph{Success condition.}
The robot must reach the object whose geometry matches the cue.

\paragraph{Dense reward structure.}
Identical reward code to find\_imposter\_color: reach-to-goal shaping, a small static-robot bonus, and the smoothness penalty, with a fixed terminal bonus on success.

\subsection{Remember Shape And Color}
\label{sec:family-remember-shape-and-color}

\paragraph{Motivation.}
The sorting task once more, now requiring both shape and color to match the item shown moments earlier -- a distractor sharing only one attribute is not close enough.

\begin{quote}\itshape
``Observe the object's shape and color, wait, then touch the object of the same shape and color.''
\end{quote}

\paragraph{Task.}
Delayed binding recall: remember both shape and color and select the matching object.

\begin{figure}[H]
\centering
\includegraphics[width=0.95\linewidth]{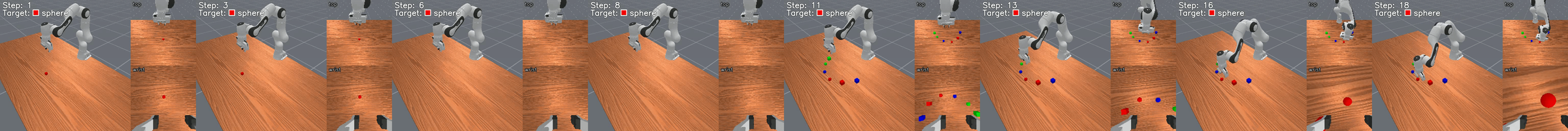}
\vspace{-8pt}
\caption{\textbf{Rollout filmstrip, \texttt{RememberShapeAndColor3x2-VLA-v0} (a representative variant).}}
\end{figure}

\begin{table}[H]
\centering\small
\caption{\textbf{Remember Shape And Color: variants, horizon, data source, median realized episode length, and information gap.} Lengths and gaps are in environment steps. Every variant contributes 250 retained demonstrations, and their oracle success rate is 1.000. Each horizon's split is given by Table~\ref{tab:splits}.}
\begin{tabular}{@{}lrlrl@{}}
\toprule
\textbf{Env ID} & \shortstack[r]{\textbf{Horizon}\\(steps)} & \textbf{Source} & \shortstack[r]{\textbf{Median}\\length} & \textbf{Gap} \\
\midrule
\texttt{RememberShapeAndColor3x2-VLA-v0} & 25 & PPO & 15 & 3 (2--4) \\
\texttt{RememberShapeAndColor3x3-VLA-v0} & 25 & PPO & 14 & 3 (2--4) \\
\texttt{RememberShapeAndColor5x3-VLA-v0} & 25 & PPO & 15 & 3 (2--4) \\
\texttt{RememberShapeAndColor3x2-Long-VLA-v0} & 600 & MP & 312 & 250 (150--350) \\
\texttt{RememberShapeAndColor3x3-Long-VLA-v0} & 600 & MP & 321 & 250 (150--350) \\
\texttt{RememberShapeAndColor5x3-Long-VLA-v0} & 600 & MP & 322 & 250 (150--350) \\
\bottomrule
\end{tabular}
\end{table}

\paragraph{Information gap: provenance.}
Every gap in the table above is read from the environment source.
\begin{itemize}
\item \texttt{RememberShapeAndColor3x2-VLA-v0} (gap 3 (2--4)), \texttt{RememberShapeAndColor3x3-VLA-v0} (), \texttt{RememberShapeAndColor5x3-VLA-v0} (): The manipulation phase begins after an empty phase whose length is drawn uniformly between 1 and 5 steps at reset, so the gap is that draw itself.
\item \texttt{RememberShapeAndColor3x2-Long-VLA-v0} (gap 250 (150--350)), \texttt{RememberShapeAndColor3x3-Long-VLA-v0} (), \texttt{RememberShapeAndColor5x3-Long-VLA-v0} (): The manipulation phase begins after an empty phase whose length is drawn uniformly between 50 and 450 steps at reset, so the gap is that draw itself.
\end{itemize}

\paragraph{Success condition.}
The robot must reach the object whose shape and color both match the cue.

\paragraph{Dense reward structure.}
Identical reward code to find\_imposter\_color: reach-to-goal shaping, a small static-robot bonus, and the smoothness penalty, with a fixed terminal bonus on success.

\subsection{Rotate Lenient}
\label{sec:family-rotate-lenient}

\paragraph{Motivation.}
A technician is told to turn a dial or connector to roughly a marked angle from memory, the way a valve gets set to about a quarter turn without a protractor in hand.

\begin{quote}\itshape
``Rotate the peg by \{angle\_deg\} degrees to match the target angle.''
\end{quote}

The braces are a placeholder: the environment substitutes that episode's own target angle, as the provenance below records.

\paragraph{Task.}
Spatial transformation: rotate a peg by a requested angle with a lenient center-position criterion.

\begin{figure}[H]
\centering
\includegraphics[width=0.95\linewidth]{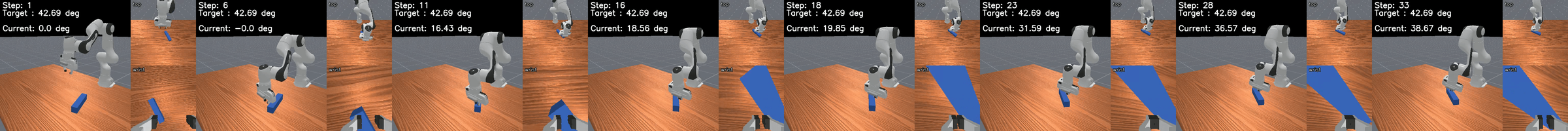}
\vspace{-8pt}
\caption{\textbf{Rollout filmstrip, \texttt{RotateLenientPos-VLA-v0} (a representative variant).}}
\end{figure}

\begin{table}[H]
\centering\small
\caption{\textbf{Rotate Lenient: variants, horizon, data source, median realized episode length, and information gap.} Lengths and gaps are in environment steps. Every variant contributes 250 retained demonstrations, and their oracle success rate is 1.000. Each horizon's split is given by Table~\ref{tab:splits}.}
\begin{tabular}{@{}lrlrl@{}}
\toprule
\textbf{Env ID} & \shortstack[r]{\textbf{Horizon}\\(steps)} & \textbf{Source} & \shortstack[r]{\textbf{Median}\\length} & \textbf{Gap} \\
\midrule
\texttt{RotateLenientPos-VLA-v0} & 60 & PPO & 29 & cue never hidden \\
\texttt{RotateLenientPosNeg-VLA-v0} & 60 & PPO & 19 & cue never hidden \\
\bottomrule
\end{tabular}
\end{table}

\paragraph{Information gap: provenance.}
Every gap in the table above is read from the environment source.
\begin{itemize}
\item \texttt{RotateLenientPos-VLA-v0} (gap cue never hidden), \texttt{RotateLenientPosNeg-VLA-v0} (): The target angle is set once at reset and never hidden. The language instruction is recomputed from the target angle at every step, so the target is available continuously via the prompt rather than through a fading visual cue, and there is no cue-hide phase to measure a retention gap over.
\end{itemize}

\paragraph{Success condition.}
The peg angle must be within \texttt{angle\_threshold} of the target and the robot must be static.

\paragraph{Dense reward structure.}
Exponential reach-to-peg shaping plus the smoothness penalty apply throughout. Once the tool center point is within a reach threshold, an exponential-in-angle-error term is added, and flat bonuses are added once the angle is within tolerance and again if the robot is also static at that point. A fixed terminal bonus replaces the shaped reward on success.

\subsection{Rotate Strict}
\label{sec:family-rotate-strict}

\paragraph{Motivation.}
The same rotation task, but now with a narrow tolerance -- like aligning a keyed connector that only seats correctly within a few degrees of the specified angle, where an approximate quarter turn is not good enough.

\begin{quote}\itshape
``Rotate the peg by \{angle\_deg\} degrees to match the target angle while keeping the center of the peg in place.''
\end{quote}

The braces are a placeholder: the environment substitutes that episode's own target angle, as the provenance below records.

\paragraph{Task.}
Spatial transformation: rotate a peg while keeping its center close to the original position.

\begin{figure}[H]
\centering
\includegraphics[width=0.95\linewidth]{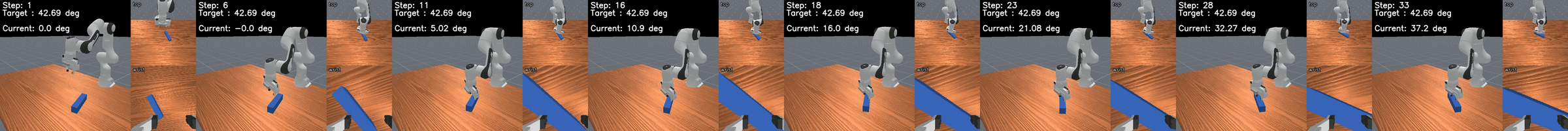}
\vspace{-8pt}
\caption{\textbf{Rollout filmstrip, \texttt{RotateStrictPos-VLA-v0} (a representative variant).}}
\end{figure}

\begin{table}[H]
\centering\small
\caption{\textbf{Rotate Strict: variants, horizon, data source, median realized episode length, and information gap.} Lengths and gaps are in environment steps. Every variant contributes 250 retained demonstrations, and their oracle success rate is 1.000. Each horizon's split is given by Table~\ref{tab:splits}.}
\begin{tabular}{@{}lrlrl@{}}
\toprule
\textbf{Env ID} & \shortstack[r]{\textbf{Horizon}\\(steps)} & \textbf{Source} & \shortstack[r]{\textbf{Median}\\length} & \textbf{Gap} \\
\midrule
\texttt{RotateStrictPos-VLA-v0} & 90 & PPO & 31 & cue never hidden \\
\texttt{RotateStrictPosNeg-VLA-v0} & 90 & PPO & 22 & cue never hidden \\
\bottomrule
\end{tabular}
\end{table}

\paragraph{Information gap: provenance.}
Every gap in the table above is read from the environment source.
\begin{itemize}
\item \texttt{RotateStrictPos-VLA-v0} (gap cue never hidden), \texttt{RotateStrictPosNeg-VLA-v0} (): The target angle is set once at reset and never hidden. The language instruction is recomputed from the target angle at every step, so the target is available continuously via the prompt rather than through a fading visual cue, and there is no cue-hide phase to measure a retention gap over.
\end{itemize}

\paragraph{Success condition.}
The peg angle must be within \texttt{angle\_threshold}, the peg must stay close to its initial XY position, and the robot must be static.

\paragraph{Dense reward structure.}
The same reach-then-angle shaping as rotate\_lenient, but gated more strictly: once reached, x/y position-holding terms are added (rewarding keeping the peg's xy position near its start), and the large angle-matching bonus and the two flat tolerance/static bonuses only fire once the peg is both reached *and* within xy position tolerance -- reflecting that this variant additionally requires the peg to stay in place while rotating. A larger fixed terminal bonus replaces the shaped reward on success.

\subsection{Seq Of Colors}
\label{sec:family-seq-of-colors}

\paragraph{Motivation.}
A worker observes a sequence of color-coded items and must later reproduce the same order while avoiding distractors. This requires retaining both the members of the set and their order.

\begin{quote}\itshape
``Observe which colored cubes appear during the cue, wait, then touch all of them in any order and press the center button.''
\end{quote}

\paragraph{Task.}
Sequential cue, unordered execution: colors are shown sequentially, but the remembered targets can be touched in any order.

\begin{figure}[H]
\centering
\includegraphics[width=0.95\linewidth]{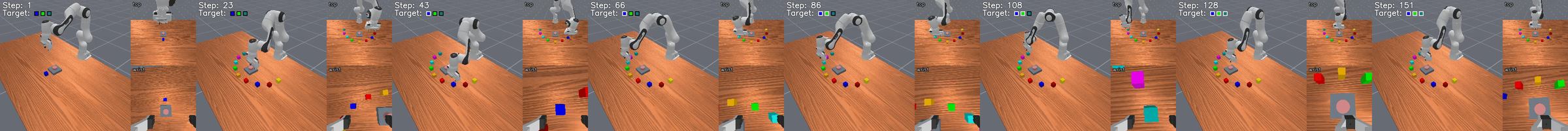}
\vspace{-8pt}
\caption{\textbf{Rollout filmstrip, \texttt{SeqOfColors3-VLA-v0} (a representative variant).}}
\end{figure}

\begin{table}[H]
\centering\small
\caption{\textbf{Seq Of Colors: variants, horizon, data source, median realized episode length, and information gap.} Lengths and gaps are in environment steps. Every variant contributes 250 retained demonstrations, and their oracle success rate is 1.000. Each horizon's split is given by Table~\ref{tab:splits}.}
\begin{tabular}{@{}lrlrl@{}}
\toprule
\textbf{Env ID} & \shortstack[r]{\textbf{Horizon}\\(steps)} & \textbf{Source} & \shortstack[r]{\textbf{Median}\\length} & \textbf{Gap} \\
\midrule
\texttt{SeqOfColors3-VLA-v0} & 400 & MP & 165 & 3 (2--4) \\
\texttt{SeqOfColors5-VLA-v0} & 400 & MP & 253 & 3 (2--4) \\
\texttt{SeqOfColors7-VLA-v0} & 400 & MP & 342 & 3 (2--4) \\
\texttt{SeqOfColors3-Long-VLA-v0} & 800 & MP & 559 & 225 (138--313) \\
\texttt{SeqOfColors5-Long-VLA-v0} & 1000 & MP & 738 & 225 (138--313) \\
\texttt{SeqOfColors7-Long-VLA-v0} & 1200 & MP & 916 & 225 (138--313) \\
\bottomrule
\end{tabular}
\end{table}

\paragraph{Information gap: provenance.}
Every gap in the table above is read from the environment source.
\begin{itemize}
\item \texttt{SeqOfColors3-VLA-v0} (gap 3 (2--4)), \texttt{SeqOfColors5-VLA-v0} (), \texttt{SeqOfColors7-VLA-v0} (): The manipulation phase begins after an empty phase whose length is drawn uniformly between 1 and 5 steps at reset, so the gap is that draw itself.
\item \texttt{SeqOfColors3-Long-VLA-v0} (gap 225 (138--313)), \texttt{SeqOfColors5-Long-VLA-v0} (), \texttt{SeqOfColors7-Long-VLA-v0} (): The manipulation phase begins after an empty phase whose length is drawn uniformly between 50 and 400 steps at reset, so the gap is that draw itself.
\end{itemize}

\paragraph{Success condition.}
The robot must touch every color that appeared in the cue sequence and avoid touching colors that were not part of that sequence.
The docstring describes set membership only. The implemented success condition additionally requires the touch order to match the cue order. The released instruction has the same mismatch, which Section \ref{sec:limitations} records.

\paragraph{Dense reward structure.}
The shaping is shared with bunch\_of\_colors: reach reward, a count-based term for touching cued colors, a static bonus, and penalties for wrong touches and cube displacement, all masked to the active phase. Terminal success receives a fixed bonus. This progress term is order-insensitive, whereas the executable success condition requires the cue order. The released instruction says that any order is allowed.

\subsection{Shell Game Color Lamp Touch}
\label{sec:family-shell-game-color-lamp-touch}

\paragraph{Motivation.}
A variant of the classic shell game where the winning cup is revealed by a lamp's color rather than which cup moved -- the player has to remember a color-to-cup mapping shown once, then act on it after the mapping is no longer displayed.

\begin{quote}\itshape
``Observe which color is under each cup, then touch the cup matching the lamp color.''
\end{quote}

\paragraph{Task.}
Hidden color-location binding: remember which color is under each cup and touch the cup matching the lamp color.

\begin{figure}[H]
\centering
\includegraphics[width=0.95\linewidth]{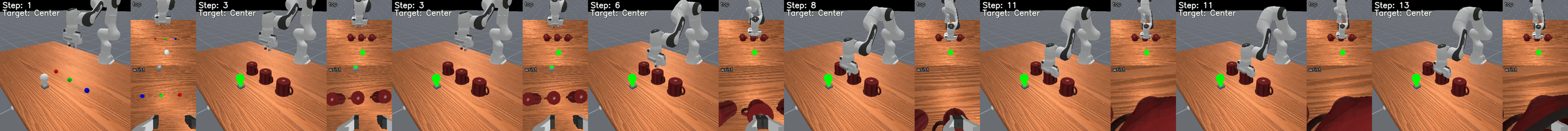}
\vspace{-8pt}
\caption{\textbf{Rollout filmstrip, \texttt{ShellGameColorLampTouch-VLA-v0} (a representative variant).}}
\end{figure}

\begin{table}[H]
\centering\small
\caption{\textbf{Shell Game Color Lamp Touch: variants, horizon, data source, median realized episode length, and information gap.} Lengths and gaps are in environment steps. Every variant contributes 250 retained demonstrations, and their oracle success rate is 1.000. Each horizon's split is given by Table~\ref{tab:splits}.}
\begin{tabular}{@{}lrlrl@{}}
\toprule
\textbf{Env ID} & \shortstack[r]{\textbf{Horizon}\\(steps)} & \textbf{Source} & \shortstack[r]{\textbf{Median}\\length} & \textbf{Gap} \\
\midrule
\texttt{ShellGameColorLampTouch-VLA-v0} & 30 & PPO & 15 & 0 \\
\bottomrule
\end{tabular}
\end{table}

\paragraph{Information gap: provenance.}
Every gap in the table above is read from the environment source.
\begin{itemize}
\item \texttt{ShellGameColorLampTouch-VLA-v0} (gap 0): The manipulation requirement begins on the very step the cue phase ends, with no intervening empty phase: a 0-step gap.
\end{itemize}

\paragraph{Success condition.}
The robot must touch the cup covering the ball whose color matches the lamp, and the robot must be static.

\paragraph{Dense reward structure.}
The same reach-plus-static shaping as shell\_game\_touch, but without a mug-shift penalty term -- the source cups are not perturbed after the cue in this variant. Masked to the manipulation phase, with a fixed terminal bonus on success.

\subsection{Shell Game Push}
\label{sec:family-shell-game-push}

\paragraph{Motivation.}
The classic shell game, but the found cup must be pushed aside rather than touched -- the same memory of which cup hides the target, expressed through a different action.

\begin{quote}\itshape
``Observe which cup hides the ball, wait, then push that cup forward.''
\end{quote}

\paragraph{Task.}
Hidden object selection: remember which cup hides the ball and push that cup forward.

\begin{figure}[H]
\centering
\includegraphics[width=0.95\linewidth]{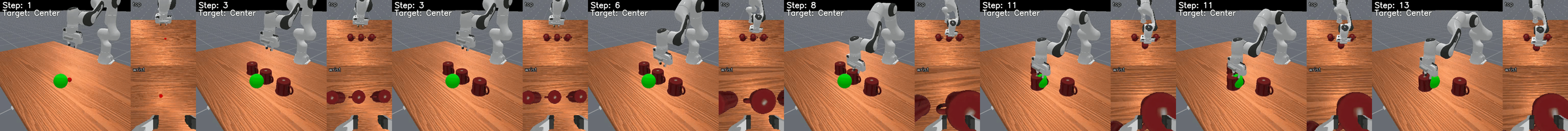}
\vspace{-8pt}
\caption{\textbf{Rollout filmstrip, \texttt{ShellGamePush-VLA-v0} (a representative variant).}}
\end{figure}

\begin{table}[H]
\centering\small
\caption{\textbf{Shell Game Push: variants, horizon, data source, median realized episode length, and information gap.} Lengths and gaps are in environment steps. Every variant contributes 250 retained demonstrations, and their oracle success rate is 1.000. Each horizon's split is given by Table~\ref{tab:splits}.}
\begin{tabular}{@{}lrlrl@{}}
\toprule
\textbf{Env ID} & \shortstack[r]{\textbf{Horizon}\\(steps)} & \textbf{Source} & \shortstack[r]{\textbf{Median}\\length} & \textbf{Gap} \\
\midrule
\texttt{ShellGamePush-VLA-v0} & 30 & PPO & 15 & 0 \\
\bottomrule
\end{tabular}
\end{table}

\paragraph{Information gap: provenance.}
Every gap in the table above is read from the environment source.
\begin{itemize}
\item \texttt{ShellGamePush-VLA-v0} (gap 0): The touch/push requirement begins on the very step the cue (ball position) is hidden, with no intervening empty phase: a 0-step gap.
\end{itemize}

\paragraph{Success condition.}
The ball center must finish inside the goal region.

\paragraph{Dense reward structure.}
Reach-to-ball shaping throughout, plus a place-toward-goal term that only activates once the tool center point has reached the ball, plus small static-robot bonuses once the ball is placed. Fixed terminal bonus on success. No explicit penalty term for disturbing the other cups.

\subsection{Shell Game Shuffle Color Lamp Touch}
\label{sec:family-shell-game-shuffle-color-lamp-touch}

\paragraph{Motivation.}
The color-lamp shell game with the added twist that the cups are physically shuffled after the mapping is shown, so the player has to track a moving target rather than recall a single static fact.

\begin{quote}\itshape
``Observe which color is under each cup, track the cups as they shuffle, then touch the cup matching the lamp color.''
\end{quote}

\paragraph{Task.}
Hidden color-location tracking: track cup-color associations through shuffling and touch the lamp-matching cup.

\begin{figure}[H]
\centering
\includegraphics[width=0.95\linewidth]{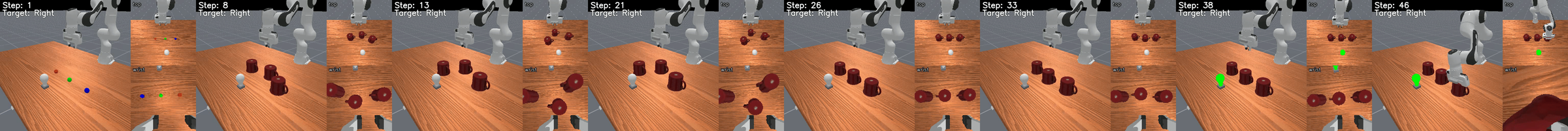}
\vspace{-8pt}
\caption{\textbf{Rollout filmstrip, \texttt{ShellGameShuffleColorLampTouch-VLA-v0} (a representative variant).}}
\end{figure}

\begin{table}[H]
\centering\small
\caption{\textbf{Shell Game Shuffle Color Lamp Touch: variants, horizon, data source, median realized episode length, and information gap.} Lengths and gaps are in environment steps. Every variant contributes 250 retained demonstrations, and their oracle success rate is 1.000. Each horizon's split is given by Table~\ref{tab:splits}.}
\begin{tabular}{@{}lrlrl@{}}
\toprule
\textbf{Env ID} & \shortstack[r]{\textbf{Horizon}\\(steps)} & \textbf{Source} & \shortstack[r]{\textbf{Median}\\length} & \textbf{Gap} \\
\midrule
\texttt{ShellGameShuffleColorLampTouch-VLA-v0} & 60 & PPO & 44 & 28 (24--31) \\
\texttt{ShellGameShuffleColorLampTouch-Long-VLA-v0} & 600 & MP & 331 & 250 (175--325) \\
\bottomrule
\end{tabular}
\end{table}

\paragraph{Information gap: provenance.}
Every gap in the table above is read from the environment source.
\begin{itemize}
\item \texttt{ShellGameShuffleColorLampTouch-VLA-v0} (gap 28 (24--31)): The manipulation phase begins after a shuffle phase whose length is drawn uniformly between 20 and 35 steps at reset.
\item \texttt{ShellGameShuffleColorLampTouch-Long-VLA-v0} (gap 250 (175--325)): The manipulation phase begins after a shuffle phase whose length is drawn uniformly between 100 and 400 steps at reset.
\end{itemize}

\paragraph{Success condition.}
The robot must touch the cup that hides the ball whose color matches the final lamp cue.

\paragraph{Dense reward structure.}
Identical reward code to shell\_game\_color\_lamp\_touch: reach-plus-static shaping with no mug-shift penalty, masked to the manipulation phase, fixed terminal bonus on success.

\subsection{Shell Game Shuffle Touch}
\label{sec:family-shell-game-shuffle-touch}

\paragraph{Motivation.}
The classic shell game played the way it is actually hustled on a street corner: the cups are shuffled after the ball is hidden, so the player has to visually track the target through the shuffle rather than remember its starting position.

\begin{quote}\itshape
``Observe which cup hides the ball, track the cups as they shuffle, then touch the correct cup.''
\end{quote}

\paragraph{Task.}
Hidden object tracking: track the ball-hiding cup through shuffles and touch it.

\begin{figure}[H]
\centering
\includegraphics[width=0.95\linewidth]{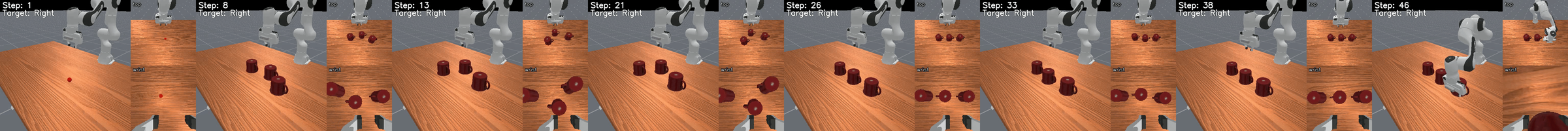}
\vspace{-8pt}
\caption{\textbf{Rollout filmstrip, \texttt{ShellGameShuffleTouch-VLA-v0} (a representative variant).}}
\end{figure}

\begin{table}[H]
\centering\small
\caption{\textbf{Shell Game Shuffle Touch: variants, horizon, data source, median realized episode length, and information gap.} Lengths and gaps are in environment steps. Every variant contributes 250 retained demonstrations, and their oracle success rate is 1.000. Each horizon's split is given by Table~\ref{tab:splits}.}
\begin{tabular}{@{}lrlrl@{}}
\toprule
\textbf{Env ID} & \shortstack[r]{\textbf{Horizon}\\(steps)} & \textbf{Source} & \shortstack[r]{\textbf{Median}\\length} & \textbf{Gap} \\
\midrule
\texttt{ShellGameShuffleTouch-VLA-v0} & 60 & PPO & 42 & 28 (24--31) \\
\texttt{ShellGameShuffleTouch-Long-VLA-v0} & 600 & MP & 337 & 250 (175--325) \\
\bottomrule
\end{tabular}
\end{table}

\paragraph{Information gap: provenance.}
Every gap in the table above is read from the environment source.
\begin{itemize}
\item \texttt{ShellGameShuffleTouch-VLA-v0} (gap 28 (24--31)): The manipulation phase begins after a shuffle phase whose length is drawn uniformly between 20 and 35 steps at reset.
\item \texttt{ShellGameShuffleTouch-Long-VLA-v0} (gap 250 (175--325)): The manipulation phase begins after a shuffle phase whose length is drawn uniformly between 100 and 400 steps at reset.
\end{itemize}

\paragraph{Success condition.}
The robot must touch the final cup position that contains the hidden ball.

\paragraph{Dense reward structure.}
Identical reward code to shell\_game\_touch, including the mug-shift penalty on the largest displacement across all three cups (touched cup included, not excluded). The only functional difference from shell\_game\_touch is the environment's own shuffle phase before manipulation begins, not the reward shaping itself.

\subsection{Shell Game Touch}
\label{sec:family-shell-game-touch}

\paragraph{Motivation.}
The simplest form of the shell game: the target is hidden once, the cups do not move afterward, and the player only has to remember where it went rather than track it moving.

\begin{quote}\itshape
``Observe which cup hides the ball, wait, then touch that cup.''
\end{quote}

\paragraph{Task.}
Short hidden object memory: observe which cup hides the ball, wait, and touch that cup.

\begin{figure}[H]
\centering
\includegraphics[width=0.95\linewidth]{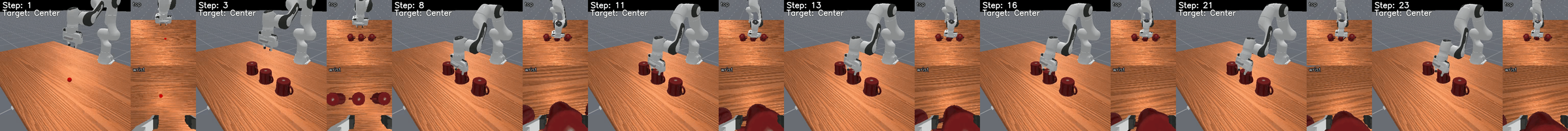}
\vspace{-8pt}
\caption{\textbf{Rollout filmstrip, \texttt{ShellGameTouch-VLA-v0} (a representative variant).}}
\end{figure}

\begin{table}[H]
\centering\small
\caption{\textbf{Shell Game Touch: variants, horizon, data source, median realized episode length, and information gap.} Lengths and gaps are in environment steps. Every variant contributes 250 retained demonstrations, and their oracle success rate is 1.000. Each horizon's split is given by Table~\ref{tab:splits}.}
\begin{tabular}{@{}lrlrl@{}}
\toprule
\textbf{Env ID} & \shortstack[r]{\textbf{Horizon}\\(steps)} & \textbf{Source} & \shortstack[r]{\textbf{Median}\\length} & \textbf{Gap} \\
\midrule
\texttt{ShellGameTouch-VLA-v0} & 30 & PPO & 22 & 0 \\
\bottomrule
\end{tabular}
\end{table}

\paragraph{Information gap: provenance.}
Every gap in the table above is read from the environment source.
\begin{itemize}
\item \texttt{ShellGameTouch-VLA-v0} (gap 0): The touch/push requirement begins on the very step the cue (ball position) is hidden, with no intervening empty phase: a 0-step gap.
\end{itemize}

\paragraph{Success condition.}
The robot must touch the cup covering the memorized ball location.

\paragraph{Dense reward structure.}
Reach-to-goal shaping plus small bonuses for being static once the touch is registered, minus a cup-displacement penalty (saturating with the largest displacement among all three cups -- left, center, and right -- with no exclusion for whichever cup is being touched) that discourages disturbing any cup during manipulation. Masked to the manipulation phase, with a fixed terminal bonus on success.

\subsection{Take It Back}
\label{sec:family-take-it-back}

\paragraph{Motivation.}
A librarian pulls a book from the shelf to check something, sets it down elsewhere, and afterward has to return it to its exact original slot -- not just any open space on the shelf, but the specific spot it came from.

\begin{quote}\itshape
``Push the cube onto the red target, and when the target changes color, return the cube to its original position.''
\end{quote}

\paragraph{Task.}
Spatial restoration: push a cube to a target, then return it to its original position after the target changes.

\begin{figure}[H]
\centering
\includegraphics[width=0.95\linewidth]{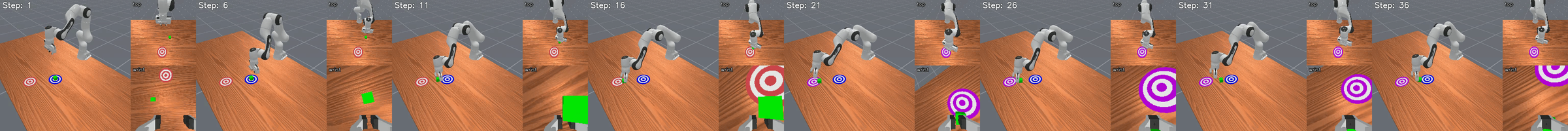}
\vspace{-8pt}
\caption{\textbf{Rollout filmstrip, \texttt{TakeItBack-VLA-v0} (a representative variant).}}
\end{figure}

\begin{table}[H]
\centering\small
\caption{\textbf{Take It Back: variants, horizon, data source, median realized episode length, and information gap.} Lengths and gaps are in environment steps. Every variant contributes 250 retained demonstrations, and their oracle success rate is 1.000. Each horizon's split is given by Table~\ref{tab:splits}.}
\begin{tabular}{@{}lrlrl@{}}
\toprule
\textbf{Env ID} & \shortstack[r]{\textbf{Horizon}\\(steps)} & \textbf{Source} & \shortstack[r]{\textbf{Median}\\length} & \textbf{Gap} \\
\midrule
\texttt{TakeItBack-VLA-v0} & 60 & PPO & 27 & undetermined \\
\bottomrule
\end{tabular}
\end{table}

\paragraph{Information gap: provenance.}
Every gap in the table above is read from the environment source.
\begin{itemize}
\item \texttt{TakeItBack-VLA-v0} (gap undetermined): The initial region is treated as a hidden object and never rendered, and the return stage begins only once the first goal is achieved -- an event timed by how fast the agent completes the first push, not a source constant, so the gap cannot be read off as a constant or a sampling range.
\end{itemize}

\paragraph{Success condition.}
The cube must first reach the goal region and then end up back inside the initial region within the same episode.

\paragraph{Dense reward structure.}
A two-stage push-then-pull sum switched by whether the cube has already reached the goal region: before that, the reward is dominated by reach-to-push-pose and place-in-goal shaping, with an added bonus once positioned to push. After that, it is dominated by reach-to-pull-pose and keep-contact shaping, with an added bonus once positioned to pull and place the cube back at its initial region. Minus the smoothness penalty throughout, fixed terminal bonus on success.

\subsection{Timed Transfer}
\label{sec:family-timed-transfer}

\paragraph{Motivation.}
A relay handoff on an assembly line has to happen inside a specific timing window signaled earlier -- handing off too early or too late both miss the window, the way a baton pass only counts inside a marked zone.

\begin{quote}\itshape
``When the white lamp turns green, start counting steps from that exact moment. Move the blue cube from the green disc to the red disc exactly on step 100 of that count.''
\end{quote}

\paragraph{Task.}
Delayed timing: after a lamp cue, move the cube at a precise future timestep.

\begin{figure}[H]
\centering
\includegraphics[width=0.95\linewidth]{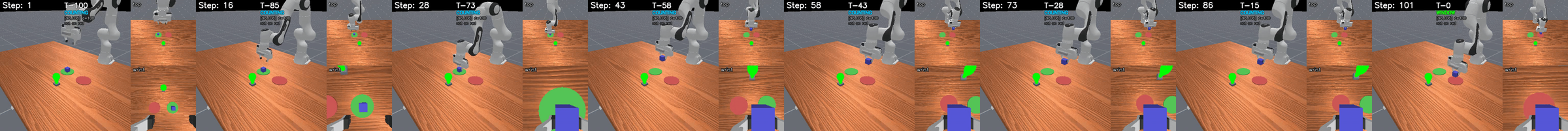}
\vspace{-8pt}
\caption{\textbf{Rollout filmstrip, \texttt{TimedTransferEasy-VLA-v0} (a representative variant).}}
\end{figure}

\begin{table}[H]
\centering\small
\caption{\textbf{Timed Transfer: variants, horizon, data source, median realized episode length, and information gap.} Lengths and gaps are in environment steps. Every variant contributes 250 retained demonstrations, and their oracle success rate is 1.000. Each horizon's split is given by Table~\ref{tab:splits}.}
\begin{tabular}{@{}lrlrl@{}}
\toprule
\textbf{Env ID} & \shortstack[r]{\textbf{Horizon}\\(steps)} & \textbf{Source} & \shortstack[r]{\textbf{Median}\\length} & \textbf{Gap} \\
\midrule
\texttt{TimedTransferEasy-VLA-v0} & 200 & MP & 106 & 100 \\
\texttt{TimedTransferMedium-VLA-v0} & 250 & MP & 152 & 150 \\
\texttt{TimedTransferHard-VLA-v0} & 300 & MP & 203 & 200 \\
\texttt{TimedTransferEasy-Long-VLA-v0} & 600 & MP & 298 & 300 \\
\texttt{TimedTransferMedium-Long-VLA-v0} & 900 & MP & 488 & 500 \\
\texttt{TimedTransferHard-Long-VLA-v0} & 1200 & MP & 962 & 1000 \\
\bottomrule
\end{tabular}
\end{table}

\paragraph{Information gap: provenance.}
Every gap in the table above is read from the environment source.
\begin{itemize}
\item \texttt{TimedTransferEasy-VLA-v0} (gap 100): The required delay is a fixed 100 steps (not resampled per episode). The cube must land on the red disc within a fixed tolerance of that many steps after the signal, so the required retention interval is exactly 100 steps.
\item \texttt{TimedTransferMedium-VLA-v0} (gap 150): The required delay is a fixed 150 steps (not resampled per episode). The cube must land on the red disc within a fixed tolerance of that many steps after the signal, so the required retention interval is exactly 150 steps.
\item \texttt{TimedTransferHard-VLA-v0} (gap 200): The required delay is a fixed 200 steps (not resampled per episode). The cube must land on the red disc within a fixed tolerance of that many steps after the signal, so the required retention interval is exactly 200 steps.
\item \texttt{TimedTransferEasy-Long-VLA-v0} (gap 300): The required delay is a fixed 300 steps (not resampled per episode). The cube must land on the red disc within a fixed tolerance of that many steps after the signal, so the required retention interval is exactly 300 steps.
\item \texttt{TimedTransferMedium-Long-VLA-v0} (gap 500): The required delay is a fixed 500 steps (not resampled per episode). The cube must land on the red disc within a fixed tolerance of that many steps after the signal, so the required retention interval is exactly 500 steps.
\item \texttt{TimedTransferHard-Long-VLA-v0} (gap 1000): The required delay is a fixed 1000 steps (not resampled per episode). The cube must land on the red disc within a fixed tolerance of that many steps after the signal, so the required retention interval is exactly 1000 steps.
\end{itemize}

\paragraph{Success condition (from the source docstring's episode-flow description).}
\begin{itemize}
\item The lamp is off.  The blue cube sits on the green disc.
\item After a brief random delay the lamp turns green (the signal).
\item The agent must internally count DELAY\_STEPS from the signal.
\item The agent picks the blue cube and places it on the red disc.
\item The cube must be on the red disc within +/-TOLERANCE\_FRAC of the target step.  Placing too early or too late is a failure.
\end{itemize}

\paragraph{Dense reward structure.}
A timing state machine gates the reward across waiting and transfer phases. Reward is zero before the cue lamp turns on. Afterward, reach and grasp bonuses accrue. Placement and static-hold bonuses are available only in the target step window, while early placement is penalized. A smoothness penalty applies throughout, with fixed failure and success terms. This family tests elapsed-time tracking, rather than retention of a hidden visual cue.

\subsection{Trace Shape}
\label{sec:family-trace-shape}

\paragraph{Motivation.}
A physical therapist demonstrates a movement pattern once and asks the patient to reproduce it without a live reference to copy -- the trajectory itself, not just its start and end points, has to be remembered.

\begin{quote}\itshape
``Watch the red cube trace a shape on the table. When the lamp turns green, pick up the green cube and trace exactly the same shape.''
\end{quote}

\paragraph{Task.}
Demonstration imitation: watch a cube trace a shape, then reproduce the path.

\begin{figure}[H]
\centering
\includegraphics[width=0.95\linewidth]{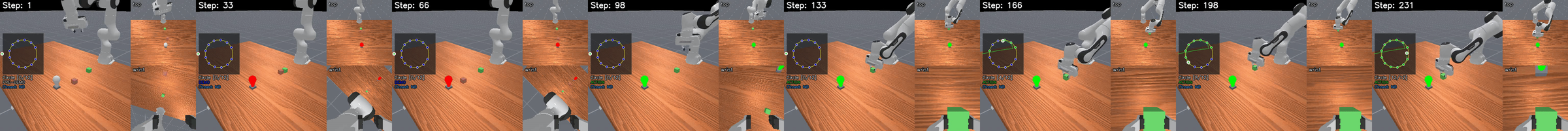}
\vspace{-8pt}
\caption{\textbf{Rollout filmstrip, \texttt{TraceShapeEasy-VLA-v0} (a representative variant).}}
\end{figure}

\begin{table}[H]
\centering\small
\caption{\textbf{Trace Shape: variants, horizon, data source, median realized episode length, and information gap.} Lengths and gaps are in environment steps. Every variant contributes 250 retained demonstrations, and their oracle success rate is 1.000. Each horizon's split is given by Table~\ref{tab:splits}.}
\begin{tabular}{@{}lrlrl@{}}
\toprule
\textbf{Env ID} & \shortstack[r]{\textbf{Horizon}\\(steps)} & \textbf{Source} & \shortstack[r]{\textbf{Median}\\length} & \textbf{Gap} \\
\midrule
\texttt{TraceShapeEasy-VLA-v0} & 250 & MP & 209 & 0 \\
\texttt{TraceShapeMedium-VLA-v0} & 300 & MP & 210 & 0 \\
\texttt{TraceShapeHard-VLA-v0} & 350 & MP & 207 & 0 \\
\bottomrule
\end{tabular}
\end{table}

\paragraph{Information gap: provenance.}
Every gap in the table above is read from the environment source.
\begin{itemize}
\item \texttt{TraceShapeEasy-VLA-v0} (gap 0), \texttt{TraceShapeMedium-VLA-v0} (), \texttt{TraceShapeHard-VLA-v0} (): The manipulation requirement begins as soon as the cue phase elapses, with no separate empty phase. The red-cube demo is hidden the same step the manipulation requirement starts: a 0-step gap.
\end{itemize}

\paragraph{Success condition.}
\begin{itemize}
\item The green cube must visit every checkpoint along the demonstrated path.
\item After that, the contour must be explicitly closed by returning near the first checkpoint (start point) within \texttt{CHECKPOINT\_THRESH}.
\end{itemize}

\paragraph{Dense reward structure.}
A checkpoint-tracing sum: a small reach term during the cue (watching the demonstration), then during the action phase, reach and grasp-indicator terms, a term proportional to the fraction of checkpoints visited so far, a nearest-unvisited-checkpoint distance term gated on already grasping the cube, and -- once every checkpoint has been visited -- a closure term that rewards returning near the starting checkpoint. Minus the smoothness penalty, fixed terminal bonus on success.

\subsection{Trace Shape Seq}
\label{sec:family-trace-shape-seq}

\paragraph{Motivation.}
The same demonstrated-motion task, now chaining several distinct shapes shown in sequence, so the reproduction has to preserve both each shape and the order they were shown in.

\begin{quote}\itshape
``Watch the red cube trace a sequence of shapes. When the lamp turns green, pick up the green cube and trace the same sequence in order. After finishing all shapes, press the button to submit your answer.''
\end{quote}

\paragraph{Task.}
Long-horizon demonstration imitation: watch and reproduce a sequence of traced shapes.

\begin{figure}[H]
\centering
\includegraphics[width=0.95\linewidth]{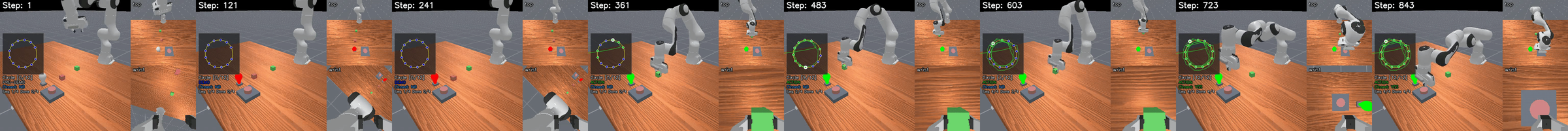}
\vspace{-8pt}
\caption{\textbf{Rollout filmstrip, \texttt{TraceShapeSeqEasy-VLA-v0} (a representative variant).}}
\end{figure}

\begin{table}[H]
\centering\small
\caption{\textbf{Trace Shape Seq: variants, horizon, data source, median realized episode length, and information gap.} Lengths and gaps are in environment steps. Every variant contributes 250 retained demonstrations, and their oracle success rate is 1.000. Each horizon's split is given by Table~\ref{tab:splits}.}
\begin{tabular}{@{}lrlrl@{}}
\toprule
\textbf{Env ID} & \shortstack[r]{\textbf{Horizon}\\(steps)} & \textbf{Source} & \shortstack[r]{\textbf{Median}\\length} & \textbf{Gap} \\
\midrule
\texttt{TraceShapeSeqEasy-VLA-v0} & 1500 & MP & 702 & 0 \\
\texttt{TraceShapeSeqMedium-VLA-v0} & 1500 & MP & 720 & 0 \\
\texttt{TraceShapeSeqHard-VLA-v0} & 1500 & MP & 706 & 0 \\
\bottomrule
\end{tabular}
\end{table}

\paragraph{Information gap: provenance.}
Every gap in the table above is read from the environment source.
\begin{itemize}
\item \texttt{TraceShapeSeqEasy-VLA-v0} (gap 0), \texttt{TraceShapeSeqMedium-VLA-v0} (), \texttt{TraceShapeSeqHard-VLA-v0} (): The manipulation requirement begins as soon as the cue phase elapses, with no separate empty phase. The red-cube demo is hidden the same step the manipulation requirement starts: a 0-step gap.
\end{itemize}

\paragraph{Success condition.}
\begin{itemize}
\item Every sequence element must be completed in order.
\item A sequence element is complete only when all its checkpoints are visited and the contour is closed (return near checkpoint[0]).
\item After all elements are complete, the robot must press the button.
\end{itemize}

\paragraph{Dense reward structure.}
Extends trace\_shape's checkpoint-tracing shaping (reach, grasp indicator, per-shape visit fraction, nearest-checkpoint term, closure term) with a sequence-progress term that credits completing the shapes of the sequence in order, and, once every shape in the sequence is closed, a reach-then-press term toward a submit button. A fixed penalty applies on an incorrect or otherwise failed submit, fixed terminal bonus on success.

\section{Full Per-Task Statistics}
\label{sec:appendix-all-tasks}

\begin{figure}[h]
  \centering
  \includegraphics[width=\linewidth,height=0.92\textheight,keepaspectratio]{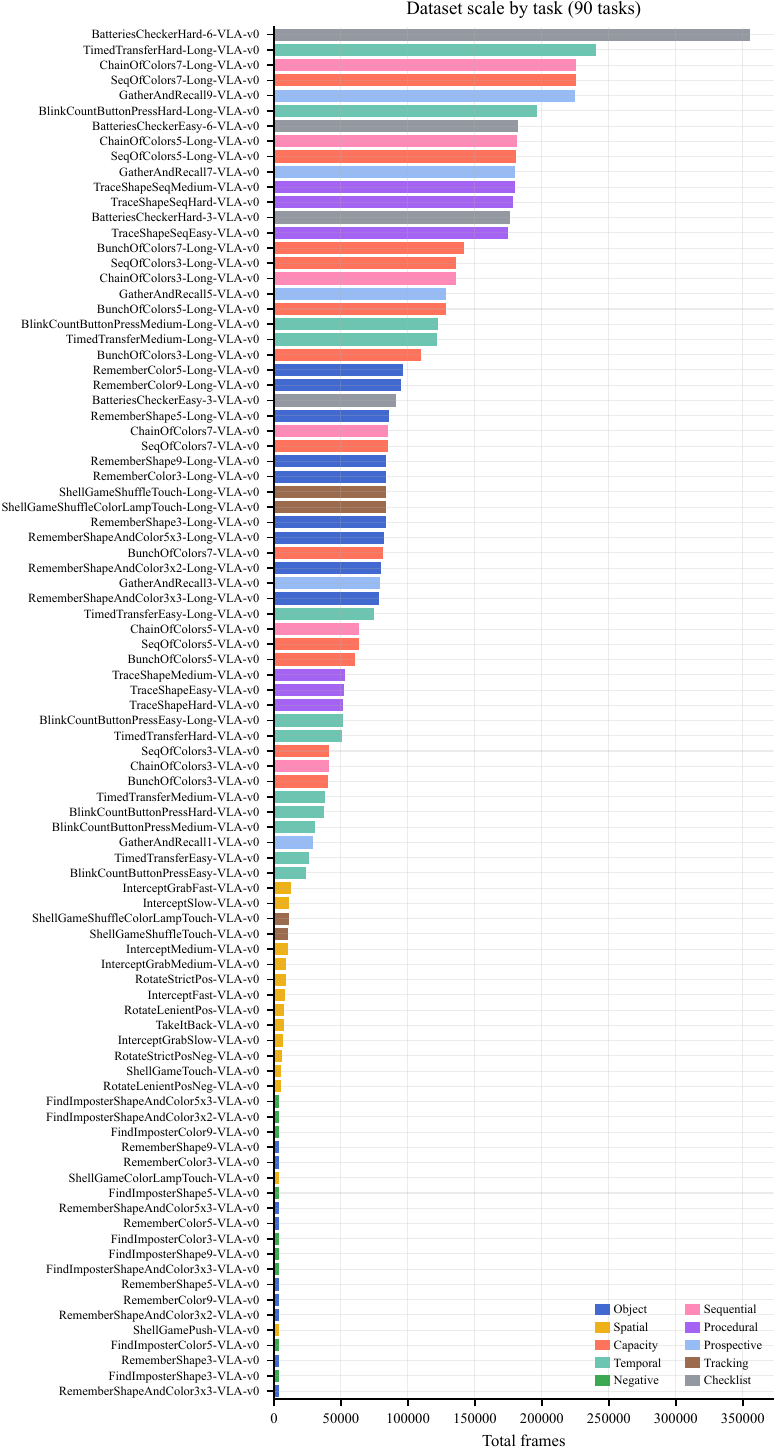}
  \vspace{-8pt}
  \caption{\textbf{Total released frames per task, sorted descending,
  colored by memory type.} The same color maps to the same memory type in
  every figure in this paper (Table~\ref{tab:taxonomy}). Numeric values for
  every task are in Table~\ref{tab:all-tasks} below.}
  \label{fig:scale}
\end{figure}

Table~\ref{tab:all-tasks} below lists all \NumTasks{} canonical
tasks.\footnote{The task registry also lists \NumExcludedTasks{}
\texttt{*-Medium-VLA-v0} environment ids beyond these \NumTasks{}. They
belong to a different project, not to MIKASA-Robo-VLA, and are excluded
from every count, table, and figure in this paper.} Figure~\ref{fig:scale} ranks those tasks by total
released frames.

\begin{landscape}
\small
\begin{longtable}{@{}lllrlrrrl@{}}
\caption{\textbf{Full per-task statistics for all \NumTasks{} canonical tasks.} Every task contributes 250 episodes. Lengths, horizons, and gaps are in environment steps. In the Gap column a bare number is a constant the environment sets, median (p25--p75) is that constant drawn afresh each episode, \emph{undetermined} marks an interval that ends when the agent reaches some state rather than after a stated number of steps, and \emph{cue never hidden} marks a task whose source states no interval exists.} \label{tab:all-tasks} \\
\toprule
\textbf{Env ID} & \textbf{Memory type} & \textbf{Split} & \textbf{Horizon} & \textbf{Source} & \textbf{Mean len.} & \textbf{Median len.} & \textbf{Oracle succ.} & \textbf{Gap} \\
\midrule
\endfirsthead
\toprule
\textbf{Env ID} & \textbf{Memory type} & \textbf{Split} & \textbf{Horizon} & \textbf{Source} & \textbf{Mean len.} & \textbf{Median len.} & \textbf{Oracle succ.} & \textbf{Gap} \\
\midrule
\endhead
\bottomrule
\endfoot
\texttt{ShellGameTouch-VLA-v0} & Spatial & Short & 30 & PPO & 21.5 & 22 & 1.000 & 0 \\
\texttt{ShellGamePush-VLA-v0} & Spatial & Short & 30 & PPO & 15.0 & 15 & 1.000 & 0 \\
\texttt{InterceptSlow-VLA-v0} & Spatial & Short & 60 & PPO & 43.8 & 43 & 1.000 & cue never hidden \\
\texttt{InterceptMedium-VLA-v0} & Spatial & Short & 60 & PPO & 41.8 & 41 & 1.000 & cue never hidden \\
\texttt{InterceptFast-VLA-v0} & Spatial & Short & 60 & PPO & 31.6 & 30 & 1.000 & cue never hidden \\
\texttt{InterceptGrabSlow-VLA-v0} & Spatial & Short & 60 & PPO & 27.6 & 24 & 1.000 & cue never hidden \\
\texttt{InterceptGrabMedium-VLA-v0} & Spatial & Short & 60 & PPO & 34.7 & 33 & 1.000 & cue never hidden \\
\texttt{InterceptGrabFast-VLA-v0} & Spatial & Short & 60 & PPO & 49.8 & 50 & 1.000 & cue never hidden \\
\texttt{RotateLenientPos-VLA-v0} & Spatial & Short & 60 & PPO & 29.3 & 29 & 1.000 & cue never hidden \\
\texttt{RotateLenientPosNeg-VLA-v0} & Spatial & Short & 60 & PPO & 19.8 & 19 & 1.000 & cue never hidden \\
\texttt{RotateStrictPos-VLA-v0} & Spatial & Short & 90 & PPO & 34.4 & 31 & 1.000 & cue never hidden \\
\texttt{RotateStrictPosNeg-VLA-v0} & Spatial & Short & 90 & PPO & 24.7 & 22 & 1.000 & cue never hidden \\
\texttt{TakeItBack-VLA-v0} & Spatial & Short & 60 & PPO & 28.9 & 27 & 1.000 & undetermined \\
\texttt{RememberColor3-VLA-v0} & Object & Short & 25 & PPO & 15.4 & 15 & 1.000 & 3 (2--4) \\
\texttt{RememberColor5-VLA-v0} & Object & Short & 25 & PPO & 15.3 & 15 & 1.000 & 3 (2--4) \\
\texttt{RememberColor9-VLA-v0} & Object & Short & 25 & PPO & 15.1 & 15 & 1.000 & 3 (2--4) \\
\texttt{RememberShape3-VLA-v0} & Object & Short & 25 & PPO & 14.9 & 15 & 1.000 & 3 (2--4) \\
\texttt{RememberShape5-VLA-v0} & Object & Short & 25 & PPO & 15.1 & 15 & 1.000 & 3 (2--4) \\
\texttt{RememberShape9-VLA-v0} & Object & Short & 25 & PPO & 15.4 & 15 & 1.000 & 3 (2--4) \\
\texttt{RememberShapeAndColor3x2-VLA-v0} & Object & Short & 25 & PPO & 15.0 & 15 & 1.000 & 3 (2--4) \\
\texttt{RememberShapeAndColor3x3-VLA-v0} & Object & Short & 25 & PPO & 14.3 & 14 & 1.000 & 3 (2--4) \\
\texttt{RememberShapeAndColor5x3-VLA-v0} & Object & Short & 25 & PPO & 15.4 & 15 & 1.000 & 3 (2--4) \\
\texttt{BunchOfColors3-VLA-v0} & Capacity & Medium & 400 & MP & 161.8 & 158 & 1.000 & 3 (2--4) \\
\texttt{BunchOfColors5-VLA-v0} & Capacity & Medium & 400 & MP & 241.4 & 242 & 1.000 & 3 (2--4) \\
\texttt{BunchOfColors7-VLA-v0} & Capacity & Medium & 400 & MP & 325.3 & 327 & 1.000 & 3 (2--4) \\
\texttt{SeqOfColors3-VLA-v0} & Capacity & Medium & 400 & MP & 165.6 & 165 & 1.000 & 3 (2--4) \\
\texttt{SeqOfColors5-VLA-v0} & Capacity & Medium & 400 & MP & 252.9 & 253 & 1.000 & 3 (2--4) \\
\texttt{SeqOfColors7-VLA-v0} & Capacity & Medium & 400 & MP & 341.3 & 342 & 1.000 & 3 (2--4) \\
\texttt{ChainOfColors3-VLA-v0} & Sequential & Medium & 400 & MP & 165.4 & 165 & 1.000 & 3 (2--4) \\
\texttt{ChainOfColors5-VLA-v0} & Sequential & Medium & 400 & MP & 253.1 & 253 & 1.000 & 3 (2--4) \\
\texttt{ChainOfColors7-VLA-v0} & Sequential & Medium & 400 & MP & 341.4 & 342 & 1.000 & 3 (2--4) \\
\texttt{ShellGameShuffleTouch-VLA-v0} & Tracking & Short & 60 & PPO & 42.1 & 42 & 1.000 & 28 (24--31) \\
\texttt{ShellGameShuffleColorLampTouch-VLA-v0} & Tracking & Short & 60 & PPO & 43.6 & 44 & 1.000 & 28 (24--31) \\
\texttt{ShellGameColorLampTouch-VLA-v0} & Spatial & Short & 30 & PPO & 15.4 & 15 & 1.000 & 0 \\
\texttt{FindImposterColor3-VLA-v0} & Negative & Short & 25 & PPO & 15.3 & 15 & 1.000 & 3 (2--4) \\
\texttt{FindImposterColor5-VLA-v0} & Negative & Short & 25 & PPO & 14.9 & 15 & 1.000 & 3 (2--4) \\
\texttt{FindImposterColor9-VLA-v0} & Negative & Short & 25 & PPO & 15.6 & 16 & 1.000 & 3 (2--4) \\
\texttt{FindImposterShape3-VLA-v0} & Negative & Short & 25 & PPO & 14.3 & 14 & 1.000 & 3 (2--4) \\
\texttt{FindImposterShape5-VLA-v0} & Negative & Short & 25 & PPO & 15.4 & 15 & 1.000 & 3 (2--4) \\
\texttt{FindImposterShape9-VLA-v0} & Negative & Short & 25 & PPO & 15.2 & 15 & 1.000 & 3 (2--4) \\
\texttt{FindImposterShapeAndColor3x2-VLA-v0} & Negative & Short & 25 & PPO & 15.7 & 16 & 1.000 & 3 (2--4) \\
\texttt{FindImposterShapeAndColor3x3-VLA-v0} & Negative & Short & 25 & PPO & 15.2 & 15 & 1.000 & 3 (2--4) \\
\texttt{FindImposterShapeAndColor5x3-VLA-v0} & Negative & Short & 40 & PPO & 15.7 & 15 & 1.000 & 3 (2--4) \\
\texttt{BatteriesCheckerEasy-3-VLA-v0} & Checklist & Medium & 540 & MP & 365.1 & 363 & 1.000 & undetermined \\
\texttt{BatteriesCheckerEasy-6-VLA-v0} & Checklist & Long & 1080 & MP & 728.4 & 727 & 1.000 & undetermined \\
\texttt{BatteriesCheckerHard-3-VLA-v0} & Checklist & Long & 1080 & MP & 705.6 & 699 & 1.000 & undetermined \\
\texttt{BatteriesCheckerHard-6-VLA-v0} & Checklist & Long & 2160 & MP & 1420.9 & 1416 & 1.000 & undetermined \\
\texttt{BlinkCountButtonPressEasy-VLA-v0} & Temporal & Short & 150 & MP & 94.6 & 94 & 1.000 & 0 \\
\texttt{BlinkCountButtonPressMedium-VLA-v0} & Temporal & Short & 200 & MP & 121.3 & 121 & 1.000 & 0 \\
\texttt{BlinkCountButtonPressHard-VLA-v0} & Temporal & Medium & 300 & MP & 150.3 & 153 & 1.000 & 0 \\
\texttt{RememberColor3-Long-VLA-v0} & Object & Medium & 600 & MP & 334.1 & 337 & 1.000 & 250 (150--350) \\
\texttt{RememberColor5-Long-VLA-v0} & Object & Medium & 600 & MP & 386.5 & 390 & 1.000 & 250 (150--350) \\
\texttt{RememberColor9-Long-VLA-v0} & Object & Medium & 600 & MP & 379.9 & 378 & 1.000 & 250 (150--350) \\
\texttt{RememberShape3-Long-VLA-v0} & Object & Medium & 600 & MP & 333.2 & 337 & 1.000 & 250 (150--350) \\
\texttt{RememberShape5-Long-VLA-v0} & Object & Medium & 600 & MP & 342.1 & 349 & 1.000 & 250 (150--350) \\
\texttt{RememberShape9-Long-VLA-v0} & Object & Medium & 600 & MP & 334.6 & 328 & 1.000 & 250 (150--350) \\
\texttt{RememberShapeAndColor3x2-Long-VLA-v0} & Object & Medium & 600 & MP & 319.4 & 312 & 1.000 & 250 (150--350) \\
\texttt{RememberShapeAndColor3x3-Long-VLA-v0} & Object & Medium & 600 & MP & 315.1 & 321 & 1.000 & 250 (150--350) \\
\texttt{RememberShapeAndColor5x3-Long-VLA-v0} & Object & Medium & 600 & MP & 329.4 & 322 & 1.000 & 250 (150--350) \\
\texttt{BunchOfColors3-Long-VLA-v0} & Capacity & Long & 700 & MP & 439.1 & 449 & 1.000 & 225 (138--313) \\
\texttt{BunchOfColors5-Long-VLA-v0} & Capacity & Long & 700 & MP & 513.9 & 524 & 1.000 & 225 (138--313) \\
\texttt{BunchOfColors7-Long-VLA-v0} & Capacity & Long & 700 & MP & 566.7 & 569 & 1.000 & 225 (138--313) \\
\texttt{SeqOfColors3-Long-VLA-v0} & Capacity & Long & 800 & MP & 543.6 & 559 & 1.000 & 225 (138--313) \\
\texttt{SeqOfColors5-Long-VLA-v0} & Capacity & Long & 1000 & MP & 724.0 & 738 & 1.000 & 225 (138--313) \\
\texttt{SeqOfColors7-Long-VLA-v0} & Capacity & Long & 1200 & MP & 901.5 & 916 & 1.000 & 225 (138--313) \\
\texttt{ChainOfColors3-Long-VLA-v0} & Sequential & Long & 800 & MP & 543.5 & 559 & 1.000 & 225 (138--313) \\
\texttt{ChainOfColors5-Long-VLA-v0} & Sequential & Long & 1000 & MP & 725.2 & 740 & 1.000 & 225 (138--313) \\
\texttt{ChainOfColors7-Long-VLA-v0} & Sequential & Long & 1200 & MP & 902.9 & 916 & 1.000 & 225 (138--313) \\
\texttt{ShellGameShuffleTouch-Long-VLA-v0} & Tracking & Medium & 600 & MP & 333.5 & 337 & 1.000 & 250 (175--325) \\
\texttt{ShellGameShuffleColorLampTouch-Long-VLA-v0} & Tracking & Medium & 600 & MP & 333.3 & 331 & 1.000 & 250 (175--325) \\
\texttt{BlinkCountButtonPressEasy-Long-VLA-v0} & Temporal & Long & 1200 & MP & 205.7 & 199 & 1.000 & 0 \\
\texttt{BlinkCountButtonPressMedium-Long-VLA-v0} & Temporal & Long & 1200 & MP & 491.2 & 490 & 1.000 & 0 \\
\texttt{BlinkCountButtonPressHard-Long-VLA-v0} & Temporal & Long & 1200 & MP & 784.9 & 788 & 1.000 & 0 \\
\texttt{TraceShapeEasy-VLA-v0} & Procedural & Medium & 250 & MP & 208.3 & 209 & 1.000 & 0 \\
\texttt{TraceShapeMedium-VLA-v0} & Procedural & Medium & 300 & MP & 210.7 & 210 & 1.000 & 0 \\
\texttt{TraceShapeHard-VLA-v0} & Procedural & Medium & 350 & MP & 207.5 & 207 & 1.000 & 0 \\
\texttt{TraceShapeSeqEasy-VLA-v0} & Procedural & Long & 1500 & MP & 698.6 & 702 & 1.000 & 0 \\
\texttt{TraceShapeSeqMedium-VLA-v0} & Procedural & Long & 1500 & MP & 718.6 & 720 & 1.000 & 0 \\
\texttt{TraceShapeSeqHard-VLA-v0} & Procedural & Long & 1500 & MP & 714.9 & 706 & 1.000 & 0 \\
\texttt{TimedTransferEasy-VLA-v0} & Temporal & Short & 200 & MP & 105.5 & 106 & 1.000 & 100 \\
\texttt{TimedTransferMedium-VLA-v0} & Temporal & Medium & 250 & MP & 152.3 & 152 & 1.000 & 150 \\
\texttt{TimedTransferHard-VLA-v0} & Temporal & Medium & 300 & MP & 203.2 & 203 & 1.000 & 200 \\
\texttt{TimedTransferEasy-Long-VLA-v0} & Temporal & Medium & 600 & MP & 298.1 & 298 & 1.000 & 300 \\
\texttt{TimedTransferMedium-Long-VLA-v0} & Temporal & Long & 900 & MP & 487.9 & 488 & 1.000 & 500 \\
\texttt{TimedTransferHard-Long-VLA-v0} & Temporal & Long & 1200 & MP & 961.6 & 962 & 1.000 & 1000 \\
\texttt{GatherAndRecall1-VLA-v0} & Prospective & Short & 200 & MP & 115.8 & 115 & 1.000 & undetermined \\
\texttt{GatherAndRecall3-VLA-v0} & Prospective & Medium & 400 & MP & 317.4 & 318 & 1.000 & undetermined \\
\texttt{GatherAndRecall5-VLA-v0} & Prospective & Medium & 600 & MP & 514.1 & 514 & 1.000 & undetermined \\
\texttt{GatherAndRecall7-VLA-v0} & Prospective & Long & 800 & MP & 720.9 & 721 & 1.000 & undetermined \\
\texttt{GatherAndRecall9-VLA-v0} & Prospective & Long & 1000 & MP & 898.1 & 900 & 1.000 & undetermined \\
\end{longtable}
\end{landscape}

\section{Reference Baseline: Full Configuration}
\label{sec:appendix-baseline}

Section~\ref{sec:baseline} gives the reference baseline's setup at the
level needed to interpret its results. This appendix records the available
training configuration and its provenance limits.

\begin{table}[h]
\centering\small
\caption{\textbf{Reference \(\pi_{0.5}\) baseline training configuration.}}
\begin{tabular}{ll}
\toprule
\textbf{Field} & \textbf{Value} \\
\midrule
Initialization & \texttt{pi05\_base} cross-embodiment checkpoint, fully fine-tuned \\
Codebase & \texttt{openpi-comet}~\citep{bai2025openpicomet}, a JAX/Flax fork of \texttt{openpi} \\
Training data & LeRobot format, \NumBaselineTasks{} tasks $\times$ \EpisodesPerTask{} episodes \\
& (\BaselineTrainEpisodes{} episodes, \BaselineTrainFrames{} frames) \\
Image input & Two cameras stacked, $\ImageShape{}$ each, upscaled to $\BaselineUpscale{}\times\BaselineUpscale{}$ \\
Action space & \texttt{pd\_ee\_delta\_pose}, \ActionDim{} active dims, padded to \BaselineActionDimPadded{} \\
Action chunk & \BaselineChunk{} steps, executed fully open-loop \\
Frame history & None -- policy sees only the current observation \\
Proprioception & Discretized, injected into the text prompt (\texttt{discrete\_state\_input=True}) \\
Global batch & \BaselineGlobalBatch{}, across \BaselineNumGpus{} GPUs (\texttt{fsdp=\BaselineNumGpus{}}) \\
Peak learning rate & Not recoverable from the source training project \\
Optimizer & Not recoverable from the source training project \\
Checkpoint evaluated & Step \BaselineSteps{} \\
\bottomrule
\end{tabular}
\end{table}

The training and evaluation lanes use distinct simulator seed ranges.
Training episodes and the \BaselineEpisodesPerTask{}-per-task evaluation
episodes (\texttt{base\_seed \BaselineEvalBaseSeed{}}) therefore use distinct
recorded seeds, although a seed alone does not establish that two simulator
versions produce distinct initial states.

\section{Comparison to Existing Benchmarks}
\label{sec:appendix-comparison}

Table~\ref{tab:comparison} compares dedicated simulation suites for
memory-dependent manipulation on the axes Section~\ref{sec:related}
discusses in prose: scale, controlled variation of memory difficulty, and
environment and learning support. Column definitions and the symbol legend
appear in the note below the table.

\begin{table*}[h]
\centering
\scriptsize
\setlength{\tabcolsep}{1.6pt}
\begin{tabular}{@{}lcrrrccccc@{}}
\toprule
& & \multicolumn{3}{c}{\textbf{Scale}}
& \multicolumn{3}{c}{\textbf{Controlled memory difficulty}}
& \multicolumn{2}{c}{\textbf{Environment \& learning}} \\
\cmidrule(lr){3-5}
\cmidrule(lr){6-8}
\cmidrule(lr){9-10}
\textbf{Benchmark} & \textbf{Backend} & \textbf{Tasks} & \textbf{Types} & \textbf{Demos}
& \textbf{Ret.} & \textbf{Load} & \textbf{H-strat.}
& \textbf{Dense rew.} & \textbf{GPU vec.} \\
\midrule
MemoryBench~\citep{fang2025sam2act} & RLBench / Coppelia & 3 & 2 & 300 & \xmark & \xmark & \xmark & \xmark & \xmark \\
LIBERO-Mem~\citep{chung2026liberomem} & LIBERO / MuJoCo & 10 & 4 & 1{,}000 & \pmark & \pmark & \pmark & \xmark & \xmark \\
RMBench~\citep{chen2026rmbench} & RoboTwin / SAPIEN & 9 & -- & 450 & \xmark & \pmark & \xmark & \xmark & \xmark \\
RoboMME~\citep{dai2026robomme} & ManiSkill / SAPIEN & 16 & 4 & 1{,}600 & \xmark & \xmark & \xmark & \xmark & \xmark \\
RoboMemArena~\citep{lei2026robomemarena} & LIBERO / MuJoCo & 26 & 4 & 2{,}600 & \xmark & \xmark & \xmark & \xmark & \xmark \\
RoboTwin-MeM~\citep{yang2026eventvla} & RoboTwin / SAPIEN & 8 & -- & 400\textsuperscript{\(\dagger\)} & \xmark & \cmark & \xmark & \xmark & \xmark \\
MEMOBench~\citep{sun2026memobench} & LIBERO / MuJoCo & 30 & 4 & 1{,}500 & \xmark & \xmark & \xmark & \xmark & \xmark \\
\midrule
\rowcolor{gray!10}
\textbf{MIKASA-Robo-VLA (ours)} & \textbf{ManiSkill / SAPIEN} & \textbf{\NumTasks{}} & \textbf{\NumMemoryTypes{}} & \textbf{\NumEpisodes{}} & \cmark & \cmark & \cmark & \cmark & \cmark \\
\bottomrule
\end{tabular}
\caption{\textbf{Comparison of dedicated simulation benchmarks for
memory-dependent robotic manipulation.}}
\label{tab:comparison}
\vspace{2pt}
\begin{minipage}{\textwidth}
\small
\textbf{Types} follows each benchmark's own taxonomy, not comparable across
suites. \textbf{Ret.}: controlled retention-interval variation.
\textbf{Load}: controlled variation of information amount or complexity.
\textbf{H-strat.}: canonical evaluation across horizon regimes.
\textbf{Dense rew.}: documented suite-wide dense rewards. \textbf{GPU
vec.}: a released benchmark-level API for GPU-parallel execution.
\cmark{}: explicit support. \pmark{}: partial or entangled support.
\xmark{}: not documented or provided suite-wide.
\textsuperscript{\(\dagger\)} RoboTwin-MeM's 50 episodes per task (of eight)
are shown for scale, not as an equivalent to other benchmarks' demonstrations.
\end{minipage}
\end{table*}

\section{Supplementary External Evaluations}
\label{sec:appendix-external}

This appendix collects three reference results that support
Section~\ref{sec:arch-comparison} and Section~\ref{sec:benchmark}'s remark
on customization. None of them is part of the canonical 90-task benchmark
evaluation. Each table's caption states its own protocol and episode count,
and the three must not be read against each other or against
Table~\ref{tab:baseline} as a single leaderboard.

\subsection{Memory architecture comparison}
\label{sec:appendix-arch-comparison}

Table~\ref{tab:arch-comparison} reproduces, at per-task resolution, the
comparison whose aggregates are quoted in Section~\ref{sec:arch-comparison}:
OpenVLA-OFT with and without a 64-token recurrent memory module ($\mu$VLA),
both fine-tuned on the same five tasks and evaluated zero-shot on 18 more,
as reported by \citet{cherepanov2026uvla}. The underlying evaluation ran on
2026-05-03 (no-memory arm) and 2026-06-02 ($\mu$VLA arm), under an earlier
release protocol that predates the wrapper and diagnostic-key differences
described in Section~\ref{sec:limitations}, so these numbers are not under
the same protocol this paper's own baseline uses.
\texttt{ShellGameTouch-VLA-v0}'s no-memory evaluation did not complete --
environment creation raised an error -- and is marked \textsc{n/a} rather
than a false zero.

\begin{table}[h]
\centering\scriptsize
\setlength{\tabcolsep}{4pt}
\caption{\textbf{Per-task success on the 23-task evaluation reported by \citet{cherepanov2026uvla}: OpenVLA-OFT without a memory module (exec-8) versus $\mu$VLA, the same backbone with a 64-token recurrent memory (receding horizon, memory updated every step).} 100 episodes per task, checkpoint step 150000. Trained tasks are marked \textsuperscript{$\dagger$}. The other 18 are zero-shot/held-out for both arms. Evaluated under an earlier release protocol (see Section~\ref{sec:limitations}). \textsc{n/a} marks an evaluation that did not complete and so is not scored.}
\label{tab:arch-comparison}
\begin{tabular}{lrr}
\toprule
\textbf{Task} & \textbf{No memory} & \textbf{$\mu$VLA} \\
\midrule
\texttt{InterceptFast} & 0.00 & 0.29 \\
\texttt{InterceptGrabFast} & 0.00 & 0.00 \\
\texttt{InterceptGrabMedium} & 0.00 & 0.00 \\
\texttt{InterceptGrabSlow} & 0.00 & 0.00 \\
\texttt{InterceptMedium\textsuperscript{$\dagger$}} & 0.36 & 0.55 \\
\texttt{InterceptSlow} & 0.05 & 0.08 \\
\texttt{RememberColor3} & 0.00 & 0.91 \\
\texttt{RememberColor5\textsuperscript{$\dagger$}} & 0.04 & 0.94 \\
\texttt{RememberColor9} & 0.03 & 0.38 \\
\texttt{RememberShape3} & 0.00 & 0.39 \\
\texttt{RememberShape5} & 0.00 & 0.42 \\
\texttt{RememberShape9} & 0.00 & 0.26 \\
\texttt{RememberShapeAndColor3x2} & 0.03 & 0.62 \\
\texttt{RememberShapeAndColor3x3\textsuperscript{$\dagger$}} & 0.03 & 0.84 \\
\texttt{RememberShapeAndColor5x3} & 0.01 & 0.27 \\
\texttt{RotateLenientPos} & 0.06 & 0.01 \\
\texttt{RotateLenientPosNeg} & 0.03 & 0.03 \\
\texttt{RotateStrictPos} & 0.00 & 0.00 \\
\texttt{RotateStrictPosNeg} & 0.00 & 0.00 \\
\texttt{ShellGamePick} & 0.00 & 0.00 \\
\texttt{ShellGamePush\textsuperscript{$\dagger$}} & 0.83 & 0.96 \\
\texttt{ShellGameTouch} & \textsc{n/a} & 0.00 \\
\texttt{TakeItBack\textsuperscript{$\dagger$}} & 0.83 & 0.98 \\
\midrule
Mean (5 trained tasks) & 0.418 & 0.854 \\
\bottomrule
\end{tabular}
\end{table}

\subsection{Execution-regime comparison}
\label{sec:appendix-exec-regime}

Table~\ref{tab:exec-regime} reports two independently trained stock
$\pi_{0.5}$ (LeRobot port, \texttt{lerobot/pi05\_base}) checkpoints on five
tasks, each evaluated twice: once under the open-loop 8-action chunk this
paper's own baseline uses, once under one-action-per-inference execution
with the same weights. Both checkpoints trained for 30{,}000 steps, global
batch 64, chunk size 8. Loader A shuffles episodes, and loader B preserves
episode order.

\begin{table}[h]
\centering\footnotesize
\setlength{\tabcolsep}{3pt}
\caption{\textbf{Reference $\pi_{0.5}$ (LeRobot port), two independently trained checkpoints, each evaluated under two execution regimes at inference: exec-8 (open-loop chunk of 8 actions) and exec-1 (receding horizon, one action per inference).} 100 episodes per task per cell. Not part of the canonical 90-task benchmark evaluation.}
\label{tab:exec-regime}
\begin{tabular}{llrrrrrr}
\toprule
\textbf{Loader} & \textbf{Regime} & \textbf{ShellGamePush} & \textbf{InterceptMedium} & \textbf{TakeItBack} & \textbf{RememberColor5} & \textbf{RSC3x3} & \textbf{Mean} \\
\midrule
A (shuffled) & exec-8 & 0.93 & 0.30 & 0.68 & 0.14 & 0.07 & 0.424 \\
A (shuffled) & exec-1 & 0.28 & 0.42 & 0.77 & 0.28 & 0.14 & 0.378 \\
B (episodic) & exec-8 & 0.95 & 0.31 & 0.77 & 0.18 & 0.03 & 0.448 \\
B (episodic) & exec-1 & 0.33 & 0.51 & 0.81 & 0.18 & 0.15 & 0.396 \\
\bottomrule
\end{tabular}
\end{table}

\subsection{Benchmark customization case study}
\label{sec:appendix-customization}

Table~\ref{tab:chunk-customization} illustrates that the benchmark's task
parameters are meant to be re-specified, not only consumed as released. Five
\texttt{*-Medium-VLA-v0} variants were created for a single test: $\mu$VLA with
its recurrent memory updated once per action chunk rather than every step,
compared across two truncated-backpropagation-through-time (TBPTT) window
lengths ($K=2$, 16 environment steps, and $K=8$, 64 steps) against a
retention interval of \ChunkHighlightDelayMin{}--\ChunkHighlightDelayMax{}
steps on \texttt{\ChunkHighlightTask{}-VLA-v0}. These five variants are
\emph{not} part of the canonical 90-task suite and are excluded from every
count, table, and figure elsewhere in this paper. They exist only as this
worked example. The window that covers the retention interval ($K=8$)
recovers \ChunkLongHighlightSuccess{} success on that task against
\ChunkShortHighlightSuccess{} for the window that does not ($K=2$), matching
the customization's own design intent. Neither arm's training run reached
its full step budget before being halted by the user, so the aggregate means
in the table are not a matched final comparison.

\begin{table}[h]
\centering\footnotesize
\caption{\textbf{Benchmark-customization case study: $\mu$VLA with memory updated once per 8-action chunk (full open loop), trained on 3 canonical tasks plus 5 custom \texttt{*-Medium-VLA-v0} variants created for this test and excluded from every count of the canonical 90-task suite.} $K$ is the TBPTT window in chunks (env-steps $=8K$). 50 episodes per task.}
\label{tab:chunk-customization}
\begin{tabular}{lrr}
\toprule
\textbf{Task} & \textbf{$K=2$ (16 steps, step 140{,}000)} & \textbf{$K=8$ (64 steps, best step 42{,}500)} \\
\midrule
\texttt{BunchOfColors3} & 0.00 & 0.00 \\
\texttt{ChainOfColors3} & 0.00 & 0.00 \\
\texttt{GatherAndRecall3} & 0.24 & 0.26 \\
\texttt{RememberColor3-Medium} & 0.32 & 0.52 \\
\texttt{RememberColor5-Medium} & 0.08 & 0.54 \\
\texttt{RememberShapeAndColor3x2-Medium} & 0.18 & 0.44 \\
\texttt{RememberShapeAndColor3x3-Medium} & 0.06 & 0.32 \\
\texttt{ShellGameColorLampTouch-Medium} & 0.30 & 0.32 \\
\midrule
Mean (8 tasks) & 0.147 & 0.300 \\
\bottomrule
\end{tabular}
\end{table}

\section{Rollout Contact Sheet}
\label{sec:appendix-contact-sheet}

\begin{figure}[htbp]
  \centering
  \includegraphics[width=\linewidth,height=0.92\textheight,keepaspectratio]{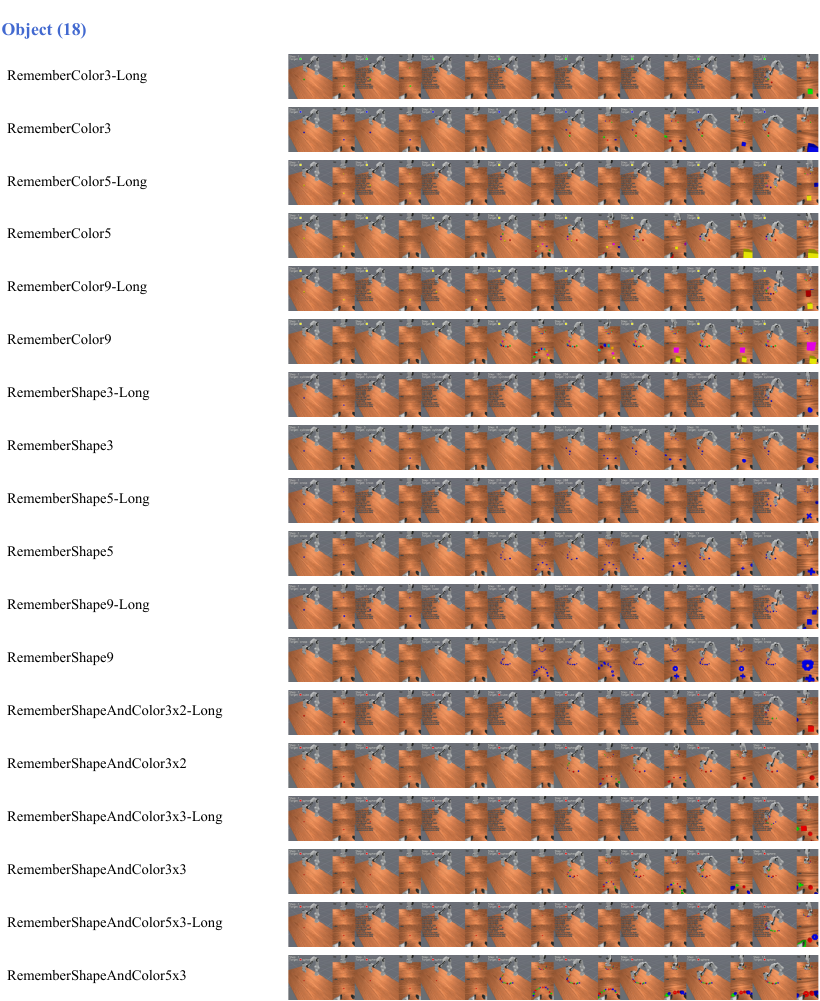}
  \vspace{-8pt}
  \caption{\textbf{Rollout filmstrips for the 18 Object tasks, one row per task, 8 evenly spaced frames per rollout, left to right.} The header color is this memory type's color throughout the paper (Table~\ref{tab:taxonomy}).}
  \label{fig:contact-sheet-object}
\end{figure}

\begin{figure}[htbp]
  \centering
  \includegraphics[width=\linewidth,height=0.92\textheight,keepaspectratio]{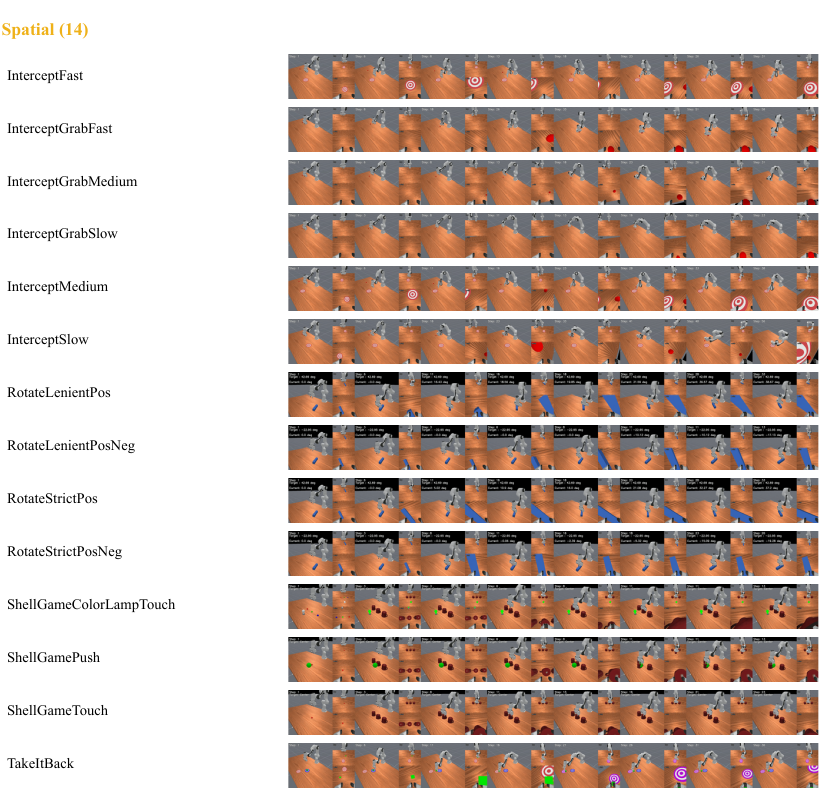}
  \vspace{-8pt}
  \caption{\textbf{Rollout filmstrips for the 14 Spatial tasks, one row per task, 8 evenly spaced frames per rollout, left to right.} The header color is this memory type's color throughout the paper (Table~\ref{tab:taxonomy}).}
  \label{fig:contact-sheet-spatial}
\end{figure}

\begin{figure}[htbp]
  \centering
  \includegraphics[width=\linewidth,height=0.92\textheight,keepaspectratio]{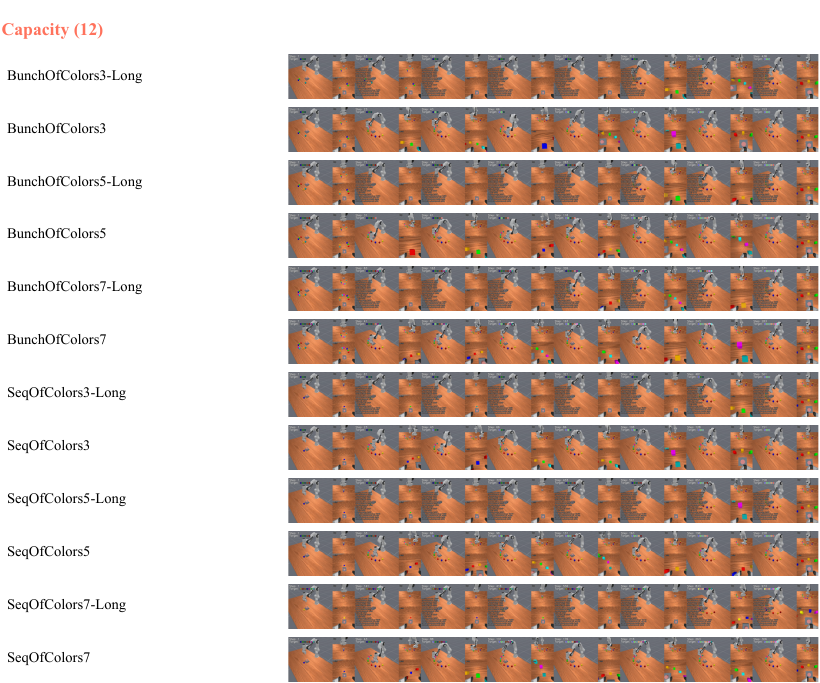}
  \vspace{-8pt}
  \caption{\textbf{Rollout filmstrips for the 12 Capacity tasks, one row per task, 8 evenly spaced frames per rollout, left to right.} The header color is this memory type's color throughout the paper (Table~\ref{tab:taxonomy}).}
  \label{fig:contact-sheet-capacity}
\end{figure}

\begin{figure}[htbp]
  \centering
  \includegraphics[width=\linewidth,height=0.92\textheight,keepaspectratio]{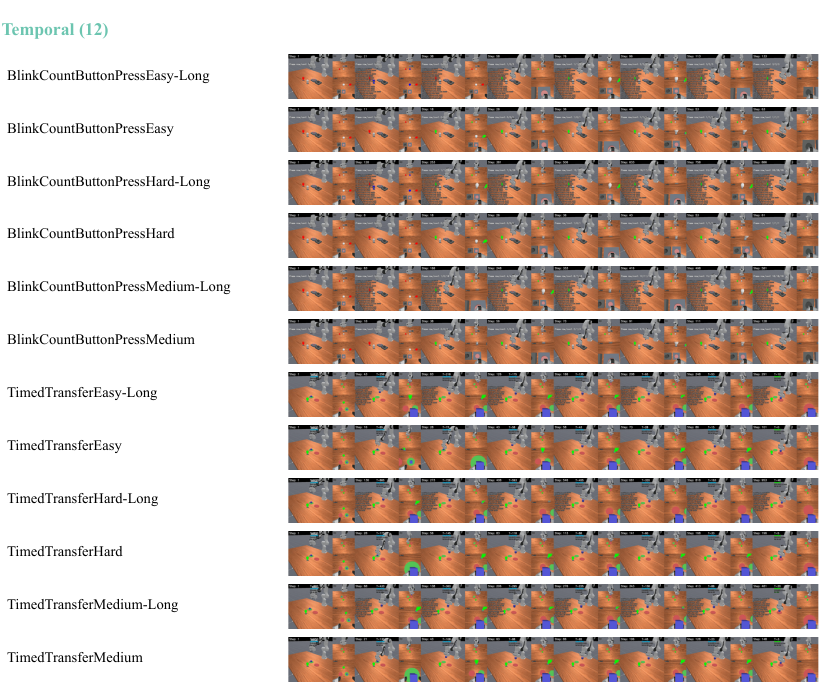}
  \vspace{-8pt}
  \caption{\textbf{Rollout filmstrips for the 12 Temporal tasks, one row per task, 8 evenly spaced frames per rollout, left to right.} The header color is this memory type's color throughout the paper (Table~\ref{tab:taxonomy}).}
  \label{fig:contact-sheet-temporal}
\end{figure}

\begin{figure}[htbp]
  \centering
  \includegraphics[width=\linewidth,height=0.92\textheight,keepaspectratio]{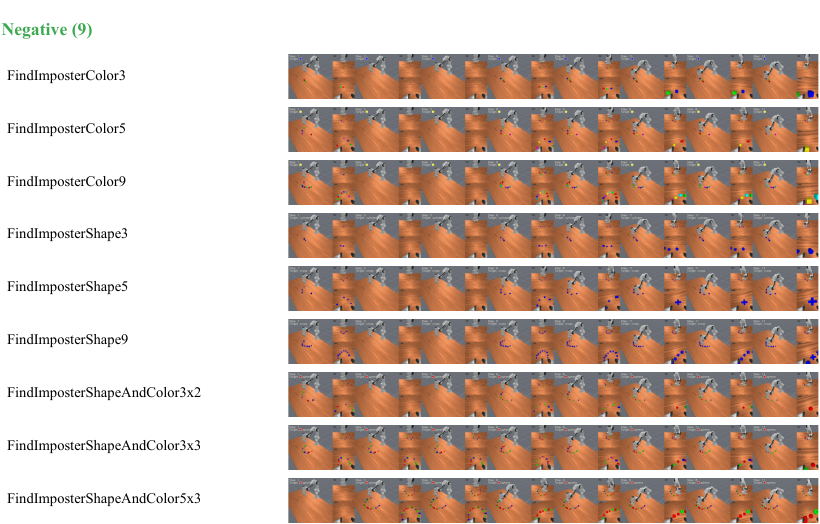}
  \vspace{-8pt}
  \caption{\textbf{Rollout filmstrips for the 9 Negative tasks, one row per task, 8 evenly spaced frames per rollout, left to right.} The header color is this memory type's color throughout the paper (Table~\ref{tab:taxonomy}).}
  \label{fig:contact-sheet-negative}
\end{figure}

\begin{figure}[htbp]
  \centering
  \includegraphics[width=\linewidth,height=0.92\textheight,keepaspectratio]{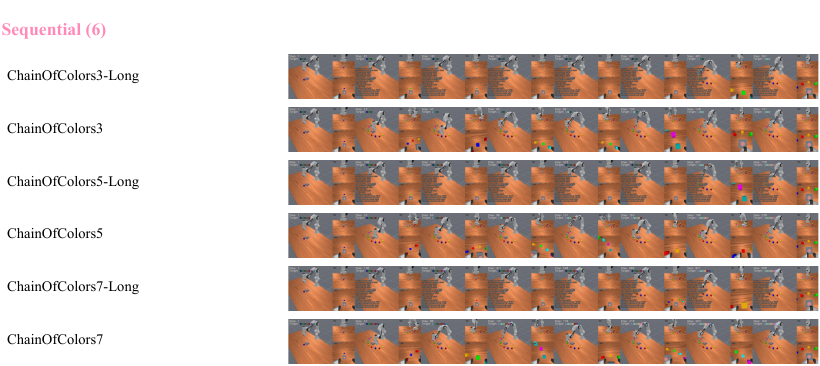}
  \vspace{-8pt}
  \caption{\textbf{Rollout filmstrips for the 6 Sequential tasks, one row per task, 8 evenly spaced frames per rollout, left to right.} The header color is this memory type's color throughout the paper (Table~\ref{tab:taxonomy}).}
  \label{fig:contact-sheet-sequential}
\end{figure}

\begin{figure}[htbp]
  \centering
  \includegraphics[width=\linewidth,height=0.92\textheight,keepaspectratio]{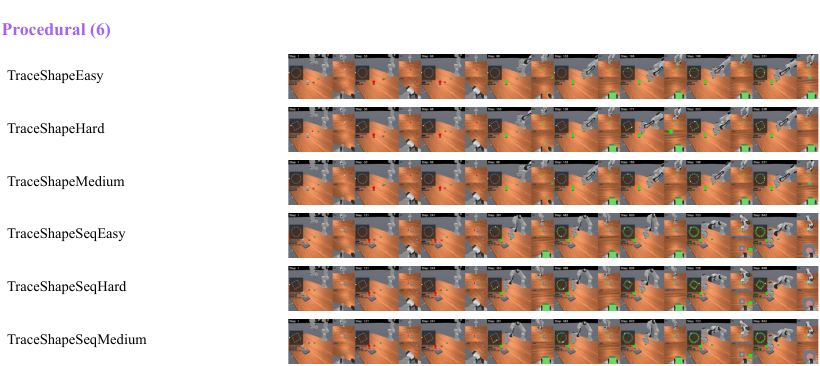}
  \vspace{-8pt}
  \caption{\textbf{Rollout filmstrips for the 6 Procedural tasks, one row per task, 8 evenly spaced frames per rollout, left to right.} The header color is this memory type's color throughout the paper (Table~\ref{tab:taxonomy}).}
  \label{fig:contact-sheet-procedural}
\end{figure}

\begin{figure}[htbp]
  \centering
  \includegraphics[width=\linewidth,height=0.92\textheight,keepaspectratio]{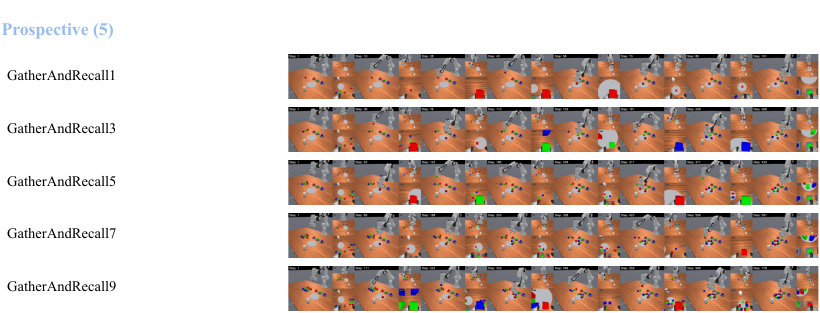}
  \vspace{-8pt}
  \caption{\textbf{Rollout filmstrips for the 5 Prospective tasks, one row per task, 8 evenly spaced frames per rollout, left to right.} The header color is this memory type's color throughout the paper (Table~\ref{tab:taxonomy}).}
  \label{fig:contact-sheet-prospective}
\end{figure}

\begin{figure}[htbp]
  \centering
  \includegraphics[width=\linewidth,height=0.92\textheight,keepaspectratio]{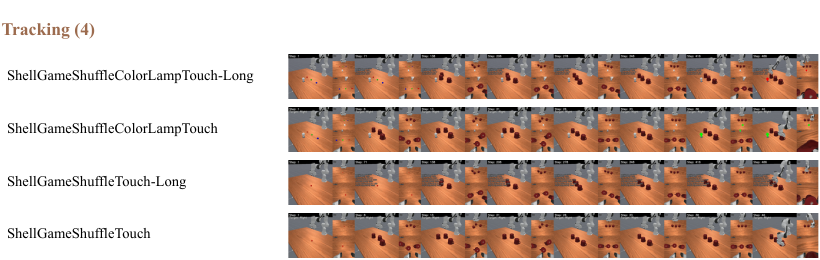}
  \vspace{-8pt}
  \caption{\textbf{Rollout filmstrips for the 4 Tracking tasks, one row per task, 8 evenly spaced frames per rollout, left to right.} The header color is this memory type's color throughout the paper (Table~\ref{tab:taxonomy}).}
  \label{fig:contact-sheet-tracking}
\end{figure}

\begin{figure}[htbp]
  \centering
  \includegraphics[width=\linewidth,height=0.92\textheight,keepaspectratio]{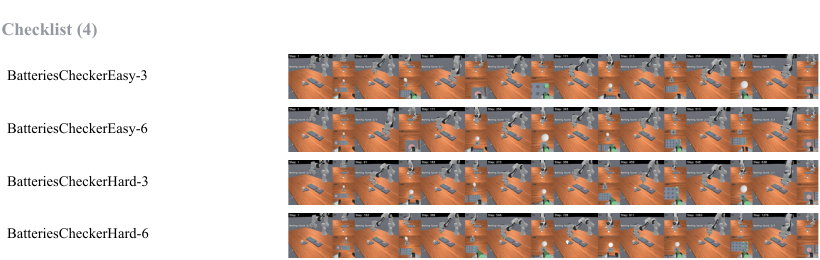}
  \vspace{-8pt}
  \caption{\textbf{Rollout filmstrips for the 4 Checklist tasks, one row per task, 8 evenly spaced frames per rollout, left to right.} The header color is this memory type's color throughout the paper (Table~\ref{tab:taxonomy}).}
  \label{fig:contact-sheet-checklist}
\end{figure}

Figures~\ref{fig:contact-sheet-object}--\ref{fig:contact-sheet-checklist} tile a
filmstrip for every one of the \NumTasks{} tasks, one
sheet per memory type, as a visual index into
Appendix~\ref{sec:appendix-families}'s per-family detail. These frames are the
render channel described there, so they carry debug text the released
observations do not. A single sheet of all
\NumTasks{} tasks is taller than a page, so it prints at a fraction of its
natural size and its task labels become unreadable. One sheet per memory type
fits the page's width instead and prints at full size.

\end{document}